\ifdefined\pdfminorversion\pdfminorversion=7\fi
\documentclass{holidaypaper}

\newcommand{\hdayfactory}{\textsc{HolidayFactory 1.0}\xspace}
\newcommand{\benchmark}{\textsc{24/7-Bench}\xspace}
\newcommand{\eris}{\textsc{Eris}\xspace}

\newcommand{\friday}{\texorpdfstring{\textsc{Friday}}{FRIDAY}\xspace}
\newcommand{\hdaynewton}{\texorpdfstring{\textsc{Holiday-\allowbreak Sim}}{Holiday-Sim}\xspace}

\providecommand{\evup}[1]{\mathop{\uparrow}\nolimits_{#1}}
\providecommand{\evdn}[1]{\mathop{\downarrow}\nolimits_{#1}}

\newcommand{\Autoref}[1]{%
  \begingroup%
  \def\algorithmautorefname{Algorithm}%
  \def\chapterautorefname{Chapter}%
  \def\sectionautorefname{Section}%
  \def\subsectionautorefname{Section}%
  \def\theoremautorefname{Theorem}%
  \def\assumptionautorefname{Assumption}%
  \def\equationautorefname{eq.}%
  \def\definitionautorefname{Definition}%
  \def\lemmaautorefname{Lemma}%
  \def\corollaryautorefname{Corollary}%
  \autoref{#1}%
  \endgroup%
}

\usepackage{float,needspace,xurl}
\usepackage{subcaption}

\newcolumntype{Y}{>{\raggedright\arraybackslash}X}
\newcommand{\src}[1]{\textsuperscript{\hyperref[src:#1]{\scriptsize #1}}}
\makeatletter
\newcommand{\promptboxflush}{\par\if@noskipsec\leavevmode\par\nobreak\vspace{1mm}\else\addvspace{5mm}\fi}
\makeatother
\newtcolorbox{promptbox}[1]{
  enhanced,breakable,sharp corners,
  colback=holidaypaper,colframe=holidayline,boxrule=0.7pt,
  borderline north={2pt}{0pt}{holidaygreen},
  title={#1},colbacktitle=holidaypaper,coltitle=holidayink,
  fonttitle=\small\bfseries,fontupper=\small,
  left=5mm,right=5mm,lefttitle=5mm,righttitle=5mm,toptitle=2mm,bottomtitle=1.5mm,
  top=2.5mm,bottom=3mm,after skip=5mm,before={\promptboxflush\noindent}
}
\DefineVerbatimEnvironment{reportprompt}{Verbatim}{
  fontsize=\footnotesize,fontfamily=phv,
  breaklines=true,breakanywhere=false,breaksymbolleft={},tabsize=2,
  listparameters={\setlength{\topsep}{0pt}\setlength{\partopsep}{0pt}}
}
\newcommand{\figureslot}[3]{%
  \IfFileExists{#1}{\includegraphics[width=\linewidth]{#1}}{%
    \begin{tcolorbox}[sharp corners,colback=holidaypaper,colframe=holidayline,boxrule=0.7pt,
      height=#3,valign=center,halign=center,left=5mm,right=5mm]
      {\small\color{holidayslate}#2}\par
      \vspace{2mm}{\footnotesize\color{holidayslate}Figure reserved for source-grounded artwork}
    \end{tcolorbox}}}

\usepackage{makecell}
\usepackage{amsthm}

\usepackage{multirow}
\usepackage{wrapfig}
\newcolumntype{C}{>{\centering\arraybackslash}X}
\usepackage[normalem]{ulem}
\usepackage[bottom]{footmisc}

\providecommand*{\algorithmautorefname}{Algorithm}
\ExplSyntaxOn
\seq_new:N \l__hday_contributors_seq
\NewDocumentCommand{\contributors}{m m}
  {
    \seq_set_from_clist:Nn \l__hday_contributors_seq {#2}
    \seq_sort:Nn \l__hday_contributors_seq
      {
        \str_compare:eNeTF {##1} > {##2}
          { \sort_return_swapped: }
          { \sort_return_same: }
      }
    \par\noindent
    {\bfseries\color{holidayink}#1}\enspace
    \seq_use:Nn \l__hday_contributors_seq { ,~ }
    \par\medskip
  }
\ExplSyntaxOff

\makeatletter
\titleformat{\paragraph}[runin]{\bfseries\color{holidayink}}{}{0pt}{}[\@addpunct{.}]
\makeatother

\title{HolidayFactory 1.0\\{\fontsize{15}{18}\selectfont Simulating, Generating, and Evaluating Industrial Robot Tasks}}
\author{Holiday Robotics\textsuperscript{\textdagger}}
\affiliation{}
\shorttitle{HolidayFactory 1.0}
\paperlabel{TECHNICAL REPORT}
\date{November 7, 2026}

\begin{document}
\maketitle
{\renewcommand{\thefootnote}{\fnsymbol{footnote}}\footnotetext[2]{See \Autoref{sec:contributors} for contributors and acknowledgments.}}
\vspace{-25pt}
\begin{abstract}
Robots are typically evaluated in episodes that begin from a reset, while industrial work consists of repeated jobs whose outcomes affect subsequent operations. 
Evaluating such work at scale requires simulation of task-critical physical interactions, construction of industrial environments, and measurements of performance as scene state persists.
We present \hdayfactory, a framework for simulating, generating, and evaluating industrial robot work. 
Given a human-specified workplace and task, the framework defines a world pairing a scenario of scene variants and process logic with an evaluation defining a job, metrics, and a rubric. 
\hdayfactory comprises three components. 
\hdaynewton extends Newton's SDF-based hydroelastic contact with additional constitutive laws and a semi-implicit primal-dual solver, couples stateful actuator dynamics to articulated motion, and derives tactile and joint observations from simulated states.
Agentic world construction generates simulation-ready assets and scene variants using geometric and interaction checks, and proposes task-conditioned evaluations for human confirmation and executable checks.
\benchmark introduces seven industrial tasks in which jobs are evaluated without resetting the scene, measuring completed, failed, and unfinished jobs, throughput, latency, and motor energy.
Controlled studies characterize the proposed contact, actuator, and tactile models and examine the effectiveness of verification-guided world construction.
Experiments on scene and evaluation generation further reveal the limitations of model critics and the difficulty of ensuring that generated evaluations faithfully capture task requirements.
We examine job-level evaluations with scripted trajectories (e.g., code-as-policy) but have not yet compared policies over persistent runs.
\end{abstract}
\vspace{-20pt}
\medskip
\begin{figure}[H]
\centering
\includegraphics[width=\linewidth]{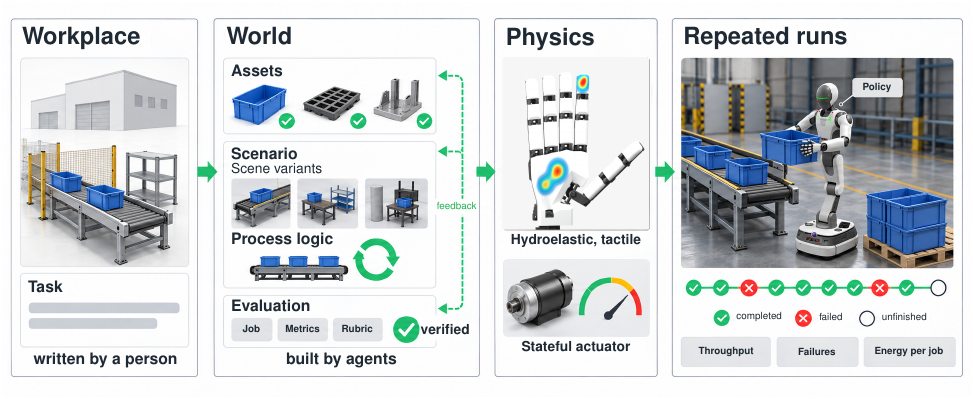}
\par\smallskip
\newcommand{\overviewtask}[2]{\begin{minipage}[t]{0.138\linewidth}\centering\includegraphics[width=\linewidth]{figures/benchmark/#1}\\[-0.2em]{\scriptsize #2}\end{minipage}}
\overviewtask{task_line_unloading.jpg}{Conveyor loading}\hfill
\overviewtask{task_flow_rack.jpg}{Flow-rack loading}\hfill
\overviewtask{task_cart_push.jpg}{Cart push}\hfill
\overviewtask{task_jig_loading.jpg}{Jig loading}\hfill
\overviewtask{task_press_tending.jpg}{Press tending}\hfill
\overviewtask{task_valve_operation.jpg}{Valve operation}\hfill
\overviewtask{task_barcode_routing.jpg}{Barcode routing}
\caption{An overview of \hdayfactory.
Human-specified workplaces and tasks guide agents to generate assets, scene variants, and process logic, and to propose evaluations for human confirmation.
\hdaynewton simulates contact, actuation, and sensor responses.
The top row is schematic, and the bottom panel shows \benchmark including seven industrial robot tasks in simulation.}
\label{fig:overview}
\end{figure}

\clearpage
\begingroup
\small
\hypersetup{linkcolor=black}
\tableofcontents
\endgroup
\clearpage
\Needspace{5\baselineskip}
\section{Introduction}
\label{sec:introduction}

Robotics has a simple scaling problem. The most trustworthy evaluation happens on the real robot, but real-robot evaluation is slow, expensive, difficult to parallelize, and hard to reproduce.
Each trial consumes hardware time and operator attention, scenes must be prepared and reset, and failures can damage the robot or its surroundings.
These costs grow quickly for generalist policies that must be tested across many tasks, objects, and environments.
Recent work therefore treats simulation not only as a training tool but also as an evaluation substrate that can be regenerated and replayed at much larger scale~\citep{li2024simpler,jain2025polaris,jangir2025robotarena}.

Most simulation benchmarks built for this purpose, however, emphasize homes and everyday activities.
BEHAVIOR-1K defines 1,000 human-centered activities across houses, gardens, restaurants, offices, and related settings, while RoboCasa and RoboCasa365 focus on large-scale everyday manipulation in household environments~\citep{li2024behavior1k,nasiriany2024robocasa,nasiriany2026robocasa365}.
Other benchmarks study whole-body humanoid control or long-horizon indoor mobile manipulation~\citep{sferrazza2024humanoidbench,jaafar2024lambda}.
These suites cover much of general-purpose embodied behavior, but each scores an episode that starts from a reset.
We instead focus on \emph{repetitive industrial work}.

Factory work is rarely a single isolated pick-and-place episode, and it is where robots are already deployed at scale. Factories installed more than 600,000 industrial robots in 2025, and the operational stock reached a record 5 million~\citep{ifr2026worldrobotics}.
Parts keep arriving, racks fill and empty, and the next job depends on the state left by previous jobs, a condition that reset-free robot learning has begun to address~\citep{sharma2021earl,sharma2023medal,wu2026cronos}.
A failed grasp may require a retry, a different stance, an intervention, or a safe stop, and naive retries often repeat the same mistake~\citep{hao2026far}.
Throughput matters, as in bin picking, where Dex-Net 4.0 reports more than 300 mean picks per hour alongside its reliability~\citep{mahler2019dexnet4}, but so do unsafe contacts, dropped objects, blocked workspaces, recovery time, and the accumulated effect of small errors.
Small per-job failure rates also compound. Under the simplifying assumption of independent outcomes, a policy with a 95\% single-job success rate completes 20 consecutive jobs only 36\% of the time, while a 99\% rate yields 82\%. Persistent scene state can additionally make job outcomes dependent.

\begin{center}
\emph{``The race is not always to the swift.''}\\[-0.2em]
{\small --- \emph{The Hare \& the Tortoise}, \emph{Aesop for Children}\footnote{\href{https://read.gov/aesop/025.html}{Library of Congress}.}}
\end{center}

We introduce \hdayfactory, a framework for simulating, generating, and evaluating industrial robot work. 
Given a human-specified workplace $\mathcal W$ and tasks $\mathcal T$, each task is associated with a world $\xi=(S,E)$ (\S~\ref{sec:overview}).
The scenario $S$ comprises scene variants and process logic, while the evaluation $E$ defines a job, its metrics, and a rubric. 
The world specification is fixed before policy evaluation, while the physical state evolves across jobs within each run without a scene reset.
Agentic construction produces and verifies assets and scene variants, writes process logic, and supports the generation of evaluations from task requirements. The current framework requires human confirmation of generated evaluation proposals. 
\hdaynewton provides the simulated dynamics and sensor observations through which a separately chosen policy $\pi$ interacts with a scene. 

\subsection{Research questions and organization}
\label{sec:research-questions}
\hdayfactory addresses a central research question:
\begin{center}
    \textit{How can industrial robot tasks be transformed into physically executable worlds \\ in which robot performance can be evaluated over repeated jobs?}
\end{center}
We approach it through three research questions.
\begin{itemize}
    \item \textbf{(RQ1)} How to simulate the contact, actuation, and sensing that determine physical outcomes
    \item \textbf{(RQ2)} How to construct task-conditioned worlds that pair executable scenarios with their evaluations
    \item \textbf{(RQ3)} How to measure robot performance across repeated jobs as the physical state persists
\end{itemize}
\textbf{(RQ1)} Simulation must represent whether a part seats in a pocket with little clearance, whether a grasp holds, and whether the motors can deliver the required effort. Differences between simulated and real interactions can affect the reliability of simulation-based evaluation~\citep{torne2024rialto,han2025re3sim,jain2025polaris,jiang2025gsworld}.
\textbf{(RQ2)} Industrial tasks require scenarios that support their intended operations. Each scenario must contain task-conditioned assets, valid scene configurations, and executable process logic. 
Scaling their construction requires verifying task-relevant constraints during generation.
\textbf{(RQ3)} Evaluation needs to translate continuous execution into measurements of work. 
Jobs require well-defined start, completion, and failure conditions, with consistent metrics and reporting criteria across policies and scene variants. 

\hdayfactory makes three contributions:
\begin{itemize}
    \item \textbf{\hdaynewton} (\S~\ref{sec:simulation}). A physics engine that extends hydroelastic contact on Newton\footnote{An open-source physics engine~\citep{newton} built on NVIDIA Warp~\citep{warp2022}.}, adds stateful actuators, and resolves both with a custom solver. 
    Contact patches also support built-in magnetic tactile sensing, while actuator and joint states provide telemetry as policy observations.
    
    \item \textbf{Agentic world} (\S~\ref{sec:worlds}). Coding agents generate and refine task-conditioned assets and scene variants through verification-guided generation, using executable checks to enforce task-relevant constraints (\S~\ref{sec:eris}, \S~\ref{sec:world-generation}). 
    The framework also generates visual backgrounds (\S~\ref{sec:photoreal-background}), while a separate agent proposes evaluation for human confirmation and deterministic compilation (\S~\ref{sec:eval-authoring}).
    
    \item \textbf{\benchmark} (\S~\ref{sec:tasks}). Seven industrial tasks define repeated jobs on persistent state. Their job outcomes and any run-level success criteria use privileged simulator state. The benchmark reports throughput, job latency, failed and unfinished jobs, and motor energy per completed job (\S~\ref{sec:evaluation}). Runs are grouped by scene variant and by the variation axes a comparison declares (\S~\ref{sec:eval-regimes}).
\end{itemize}

The experiments assess physical and tactile responses, asset and scene generation, and the verification and semantic agreement of generated evaluations. They do not yet establish end-to-end generation of the seven benchmark worlds or compare policies across those tasks.

The remainder of this paper is organized as follows. \Autoref{sec:overview} establishes the common terminology and framework overview. \Autoref{sec:simulation} presents \hdaynewton for physical simulation. \Autoref{sec:worlds} describes agentic world construction and verification. \Autoref{sec:tasks} introduces \benchmark for evaluating persistent industrial work. \Autoref{sec:discussion} discusses limitations and open research questions.

\section{Overview}
\label{sec:overview}

\subsection{Preliminaries}
\label{sec:overview-preliminaries}

A \emph{workplace} $\mathcal W$ is a human-specified description of an industrial site, including its equipment, the objects it handles, and where the robot operates.
A workplace may host multiple \emph{tasks}, each specifying the work requested of a robot policy.
A task $\mathcal T$ is expressed in natural language or structured fields, such as continuously loading totes onto a conveyor.
\autoref{fig:workplace-overview} illustrates four tasks within an inbound conveyor workplace.

For each task $\mathcal T$, a \emph{world} $\xi=(S,E)$ pairs a \emph{scenario} $S$ with an \emph{evaluation} $E$.
The scenario $S=(s_{1:N_s},\ell)$ consists of \emph{scene} variants $s_1,\ldots,s_{N_s}$ and \emph{process logic} $\ell$.
Each scene specifies the robot and objects, their geometry, physical parameters, initial state, background, and cameras.
The process logic governs workplace operations through continuous effects, such as conveyor motion, triggered events, such as press activation, and queues that deliver or remove parts.
Following scenario-based testing in automated driving, a scene describes a snapshot of the environment and a scenario its evolution over time~\citep{ulbrich2015scene}.
Our scenario corresponds to a logical scenario, while each run instantiates a concrete scenario using one scene variant $s_i$ and process logic $\ell$~\citep{menzel2018scenarios}.

A \emph{job} is a unit of requested work, such as moving one tote from a pallet onto a conveyor, with its own progress and outcome.
The \emph{evaluation} $E=(j,m_{1:N_m},c)$ defines the job $j$ through events on privileged simulator state marking its start, completion, and failure.
The \emph{metrics} $m_{1:N_m}$ aggregate job records and events into quantities such as completed jobs, job latency, and energy.
The \emph{rubric} $c$ specifies run termination, reporting rules, and optional success criteria (\S~\ref{sec:evaluation}).

A \emph{run} executes a policy $\pi$ on task $\mathcal T$ in one scene variant $s_i$ of its world.
An \emph{evaluation regime}, defined separately from the world, specifies how scene variants and experimental conditions are sampled across runs (\S~\ref{sec:eval-regimes}).
The same evaluation applies to different policies, while their observations, control authority, and callable skills are part of the experimental condition.
The world specification remains fixed during policy evaluation, although the scene state evolves within each run.
\Autoref{tab:nomenclature} summarizes the terminology.

For a run in scene variant $s=s_i$,
\begin{equation}
  x_{t+1}=F_{s,\ell}(x_t,u_t),\qquad
  o_t=O_s(x_t),\qquad
  u_t=\pi(o_{0:t},\mathcal T),\qquad
  \mathcal Q=\operatorname{Eval}_E(\tau),
  \label{eq:run-execution}
\end{equation}
where $x_t$, $u_t$, and $o_t$ denote the simulation state, action, and observation, respectively; $F_{s,\ell}$ advances the state under scene $s$ and process logic $\ell$, while $O_s$ produces the policy observations, including images and nonvisual sensor readings, and $\tau=(x_{0:N_{\mathrm{step}}},o_{0:N_{\mathrm{step}}},u_{0:N_{\mathrm{step}}-1})$ is the recorded trajectory over $N_{\mathrm{step}}$ steps. The evaluation report $\mathcal Q$ contains job records, metric values, the termination reason, and, when required by the rubric, a verdict $Y$. A report-only rubric does not assign a verdict.
Core execution symbols are summarized in \autoref{tab:core-notation}.
Generated backgrounds affect $O_s$ but not $F_{s,\ell}$, allowing recorded trajectories to be re-rendered without repeating the physical simulation.

\makeatletter
\providecommand{\tbar}{\textcolor{gray!55}{\rule[-\dp\@arstrutbox]{0.6pt}{\dimexpr\ht\@arstrutbox+\dp\@arstrutbox\relax}}\hspace{4pt}}
\makeatother
\begin{figure}[t]
\centering
\input{report/figure_workplace_overview}
\vspace{1.5em}
\footnotesize
\captionsetup{type=table}
\caption{Terminology of \hdayfactory.
}\label{tab:nomenclature}
\setlength{\tabcolsep}{4pt}
\renewcommand{\arraystretch}{1.15}
\begin{tabularx}{\linewidth}{@{}lY@{}}
\toprule
\tablehead{Term} & \tablehead{Definition}\\
\midrule
Workplace $\mathcal W$ & Human-specified description of an industrial site, its equipment, objects, and robot workspace. \\
Task $\mathcal T$ & Human-specified instruction given to a robot policy.\\
World $\xi$ &  Scenario and evaluation associated with one task.\\
\tbar{}Scenario $S$ & Executable environment of a task, made of scene variants and the process logic that changes them over time.\\
\tbar\tbar{}Scene $s_i$ & One variant of the objects and robot, with their physical parameters and placement, and the background.\\
\tbar\tbar{}Process logic $\ell$ & What the site does on its own, such as per-step effects and changes triggered by a condition.\\
\tbar{}Evaluation $E$ & Job definition, metrics, and rubric for a task.\\
\tbar\tbar{}Job $j$ & Requested piece of work defined by start, completion, and failure events on privileged state.\\
\tbar\tbar{}Unit & Object that a job acts on, such as one tote.\\
\tbar\tbar{}Metric $m$ & Quantity aggregated from job records and events, such as completed jobs, latency, or energy.\\
\tbar\tbar{}Rubric $c$ & Run termination and reporting rules, with optional success criteria.\\
Policy $\pi$ & Method under test, selected independently of the world.\\
\bottomrule
\end{tabularx}
\end{figure}

\begin{table}[t]
\centering\small
\caption{Core execution notation. World components are defined in Table~\ref{tab:nomenclature}; local physical quantities are defined where used.}
\label{tab:core-notation}
\begin{tabularx}{\linewidth}{@{}lX@{}}
\toprule
\tablehead{Symbol} & \tablehead{Meaning}\\
\midrule
$N_s,N_m$ & Numbers of scene variants and metric definitions.\\
$x_t,u_t,o_t$ & Simulator state, policy action, and observation at run step $t$.\\
$\pi$ & Policy mapping observation history and task to an action.\\
$F_{s,\ell},O_s$ & State transition under scene $s$ and process logic $\ell$, and observation model for $s$.\\
$\tau,N_{\mathrm{step}}$ & Recorded trajectory and its actual number of steps.\\
$\mathcal Q=\operatorname{Eval}_E(\tau)$ & Evaluation report under specification $E$.\\
$Y$ & Optional run verdict: pass, fail, or undecided.\\
$\Delta t,h,N_{\mathrm{sub}}$ & Frame interval, physics substep interval, and substeps per frame; $h=\Delta t/N_{\mathrm{sub}}$.\\
$T_{\mathrm{run}},t_{\mathrm{run}}$ & Maximum run duration and actual elapsed simulation time.\\
$\tau_{\mathrm{job}}$ & Timeout duration measured from a job's start.\\
$N_{\mathrm{goal}},N_{\mathrm{done}}$ & Target and actual completed-job counts.\\
$N_{\mathrm{env}}$ & Number of parallel simulation instances.\\
\bottomrule
\end{tabularx}
\end{table}

\paragraph{Embodiment} We use \friday, a wheeled humanoid robot about 1.8\,m tall with a differential-drive base, a five-joint waist, two seven-joint arms, a three-joint head, and two five-finger hands whose palms and inner finger surfaces carry compliant tactile skin, 75 joints in all.
Head and wrist cameras, together with an IMU, LiDAR, odometry, joint telemetry, and the tactile skin, supply the observations $o_t$ (\S~\ref{app:robot-simulation}).

\subsection{\hdaynewton}
\label{sec:overview-newton}

\hdaynewton provides the physical state transitions and sensor responses used to execute robot policies in \hdayfactory (\S~\ref{sec:simulation}).
It extends Newton's sparse-SDF-based hydroelastic contact with contact constitutive models, a semi-implicit primal-dual solver, and stateful actuator dynamics.
The contact patches also provide inputs to magnetic tactile sensing, while the actuator and mechanism states produce joint telemetry.

Each scene $s_i$ supplies the geometry and physical parameters for these models, while process logic $\ell$ governs workplace operations during execution.

\subsection{Agentic world}
\label{sec:overview-world}

An agentic pipeline generates task-conditioned assets, scene variants, and evaluation proposals (\S~\ref{sec:worlds}).
Asset generation produces simulation-ready objects and checks their geometry and task-relevant interactions in \hdaynewton.
Scenario generation assembles these assets into scene variants with physical parameters and photorealistic backgrounds, while a scenario-authoring agent writes the process logic $\ell$.
Evaluation generation proposes a job, metrics, and rubric for human confirmation before compilation and executable checks.

Formally,
\begin{equation}
  S\sim G(\mathcal T,\mathcal W)
  \ \text{with}\ V_{\mathrm{scen}}(S)=1,\qquad
  E\sim A(\mathcal T,S)
  \ \text{with}\ V_{\mathrm{eval}}(E,S)=1,
  \label{eq:agentic-world}
\end{equation}
where $G$ assembles scene variants with the specified process logic, $A$ generates evaluations with human confirmation, and $V_{\mathrm{scen}}$ and $V_{\mathrm{eval}}$ denote their respective checks.

\begin{table}[t]
\centering\scriptsize
\caption{Comparison of representative robot simulation benchmarks.}
\label{tab:benchmark-scope}
\setlength{\tabcolsep}{2pt}
\renewcommand{\arraystretch}{1.18}
\begin{tabularx}{\linewidth}{@{}>{\raggedright\arraybackslash}p{0.13\linewidth}YYYYYY@{}}
\toprule
\tablehead{Benchmark} & \tablehead{Scale} & \tablehead{Environment} & \tablehead{Horizon} & \tablehead{Simulator} & \tablehead{Scored unit} & \tablehead{Metrics}\\
\midrule
LIBERO \citep{liu2023libero} & 130 tasks & Tabletop & 30\,s & MuJoCo (robosuite) & Episode & Success rate\\
\addlinespace
ManiSkill3 \citep{taomaniskill3} & 12 domains & Tabletop\newline Indoor & 2.5-12.5\,s$^{\dagger}$ & PhysX (SAPIEN) & Episode & Success rate\newline Episode return\\
\addlinespace
RoboCasa365 \citep{nasiriany2026robocasa365} & 365 tasks & Kitchen & 15\,s-6\,min & MuJoCo (robosuite) & Episode & Success rate\\
\addlinespace
BEHAVIOR-1K \citep{li2024behavior1k} & 50 tasks & Household & 2.4-29\,min$^{\dagger}$ & PhysX (OmniGibson) & Episode & Goal completion\newline Execution time\newline Motion distance\\
\addlinespace
RoboTwin 2.0 \citep{chen2025robotwin2} & 50 tasks & Tabletop & 400-1,700 steps$^{\dagger}$ & PhysX (SAPIEN) & Episode & Success rate\\
\addlinespace
HumanoidBench \citep{sferrazza2024humanoidbench} & 27 tasks & Obstacle courses\newline Indoor & 10-20\,s & MuJoCo & Episode & Success rate\newline Episode return\\
\addlinespace
Factory \citep{narang2022factory} & 3 tasks & Assembly cell & 1.7\,s-2.3\,min & PhysX (Isaac Gym) & Episode & Success rate\\
\midrule
\rowcolor{holidaygreendark!15}
\benchmark & 7 workplaces\newline Open task set & Industrial cell & $\sim$ hours & \hdaynewton & Job\newline Task & Success rate\newline Throughput\newline Job latency\newline Failures\newline Violations\newline Energy \& power\\
\bottomrule
\end{tabularx}
\smallskip
\begin{minipage}{\linewidth}
\scriptsize
Entries that the cited papers do not state follow each benchmark's released code and documentation~\citep{maniskill_eval_docs,maniskill_tasks_docs}, and the BEHAVIOR-1K entries follow its 2025 challenge~\citep{behavior_eval_docs}. $^{\dagger}$Episode limits of the tabletop gripper tasks of ManiSkill3. BEHAVIOR sets each limit at twice the mean human demonstration time of the task from the 2025 challenge demonstrations. A RoboTwin step is one policy action, a planned motion of variable duration.
\end{minipage}
\end{table}

\subsection{\texorpdfstring{\benchmark}{24/7-Bench}}
\label{sec:overview-benchmark}

\benchmark comprises seven industrial tasks across seven workplaces (\S~\ref{sec:tasks}).
Each task involves repeated jobs within a scene whose state persists between jobs.
Some tasks use conveyors, events, or queues to supply and process work, while others operate on a fixed initial set of units.
The benchmark records completed, failed, timed-out, and unfinished jobs, reporting throughput, job latency, violations, and motor energy under specified evaluation regimes.
A rubric may additionally define a run-level success criterion.
\autoref{tab:benchmark-scope} compares its execution horizon, scoring units, and metrics with representative simulation benchmarks.
\Needspace{5\baselineskip}

\section{\hdaynewton}
\label{sec:simulation}

\begin{figure}[t]
\centering
\includegraphics[width=\linewidth]{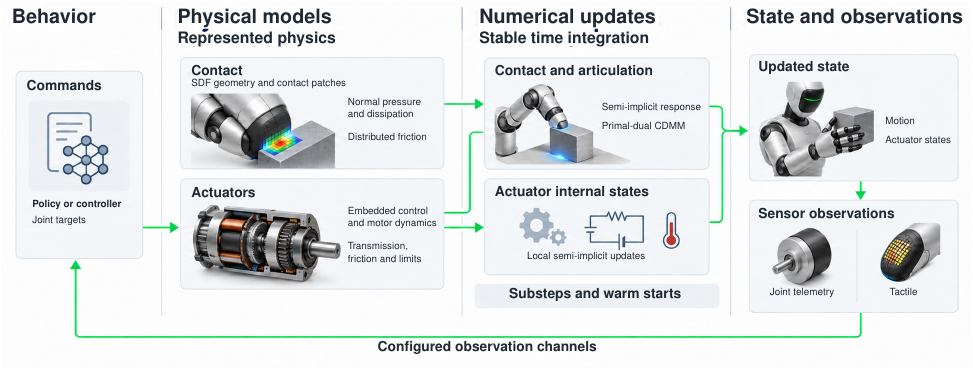}
\caption{An overview of \hdaynewton. 
Illustrations are schematic, not simulator outputs.
}
\label{fig:simulation-overview}
\end{figure}

\hdaynewton combines Newton's sparse-SDF hydroelastic contact with the constitutive models and a semi-implicit primal-dual solver, and couples stateful actuators to articulated dynamics (\autoref{fig:simulation-overview}).
For contact, it models pressure, damping, and friction over finite patches and resolves their coupled response within each simulation step (\S~\ref{sec:contact}).
For actuation, it models motor controllers, electrical limits and thermal dynamics, transmissions, and internal friction, contributing joint effort and impedance to the articulated solve (\S~\ref{sec:actuator}).

\subsection{SDF-hydroelastic contact}
\label{sec:contact}

Precision industrial manipulation depends on contact fidelity. A machined part must seat in a jig pocket with little clearance, a tote must rest stably on a stack, and a grasp must hold a heavy box without slipping or twisting.
Representative benchmarks use convex decomposition and sparse point contacts (\autoref{tab:benchmark-contact}), which can fill in narrow features and collapse distributed loads onto a handful of points.
Hydroelastic contact, introduced by \citet{elandt2019pressure} and implemented in Drake~\citep{drake,masterjohn2022velocity} with tetrahedral volume meshes, addresses the distributed-load limitation by integrating a pressure field over a finite contact patch. Preparing such meshes for arbitrary CAD assets can be laborious, and contact queries can be costly at benchmark scale.
Newton~\citep{newton} implements the same pressure-balance principle on sparse signed distance fields (SDFs), avoiding tetrahedral meshing.

Building on Newton, \hdaynewton addresses three challenges in precision industrial manipulation: contact modeling, numerical stability, and memory efficiency.
First, it extends Newton's linear pressure law with compression hardening, compression-dependent damping, speed-dependent friction, and planar ground contact (\S~\ref{sec:contact-modeling}).
Second, a semi-implicit primal-dual solver resolves stiff, coupled contacts across articulated bodies (\S~\ref{sec:contact-solver}).
Third, asset-level SDF budgets and instance-chunked collision control memory consumption, which can otherwise be dominated by the finest-resolution asset (\S~\ref{sec:eris-verification} and \S~\ref{app:collision-memory}).
The resulting contact patches also support magnetic tactile sensing (\S~\ref{sec:tactile}).

\begin{table}[t]
\centering\footnotesize
\caption{Rigid-body contact formulations in representative robot simulation benchmarks. Asterisks mark the default solver of each simulator.}
\label{tab:benchmark-contact}
\setlength{\tabcolsep}{3pt}
\begin{tabularx}{\linewidth}{@{}>{\raggedright\arraybackslash}p{0.22\linewidth}YYYY@{}}
\toprule
\tablehead{System} & \tablehead{Simulator engine} & \tablehead{Contact geometry} & \tablehead{Contact physics} & \tablehead{Contact solver}\\
\midrule
LIBERO \citep{liu2023libero} & MuJoCo (robosuite) & Convex decomp. & Compliant point & Newton$^{*}$\\
ManiSkill3 \citep{taomaniskill3} & PhysX (SAPIEN) & Convex decomp. & Point & TGS$^{*}$\\
RoboCasa365 \citep{nasiriany2026robocasa365} & MuJoCo (robosuite) & Convex decomp. & Compliant point & Newton$^{*}$\\
BEHAVIOR-1K \citep{li2024behavior1k} & PhysX (OmniGibson / Isaac Sim) & Convex decomp. & Point & TGS$^{*}$\\
RoboTwin 2.0 \citep{chen2025robotwin2} & PhysX (SAPIEN) & Convex decomp. & Point & TGS$^{*}$\\
HumanoidBench \citep{sferrazza2024humanoidbench} & MuJoCo & Convex decomp. & Compliant point & Newton$^{*}$\\
Factory \citep{narang2022factory} & PhysX (Isaac Gym) & SDF & Point & TGS \\
\midrule
\rowcolor{holidaygreendark!15}
\hdayfactory & \hdaynewton & SDF & Hydroelastic & CDMM\\
\bottomrule
\end{tabularx}
\smallskip
\begin{minipage}{\linewidth}
\scriptsize
TGS denotes temporal Gauss-Seidel, CDMM denotes Constrained Dynamics via Method of Multipliers (\S~\ref{sec:contact-solver}), and Newton in the solver column is MuJoCo's Newton solver, unrelated to the Newton engine. Where the cited papers do not state a setting, the entry follows the benchmark's released assets and configuration and the engine documentation~\citep{mujoco_docs,physx_docs}. These entries list only the configurations used for comparison, and each engine supports other contact representations as well.
\end{minipage}
\end{table}

\subsubsection{Geometry and constitutive laws}
\label{sec:contact-modeling}

\hdaynewton adopts Newton's sparse-grid SDF representation to model nonconvex contact geometry, including narrow gaps, at lower memory cost than uniformly fine volumetric grids (\S~\ref{app:sdf-queries}).
For a body $A$, the signed distance field $\phi_A(\mathbf{x})$ is negative inside the body and positive outside. 
It also supplies the depth-dependent pressure field of the hydroelastic model through Eq.~\eqref{eq:sdf-pressure-field}. 
Factory~\citep{narang2022factory} also uses SDFs to represent tight-clearance assembly geometry, but resolves contact at points and has no pressure patches.

\begin{figure}[t]
\centering
\includegraphics[width=\linewidth]{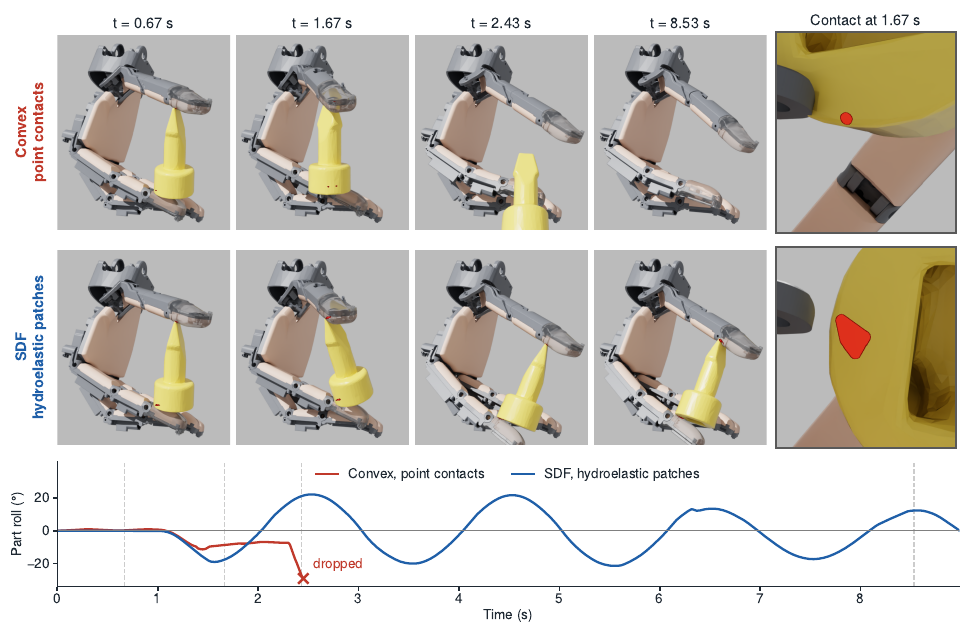}
\caption{\friday fingertips rolling a slender tapered part under the same finger motion. Top, point contacts on convex pieces. In this configuration, the part drops at 2.4\,s. Bottom, sparse-grid SDFs with hydroelastic patches hold the part through the whole motion. The right column shows the contact on the part at 1.67\,s, a single point against a pressure patch.}
\label{fig:friday-hand-collider}
\end{figure}

Where two bodies overlap, the contact surface is the set of points at which their pressure fields balance~\citep{elandt2019pressure}:
\begin{equation}
\mathcal{S}_{AB} = \left\{ \mathbf{x}\in\Omega_A\cap\Omega_B \;\middle|\; p_A(\mathbf{x})=p_B(\mathbf{x}) \right\},
\label{eq:hydroelastic-contact-surface}
\end{equation}
where $\Omega_A$ and $\Omega_B$ are the volumes occupied by the two bodies.
Newton extracts this surface directly from the sparse-grid SDFs, using hierarchical voxel pruning and marching cubes, and can reduce the patch to a few representative contacts by moment matching while preserving its friction moment.

\hdaynewton specifies two mappings around this surface, from signed distance to pressure and from a pressure patch to contact force and friction.
With a limited set of contacts, the force that supports an object and the moment that resists its tilting depend on how many points the simulator keeps. With a pressure patch, both follow from where and how deeply the bodies overlap.
Both mappings matter across the range of stiffness in industrial contact, from stiff machined parts to compliant pads and objects. 
On the \friday hand, compliant fingertips hold a slender tapered part through patch pressure and friction while it rolls. With point contacts the same motion drops the part (\autoref{fig:friday-hand-collider}).

For the first mapping, signed distance becomes pressure through a constitutive law, which is linear by default:
\begin{equation}
p_A(\mathbf{x}) = k_{h,A}\left[-\phi_A(\mathbf{x})\right]_+,
\qquad [z]_+ = \max(z,0).
\label{eq:sdf-pressure-field}
\end{equation}
Here, $\phi_A$ is the signed distance after any surface offset, and $k_{h,A}$, in $\mathrm{N}/\mathrm{m}^3$, relates pressure to depth. Unlike a point-contact stiffness, the resulting force also grows with the contact area.
Measurements on a silicone robotic fingerpad show that force rises from 1.1 to 11.7\,N as the indentation grows from 1.6 to 4.3\,mm and the contact area spreads from 48 to 87\,mm$^2$~\citep{achilli2022fingerpad}.
To represent additional depth-dependent stiffening, \hdaynewton uses the following law for interior depth $x_A=[-\phi_A(\mathbf{x})]_+$:
\begin{equation}
p_A(\mathbf{x}) = k_{h,A}x_A\left[1+\beta_A\left(\frac{x_A}{\ell_A}\right)^2\right],
\qquad \beta_A\geq0,\quad\ell_A>0.
\label{eq:nonlinear-pad-pressure}
\end{equation}
Here, $k_{h,A}$ sets the initial pressure gradient, $\beta_A$ the amount of hardening, and $\ell_A$ a characteristic depth. Setting $\beta_A=0$ recovers the linear law.
The law does not model the interior of the pad. The depth $\ell_A$ need not equal the pad thickness, and plasticity, creep, and lateral bulging are not represented (\S~\ref{par:nonlinear-pressure}).

The same mapping covers the limit of an infinitely stiff body. Against rigid ground, pressure balance places the contact surface on the plane, so \hdaynewton integrates the compliant body's pressure directly over it without building a ground SDF.
An unbounded plane needs no volumetric field, yet wheels still need patch-level pressure and friction because slip speeds vary across the footprint and produce a net force and moment.
For a ground plane $\Pi_g$ with unit normal $\mathbf{n}_{\Pi_g}$ pointing toward the body, the patch and elastic normal force are
\begin{equation}
\mathcal{S}_{A\Pi_g}=\Omega_A\cap\Pi_g,
\qquad \mathbf{f}_{n}^{\mathrm{el}} = \mathbf{n}_{\Pi_g}\int_{\mathcal{S}_{A\Pi_g}} p_A(\mathbf{x})\,\mathrm{d}A,
\label{eq:plane-hydroelastic-contact}
\end{equation}
evaluated with two-dimensional quadrature over the footprint.

For the second mapping, a pressure patch becomes contact force by integrating the traction over the contact surface:
\begin{equation}
\begin{aligned}
\mathbf{f}_{B\to A} &= \int_{\mathcal{S}_{AB}} \mathbf{t}_{B\to A}(\mathbf{x})\,\mathrm{d}A, \qquad
\boldsymbol{\tau}_{B\to A,\mathbf{o}} &= \int_{\mathcal{S}_{AB}} (\mathbf{x}-\mathbf{o})\times \mathbf{t}_{B\to A}(\mathbf{x})\,\mathrm{d}A,
\end{aligned}
\label{eq:hydroelastic-contact-wrench}
\end{equation}
where $\mathbf{t}_{B\to A}$ is the total traction that body $B$ exerts on body $A$, including friction and damping, and $\mathbf{o}$ is a reference point.
Because the patch grows as the bodies press together, the net force rises faster than linearly even under a linear pressure law. When a resting object tilts, the pressure shifts toward the lower side and produces a moment that opposes the tilt.
The model represents local compliance between nominally rigid bodies. Bulk deformation is not modeled.
Because the patch is reduced to representative contacts, force and moment are approximated, with accuracy set by grid resolution and contact reduction.

\hdaynewton adds Hunt-Crossley damping~\citep{hunt1975coefficient}, which grows with compression and helps a part settle.
A purely elastic contact would return all stored energy, so a dropped part would keep bouncing. This damping removes energy in proportion to both compression and approach speed.
The stiffness $K_i$ of each representative contact is derived from the pressure patch and held fixed within each substep.
For penetration $\delta_i\geq 0$ and normal velocity $v_{n,i}$, positive when the bodies separate, $f_{n,i} = K_i\delta_i\left[1-\alpha v_{n,i}\right]_+$
with zero force outside contact and a dissipation coefficient $\alpha$, in $\mathrm{s}/\mathrm{m}$, shared across contacts.
The corresponding damping coefficient $c_{\mathrm{eff},i}=K_i\alpha\delta_i$
vanishes at first touch and grows as the contact is compressed, so the same approach speed is damped more strongly in a deeper contact.
Loading and unloading therefore follow different force curves without any stored material history, and the force never becomes tensile.

\hdaynewton applies friction over the whole patch to model a grasp's resistance to twisting.
Each representative contact applies friction bounded by its share of the normal load at its own lever arm, approximating the resisting torque from the pressure distribution.
A point-contact manifold can also resist twisting through tangential forces at spatially separated contacts, with the resulting torque depending on their locations and load distribution.
\hdaynewton also lets this friction weaken once sliding starts, so that an interface can hold at rest yet resist less while it slides. With normal force $f_{n,i}\geq 0$, tangential force $\mathbf{f}_{t,i}$, and slip speed $v_{\mathrm{slip},i}=\|\mathbf{v}_{t,i}\|$ at contact $i$, the friction law satisfies
\begin{equation}
\begin{aligned}
\mu_i(v_{\mathrm{slip},i}) &= \mu_{s,i}\left[r+(1-r)e^{-v_{\mathrm{slip},i}/v_c}\right],\\
\|\mathbf{f}_{t,i}\| &\leq \mu_i(v_{\mathrm{slip},i}) f_{n,i},\qquad
\mathbf{f}_{t,i} = -\mu_i(v_{\mathrm{slip},i}) f_{n,i}\frac{\mathbf{v}_{t,i}}{v_{\mathrm{slip},i}}, \qquad v_{\mathrm{slip},i}>0.
\end{aligned}
\label{eq:contact-coulomb-friction}
\end{equation}
Here, $\mu_{s,i}$ is the static coefficient, $r\in[0,1]$ sets the sliding coefficient $r\mu_{s,i}$, and $v_c>0$ is the transition speed. The same exponential decay is used in continuum contact modeling~\citep{comsol_friction}.
Setting $r=1$ recovers constant Coulomb friction. The parameters $r$ and $v_c$ are currently shared across contacts.
Because representative contacts can stick or slip independently, the patch also gives a discrete approximation of partial slip. Tangential elastic deformation, micro-slip, and rolling resistance are not resolved.

Sparse-grid SDFs are used for all volumetric collision geometry, including simple shapes such as cubes, so that every contact receives the same hydroelastic treatment, and only the ground uses the planar path above.

\begin{figure}[t]
\centering
\includegraphics[width=\linewidth]{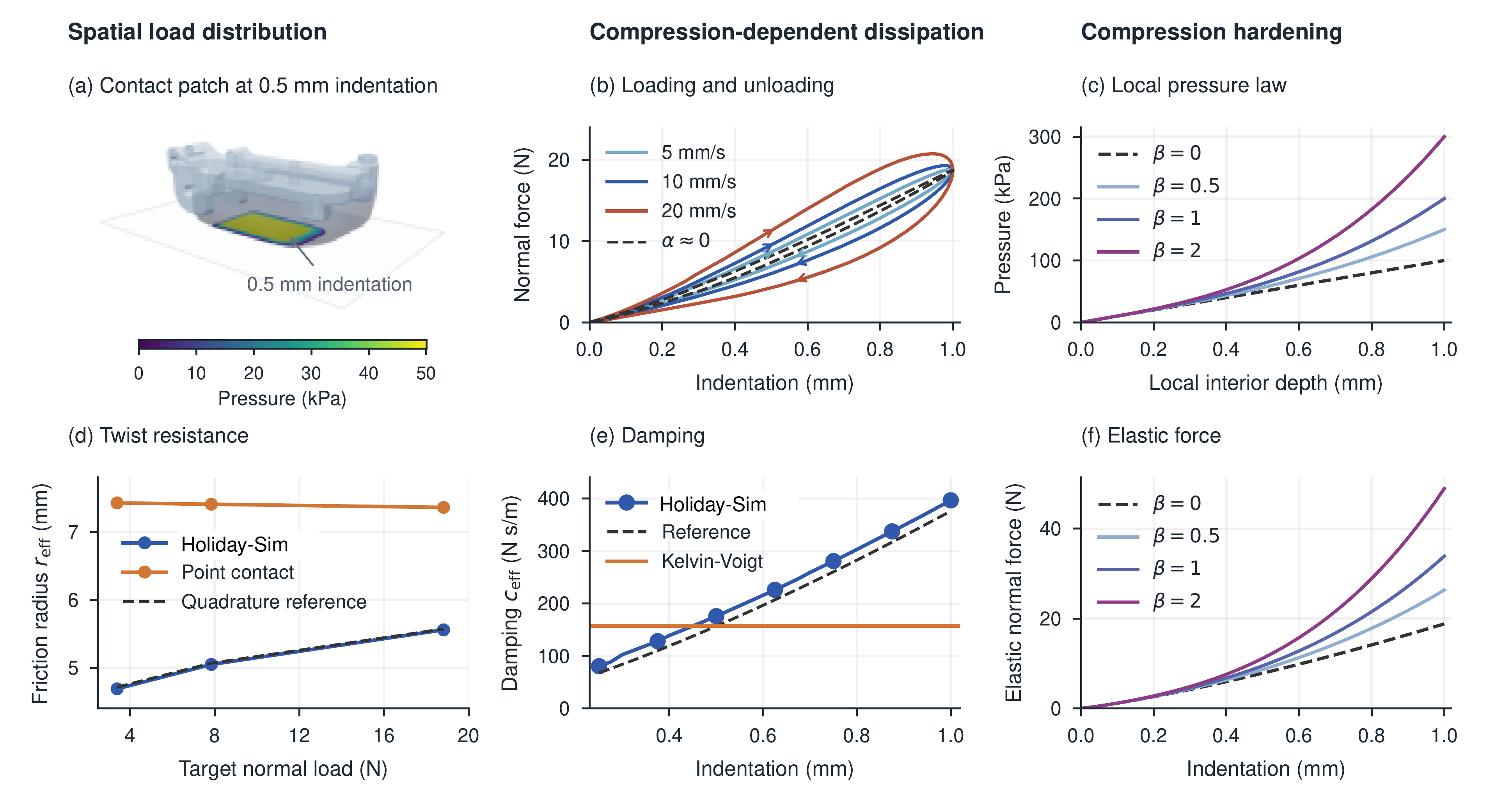}
\caption{\friday fingertip contact response. Left, (a) pressure patch at 0.5\,mm indentation and (d) effective friction radius under pure twist, for \hdaynewton and a point-contact configuration. Middle, (b) prescribed loading and unloading loops and (e) the effective damping they imply. Right, (c) depth-hardening pressure laws for several $\beta$ and (f) the elastic forces they integrate to. Dashed curves are quadrature references of the same law, and the first two columns use the linear law. 
}
\label{fig:fingertip-patch-friction}
\end{figure}

\begin{figure}[t]
\centering
\includegraphics[width=\linewidth]{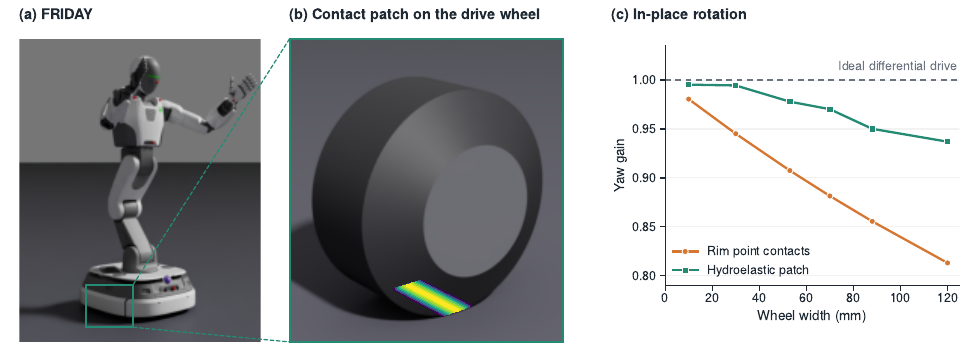}
\caption{Wheel-ground contact. (a) \friday with its right drive wheel marked behind the base skirt. (b) The hydroelastic pressure sampled on the authored wheel profile under a static load, drawn through the wheel and stretched along the rolling direction for visibility. Rim contacts place one point at each tread edge instead. (c) Yaw gain in the two-wheel width sweep.}
\label{fig:tire-ground-contact}
\end{figure}

\paragraph{Contact probes}
Controlled probes isolate three responses of the contact model from any benchmark task and compare each with a quadrature of the same constitutive law.
Under pure twist of the \friday index fingertip against a rigid plane, the effective friction radius $r_{\mathrm{eff}}=|\tau_n|/(\mu N)$ is the pressure-weighted distance from the rotation axis for normal load $N$ and resisting torque $\tau_n$ (\autoref{fig:fingertip-patch-friction}(a,d)).
A dense quadrature of the linear pressure law gives a radius that increases with indentation. \hdaynewton, using Newton's moment-matching reduction, follows this trend. The tested point-contact configuration gives a larger, nearly constant radius.
Under prescribed loading and unloading of the same fingertip, the computed damping grows with indentation as the Hunt-Crossley law predicts, unlike a constant damper (\autoref{fig:fingertip-patch-friction}(b,e)).
Direct quadrature shows that raising $\beta$ keeps the response to light contact unchanged and increases resistance at depth (\autoref{fig:fingertip-patch-friction}(c,f)). This stiffening of the local law is separate from the growth of the patch noted above.
On a simplified differential-drive rig, the planar hydroelastic path is compared with a rim-point baseline across wheel widths (\autoref{fig:tire-ground-contact}). The loss of yaw gain with width is the turn-slip resistance that tire models integrate over the footprint~\citep{pacejka2012tire} and that point-contact kinematics of skid-steered robots omit~\citep{mandow2007skidsteer}. Here it follows from the same patch contact as every other body, without a tire model or a hand-set torsional friction coefficient. Tire deformation, tread, and hysteresis are not modeled. These probes compare the model with a numerical reference and involve no hardware measurement. 

\subsubsection{Semi-implicit contact solver}
\label{sec:contact-solver}

The contact laws above specify the forces that arise from local deformation and motion. The solver must resolve these forces together with the motion of the contacting bodies.
Hydroelastic patches produce many coupled contacts on each body or articulated chain, with stiffnesses ranging from compliant pads to near-rigid supports.
Within a fixed iteration budget, numerical errors can cause resting parts to rattle, grasped objects to slip, or supported loads to sink.

\hdaynewton uses a semi-implicit discretization to limit the nonlinear coupling within each solve while retaining the dependence of contact forces on the unknown next-step motion.
Contact geometry and selected constitutive quantities are held fixed within each substep, while stiffness and damping enter the implicit force--motion relation.
This avoids solving simultaneously for evolving contact geometry, material coefficients, and body motion; geometry and coefficients are refreshed at the next substep.
It folds Hunt-Crossley damping into the implicit solve through a single contact time scale, so stiffness and damping act together on the unknown next-step velocity (\S~\ref{app:solver-discretization}).

Friction, including its speed weakening, is solved in the same step without adding nonlinearity to the iteration. The patch-dependent resistance to sliding and twisting (\S~\ref{sec:contact-modeling}) is resolved together with the next-step motion. The coefficient $\mu_i^k=\mu_i(\|\mathbf{v}_{t,i}^k\|)$ is evaluated from the local slip speed at the start of each substep and held fixed during the solve (\S~\ref{app:solver-discretization}).

The resulting system is solved by Constrained Dynamics via Method of Multipliers (CDMM), a primal-dual iteration that retains both motion and contact reactions as unknowns. The comparison by \citet{macklin2020primaldual} shows that large stiffness ratios can impede primal descent, whereas large mass ratios can impede dual descent. This motivates the variable choice but does not guarantee convergence of the \hdaynewton solver.
The algorithmic core of CDMM builds on prior work~\citep{lee2025variations} that separates system dynamics from local contact relations using auxiliary contact variables and augmented-Lagrangian multiplier updates. \hdaynewton specializes the local response to the hydroelastic stiffnesses and semi-implicit Hunt-Crossley law above.

Splitting motion from contact lets CDMM handle many stiff hydroelastic contacts without a dense contact solve.
The motion update accounts for how contact forces act through articulated dynamics, and the contact update enforces the local material and friction response. Repeated exchange between the two brings predicted motion and contact reactions into agreement, so coupling between contacts is retained.
Each articulation or free body is solved with an augmented inertia matrix whose factorization is reused across iterations while the geometry, implicit coefficients, and augmentation parameter remain fixed. Local contact responses are evaluated in parallel.

A patch is resampled at every step, so the same support can reappear as new contact samples even while the physical interaction persists.
\hdaynewton warm-starts such samples. It reuses previous impulses for matched contacts in the updated contact frame and estimates impulses for new or unmatched contacts from their compression, stiffness, damping, and approximate inertial response.
Together with the current generalized velocity and the contact slack variables reconstructed from it, these estimates initialize both the motion and reaction sides of the solve.
Substepping complements this initialization by refreshing contact geometry and material coefficients more frequently.
The same time interval can be divided into smaller substeps, each with fewer solver iterations.
\citet{macklin2019smallsteps} show in XPBD that this allocation can reduce constraint error and numerical damping.
In \hdaynewton, successive substeps warm-start from matched reactions.
Additional substeps nevertheless require repeated collision processing and factorization, so the balance between substeps and solver iterations depends on the scene.

The numerical changes to CDMM are guided by the inertia behind each contact, which determines both how its update should be scaled and how stiff a response the time step can resolve.
Contact preconditioning uses this inertia to improve convergence. Contacts on light parts and heavy articulated mechanisms can converge at different rates, so \hdaynewton scales each contact block by an approximate inertial metric that incorporates contact directions and connectivity, applying the same scalar to its normal and tangential coordinates so that the isotropic friction cone is not distorted. The scaling is applied consistently to the dynamics and local contact updates and leaves the material law unchanged.
Near-rigid regularization uses it to keep stiff contacts stable. Shallow contacts whose stiffness exceeds what their inertia and the time step can resolve are softened, while deeper compression recovers the nominal stiffness (\S~\ref{app:solver-discretization}).

\paragraph{Comparison with MuJoCo}

\begin{wraptable}{r}{0.22\linewidth}
\vspace{-20pt}
\centering
\footnotesize
\caption{Battery-lift success across four time steps, with five rollouts per step size.}
\vspace{-5pt}
\label{tab:battery-lift-summary}
\setlength{\tabcolsep}{4pt}
\begin{tabular}{@{}lcc@{}}
\toprule
Backend & Iter. cap & Lifted \\
\midrule
MuJoCo & 15  & 1/20 \\
MuJoCo & 100 & 0/20 \\
\rowcolor{holidaygreendark!15} CDMM   & 32  & 20/20 \\
\bottomrule
\end{tabular}
\vspace{-10pt}
\end{wraptable}

\begin{figure}[t]
\centering
\includegraphics[width=0.85\linewidth]{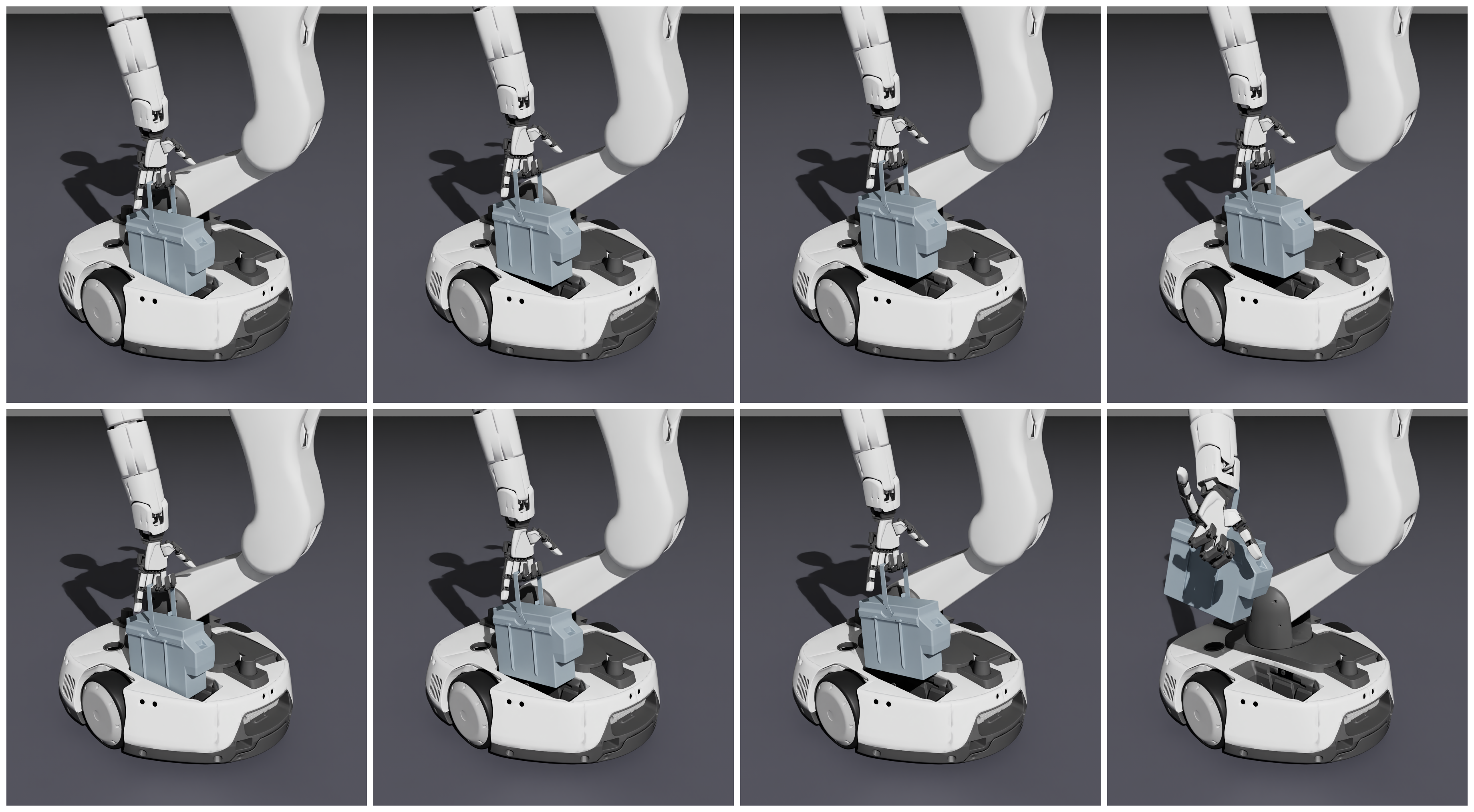}
\caption{Battery lifting with CDMM (top) and MuJoCo (bottom) at $h=1/2400$\,s using the same collision pipeline, initial state, and joint position targets.
Columns show $t = 0$, $1.150$, $1.183$, and $1.185$\,s, the last two being five substeps apart. CDMM completes the lift, whereas the MuJoCo state jumps abruptly and becomes nonfinite at $t=1.189$\,s.}
\label{fig:contact-solver-comparison}
\end{figure}

We compare CDMM with Newton's MuJoCo backend in a battery-lifting task involving compliant fingertips, lightweight finger links, and a substantially heavier rigid battery. 
Both backends share the SDF-hydroelastic collision pipeline, initial state, and joint targets, but retain different contact treatments. 
CDMM completes more lifts under the tested configurations (\autoref{tab:battery-lift-summary}). 
However, the grasp and targets were tuned using CDMM, so the comparison does not isolate solver performance. \autoref{fig:contact-solver-comparison} illustrates an additional rollout.

\subsubsection{Magnetic tactile sensing}
\label{sec:tactile}

Tactile sensing captures contact onset and load changes that may be obscured from vision. 
The \friday hand uses compliant skin with embedded magnets whose contact-induced displacement changes the magnetic field measured by Hall sensors~\citep{bhirangi2022reskin,tomo2018uskin}.
Existing simulators model skin deformation using finite-element methods~\citep{si2024difftactile} or approximate contact-response models~\citep{si2021taxim,akinola2024tacsl,dang2026hydroshear}.
\hdaynewton instead uses hydroelastic contact patches (\S~\ref{sec:contact-modeling}) to generate magnetic tactile responses, sharing the contact representation between dynamics and sensing.

The patches give the sensor its load distribution without a second contact computation.
The same net load applied over a broad face, a narrow ridge, or two separate patches deforms the skin differently, and the patches already resolve these differences down to the SDF and patch resolution.
The sensor reconstructs elastic pressure from each contact's depth and stiffness with the linear law of Eq.~\eqref{eq:sdf-pressure-field} and weights it by contact area to obtain the distributed normal load.

Following HydroShear~\citep{dang2026hydroshear}, the sensor turns these loads into magnet displacements by superposing spatially weighted contributions and does not solve volumetric elasticity.
With reconstructed pressure $p_i$ and area $A_i$ at contact position $\mathbf{x}_i$, the displacement of magnet $m$ is
\begin{equation}
\mathbf{u}^{\mathrm{press}}_m
\approx \sum_i \mathbf{G}_m(\mathbf{x}_i,\mathbf{n}_i)\,p_iA_i
\approx \int_{\mathcal{S}}\mathbf{G}_m(\mathbf{x},\mathbf{n})\,p(\mathbf{x})\,\mathrm{d}A ,
\label{eq:tactile-pressure-displacement}
\end{equation}
where $\mathbf{G}_m$, in $\mathrm{m}/\mathrm{N}$, is the magnet's displacement per unit normal load at $\mathbf{x}_i$ along the inward normal $\mathbf{n}_i$, representing inward compression and radial deformation around the loaded region.

Sliding contact also shears the skin, and \hdaynewton stores the shear history on the sensor.
Each query point on the sensor surface records its contact position at onset as a reference attached to the object. The tangential displacement of that reference adds to magnet displacement and resets when the query loses contact.
Unlike HydroShear, which tracks predefined object-surface points with Coulomb-limited history~\citep{dang2026hydroshear}, this sensor-side history needs no tactile samples prepared for each new object (\S~\ref{par:tactile-shear-history}).

Hall readings follow from the displaced magnets through the point-dipole field:
\begin{equation}
\mathbf{B}(\mathbf{r})=\frac{\mu_0}{4\pi r^3}\left[3(\mathbf{m}\cdot\hat{\mathbf{r}})\hat{\mathbf{r}}-\mathbf{m}\right],
\label{eq:tactile-dipole}
\end{equation}
where $\mathbf{B}$ is the flux density from one magnet with moment $\mathbf{m}$, held fixed in the sensor frame, $\mathbf{r}$ points from the displaced magnet to the Hall sensor, $r=\|\mathbf{r}\|$, $\hat{\mathbf{r}}=\mathbf{r}/r$, and $\mu_0$ is the vacuum permeability.
Each Hall sensor sums these contributions in its own frame.

The coupling is one-way. The reconstructed deformation does not feed back into collision geometry or the dynamics, and the sensor reads reconstructed elastic loading rather than the solver's final reactions (\S~\ref{par:tactile-coupling}).

\begin{figure}[t]
\centering
\includegraphics[width=\linewidth]{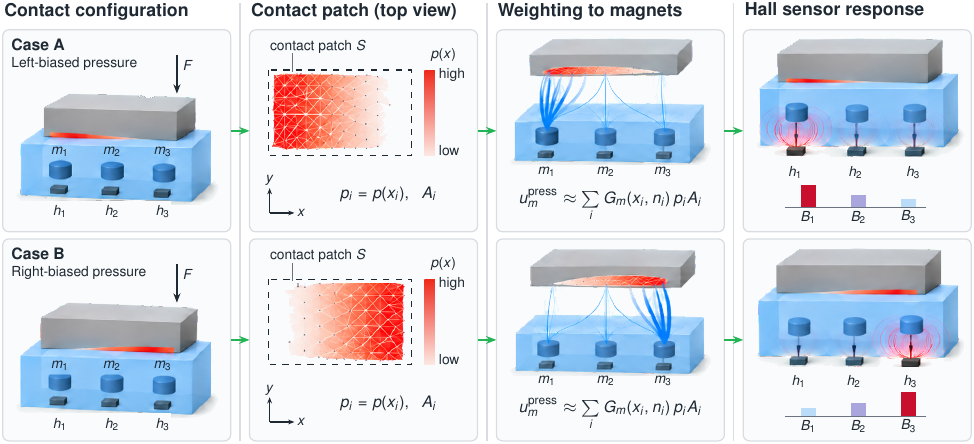}
\caption{Two contacts with the same net load but different patch distributions on the \friday tactile skin. The reconstructed pressure displaces the embedded magnets differently and produces different Hall readings.}
\label{fig:tactile-pressure-displacement}
\end{figure}

\autoref{fig:tactile-pressure-displacement} shows two contacts of equal net load whose different patch distributions give different Hall readings, which a single net-force sensor model cannot reproduce.

\begin{figure}[t]
\centering
\includegraphics[width=\linewidth]{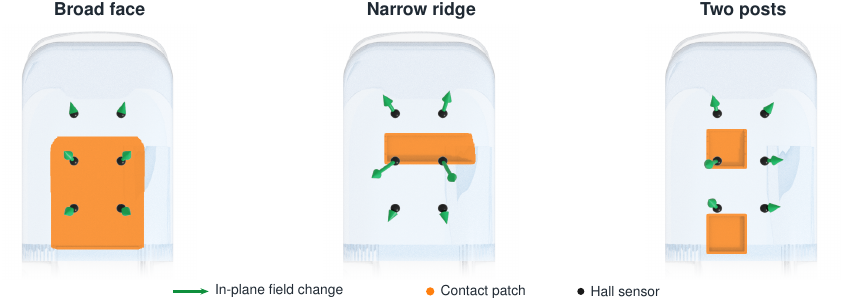}
\caption{Simulated Hall responses of a \friday fingertip pad to three rigid probes at the same net normal load of 1\,N, viewed from the contact side with the fingertip up. 
The orange region is the hydroelastic contact patch, black dots mark the Hall sensors, and arrows show the in-plane field change on a common scale. 
The out-of-plane component is not drawn.}
\label{fig:tactile-probes}
\end{figure}

\paragraph{Patch-dependent readings}
We pressed three rigid probes against a \friday fingertip pad until each reached the same net normal load of 1\,N (\autoref{fig:tactile-probes}).
The Hall readings differ with the shape and position of the probe, whereas a model driven by net force alone would return one reading for all three.
These are model outputs with default kernel parameters, so the figure shows what the sensor model distinguishes rather than how closely it matches hardware. 

\subsection{Stateful actuator}
\label{sec:actuator}

The motor, its drive, and its transmission decide which forces the robot can deliver and how quickly it responds.
The benchmarks in \autoref{tab:benchmark-actuation} use joint-level torque commands or PD drives without internal actuator states, which capture the commanded behavior but omit the dynamics that govern tracking, sustained force, and motion near rest.
\hdaynewton carries the actuation chain from a commanded target to the delivered joint effort with states advanced at every step. These are the shaped command and error integral of the drive, the winding current and temperature of the motor, the chamber effort of a single-acting transmission, and the bristle deflection of the internal friction (\S~\ref{sec:actuator-modeling}).
The articulated solve receives these states through their implicit tangents (\S~\ref{sec:actuator-solver}), and the joint telemetry a policy observes is reported from the same states (\S~\ref{sec:telemetry}).

\begin{table}[t]
\centering\small
\caption{Actuation models in representative robot simulation benchmarks.}
\label{tab:benchmark-actuation}
\begin{tabularx}{\linewidth}{@{}>{\raggedright\arraybackslash}p{0.19\linewidth}YYY@{}}
\toprule
\tablehead{System}
& \tablehead{Actuation model}
& \tablehead{Actuation characteristics}
& \tablehead{Internal actuator states}\\
\midrule
LIBERO
& Ideal torque/servo actuation
& Torque limits
& None modeled\\

RoboCasa365
& Ideal torque/servo actuation
& Torque limits
& None modeled\\

ManiSkill3
& Ideal PD drives
& Torque limits
& None modeled\\
\midrule
\rowcolor{holidaygreendark!15} \hdayfactory
& Embedded control $\rightarrow$ motor/transmission
& Electrical limits; transmission and friction dynamics
& Electrical, thermal, friction, hydraulic, and controller states\\
\bottomrule
\end{tabularx}
\smallskip
\begin{minipage}{\linewidth}
\footnotesize
``Ideal'' denotes actuation without separate drive dynamics, and its output is still limited.
\end{minipage}
\end{table}

\subsubsection{Drive and transmission models}
\label{sec:actuator-modeling}

\hdaynewton models command latency, first-order target filtering, and position slew limits, using either motor-position PID with current or voltage commands or cascaded joint-position and velocity control. 
The drive includes gravity and friction compensation, current feedforward, and saturation-aware anti-windup.

Motor current and electromagnetic torque follow
\begin{equation}
L\dot{i}=V-R(T)i-k_e\omega_m,
\qquad \tau_{\mathrm{em}}=k_t i,
\label{eq:actuator-electrical}
\end{equation}
where $L$ is winding inductance, $R(T)$ temperature-dependent resistance, $i$ current, $V$ terminal voltage, $\omega_m$ motor speed, and $k_e$ and $k_t$ the back-EMF and torque constants.
Current limits and the terminal-voltage range, bounded by bus voltage and maximum duty cycle, constrain the output. The inductance delays the current buildup, and torque falls once back-EMF consumes the available voltage.
A lumped thermal state heats the winding through copper loss $i^2R(T)$ and cools it toward ambient. The resulting increase in winding resistance can reduce available torque when voltage-limited.
Current, terminal power, copper loss, and temperature are available as diagnostics (\S~\ref{par:actuator-electrical}).

A motor's effect on joint motion depends on its transmission, so the model includes geometric transmissions as well as constant gearing.
With an offset slider-crank, the same motor produces different joint torque and reflected inertia across the joint range. Writing motor position as a function of joint position gives
\begin{equation}
\theta_m=\phi(q),\qquad r(q)=\frac{\mathrm{d}\phi}{\mathrm{d}q},\qquad
\omega_m=r(q)\dot{q},\qquad I_{\mathrm{ref}}(q)=I_r r(q)^2,
\label{eq:actuator-transmission}
\end{equation}
where $I_r$ is rotor inertia and the reflected-inertia relation applies while the motor is coupled.
Motor-side effort is mapped through the local transmission ratio $r(q)$ and a scalar efficiency, and the dynamics include the inertial bias $I_r r(q)r'(q)\dot{q}^{2}$, with $r'(q)=\mathrm{d}r/\mathrm{d}q$, that the changing reflected inertia produces (\S~\ref{par:actuator-transmission}).

Small commands and motion reversals expose internal friction that viscous damping cannot represent, so the model adds motor-side LuGre friction~\citep{canudas1995friction}.
Unlike the memoryless contact-friction coefficient of \S~\ref{sec:contact-modeling}, LuGre friction carries a bristle state that represents presliding deformation and history-dependent resistance, and a Stribeck envelope governs the transition toward sliding. Together these reproduce small-load holding, breakaway, and reversal.
Optional temporally correlated variation of the LuGre parameters, a low-speed torque disturbance, and motor-angle-dependent cogging add the variability seen in precision motion and position holding (\S~\ref{par:actuator-variability}).

The \friday fingers use single-acting hydraulic drives. 
A motor drives a piston through a lead screw, hydraulic pressure closes the finger, and a return spring opens it.
The drive can exert force only in the closing direction.
\hdaynewton models this response with a nonnegative chamber-effort state that builds and decays with first-order dynamics, and flow damping that resists motion in both directions (\S~\ref{par:actuator-single-acting}).
The return spring and damper act through the transmission independently of electrical saturation.
Without chamber dynamics, a coupling switch based on the sign of the requested torque approximates the unilateral drive.

\begin{figure}[t]
\centering
\includegraphics[width=\linewidth]{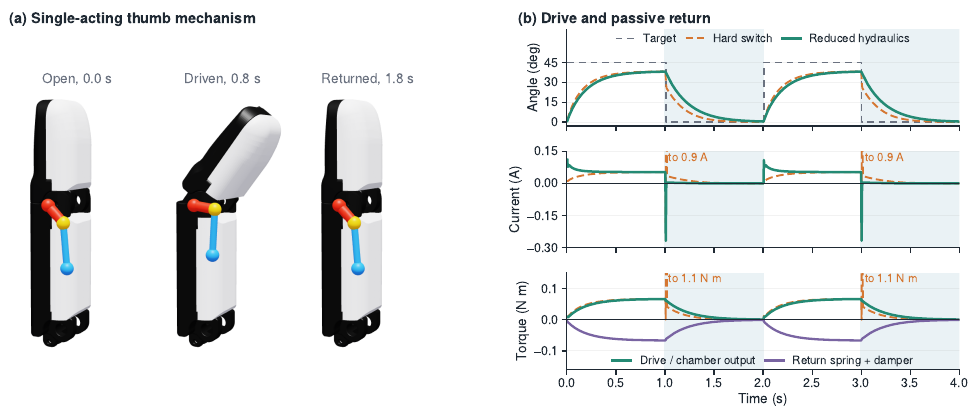}
\caption{Single-acting drive and passive return in the \friday thumb example. 
(a) Snapshots from one cycle of the single-acting hydraulic finger. The colored linkage is a kinematic visualization. (b) Joint motion, motor current, and joint-side torques over two cycles. 
Shaded intervals command release. 
Solid curves use the reduced hydraulic model with the hand's parameters, and dashed curves replace it with a coupling switch on the sign of the requested torque, which drops the motor's damping for one step at release and re-engages the motor at its current limit.
}
\label{fig:single-acting-response}
\end{figure}

In the \friday thumb example, the motor drives the joint toward its target, while commanded release allows chamber effort to decay as the spring reopens the joint against flow resistance (\autoref{fig:single-acting-response}).
The drive and return torques capture the asymmetric response absent from a bidirectional servo model.
In contrast, the coupling-switch approximation drops motor damping for one step at release.

\subsubsection{State integration and dynamics coupling}
\label{sec:actuator-solver}

Actuators can have high stiffness and internal dynamics that evolve much faster than the robot's motion. Applying their effects as explicit joint forces can therefore make the result strongly dependent on the step size.
\hdaynewton updates internal states locally and incorporates the dominant inertial, stiffness, and damping effects into the articulated solve through a semi-implicit contribution. This couples the actuator response to the robot's motion without a fully nonlinear electromechanical solve.

Contact and actuators are solved together, so the actuator's impedance contributes to the contact response.
Reflected rotor inertia and the controller and passive terms enter the articulated dynamics used by the contact iteration, schematically as
\begin{equation}
\mathbf M_{\mathrm{aug}}=\mathbf M+\operatorname{diag}\!\left(I_{\mathrm{ref}}+hD_{\mathrm{act}}+h^2K_{\mathrm{act}}\right).
\label{eq:actuator-implicit-matrix}
\end{equation}
Here, $\mathbf M$ includes the joint armature, $h$ is the substep duration, and $K_{\mathrm{act}}$ and $D_{\mathrm{act}}$ collect local implicit coefficients expressed in joint coordinates, arising from controller response, passive springs and damping, and a nonnegative approximation to the LuGre stiffness. Corresponding force corrections accompany these terms.

The controller's contribution to the implicit response depends on the control mode, so the same numerical gain need not produce the same joint impedance.
Voltage-controller gains are converted to motor effort through $k_t/R(T)$ and scaled by the current that builds within the substep. Back-EMF adds implicit damping where the limits leave it active, while cascade gains act in joint coordinates.

Implicit stiffness and damping terms are retained when the output is clipped, providing regularization as a drive saturates or a single-acting drive switches to passive return.
For the controller, this applies at a current or voltage limit. Similarly, hydraulic flow damping is retained across the zero-output boundary, because a return spring or external load can reverse motion within a substep.

The LuGre term enters the solve through a state- and direction-dependent stiffness approximation. After the solve, the bristle state advances with the solved joint velocity, which removes the artificial creep of advancing it with the pre-solve velocity. Electrical and thermal states use exponential updates with their driving quantities held fixed over the substep. 

\subsubsection{Joint telemetry}
\label{sec:telemetry}

\hdaynewton generates joint telemetry from actuator and mechanism states, with configurable sampling and measurement errors.
The reported channels include joint position, estimated velocity, motor current, and joint effort.
Encoder resolution, filtering, calibration error, and correlated noise can obscure small motions, delay velocity estimates, and cause reported effort to differ from delivered joint torque.

Motor position is quantized at the encoder resolution and mapped to joint position through the local transmission ratio, with an optional joint-position bias. Velocity is estimated from successive samples over the actual elapsed interval and optionally low-pass filtered. Outputs are held between samples, so a faster physics rate does not raise the telemetry rate.

Effort can be reported from motor current, as on hardware.
With measured current $\hat{i}$, torque constant $k_t$, transmission ratio $r(q)$ from Eq.~\eqref{eq:actuator-transmission}, and efficiency $\eta$, the reported effort is
\begin{equation}
  \hat{\tau}_{\mathrm{current}}
  = a_{\tau}\,\eta\,r(q)k_t\hat{i}+b_{\tau}+\epsilon_{\tau},
  \label{eq:telemetry-current-effort}
\end{equation}
where $a_{\tau}$ and $b_{\tau}$ are calibration scale and bias, and $\epsilon_{\tau}$ is optional correlated error.
Because internal friction and passive mechanisms sit between current and output, this estimate differs from the net joint effort. An alternative channel reports the computed net effort before calibration and noise.

Observation errors stay separate from physical disturbances. An Ornstein-Uhlenbeck process adds correlated error to the effort channel without changing the forces on the robot.
These outputs reach a policy only when a run exposes them as policy inputs, and their parameters are recorded with the benchmark settings (\S~\ref{app:robot-simulation}).

\Needspace{5\baselineskip}
\section{Agentic world}
\label{sec:worlds}
A task-conditioned world $\xi=(S,E)$ needs assets that support the interactions of the task, scene variants with process logic, and an evaluation consistent with both, each verified before use.
\eris generates and verifies assets (\S~\ref{sec:eris}), inverse scene generation builds and verifies scene variants and adds backgrounds (\S~\ref{sec:world-generation}), and evaluation generation turns the task into a confirmed, compiled, and checked evaluation (\S~\ref{sec:eval-authoring}).

\subsection{Constraint-driven asset generation}
\label{sec:eris}
\begin{figure}[t]
\centering
\includegraphics[width=1.0\linewidth]{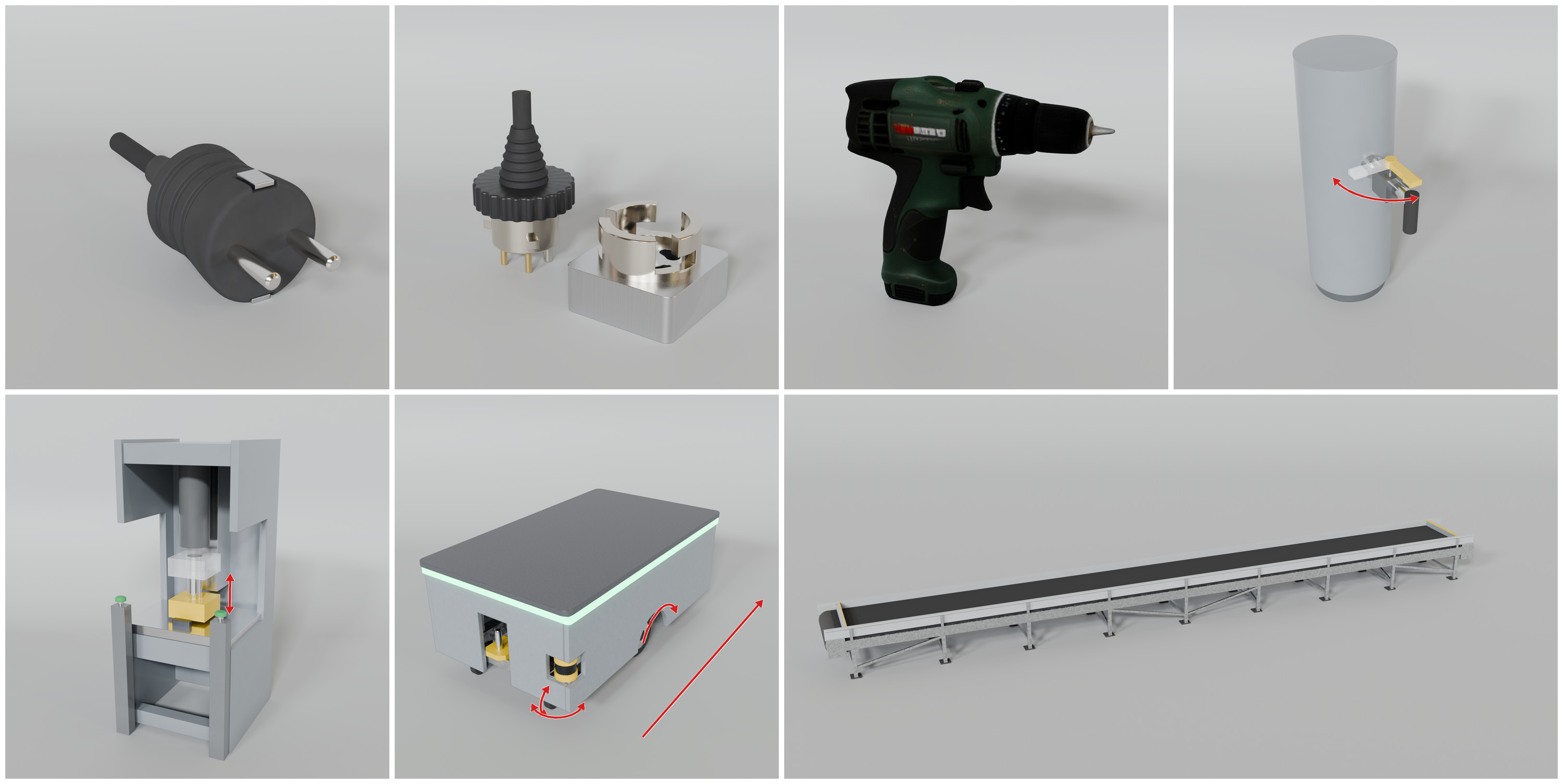}
\caption{Example assets constructed by \eris. Red arrows indicate articulated degrees of freedom.}
\label{fig:eris-gallery}
\end{figure}

We introduce \eris, a constraint-driven asset generation system that selects, constructs, verifies, and revises each asset according to the guarantees its downstream interaction requires, not according to its object category. \autoref{fig:eris-gallery} illustrates the range of rigid and articulated assets produced by the system.

Visual fidelity and diversity alone are insufficient for industrial manipulation.
Manipulation tasks involve tight fixtures, part-to-part contact, and articulated mechanisms, where small errors in dimensions or clearances decide whether an interaction is feasible.
A visually plausible fixture can still block insertion if its mating geometry is slightly off.
Collision geometry that is finer than needed or badly partitioned inflates memory when many simulation instances run in parallel.

\begin{figure}[t]
\centering
\includegraphics[width=1.0\linewidth]{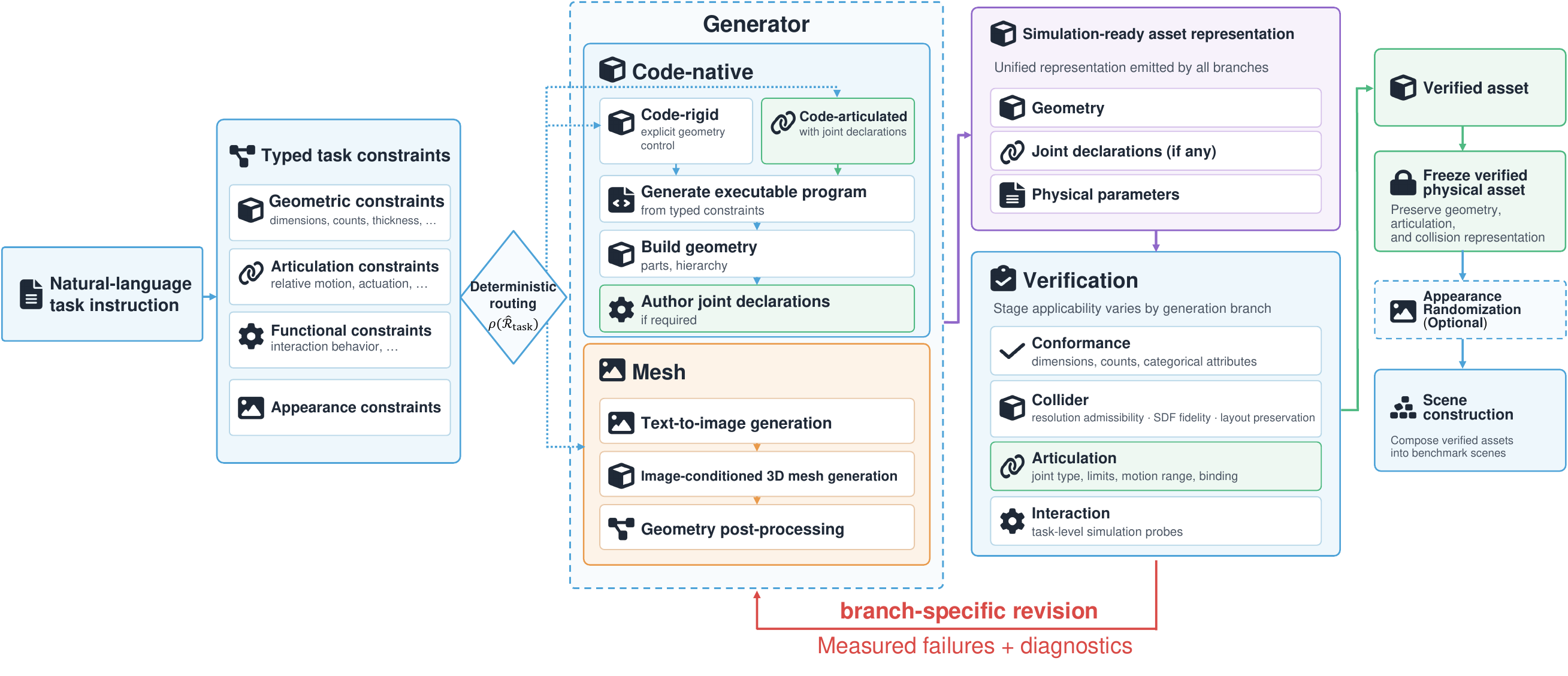}
\caption{
An overview of \eris.
A natural-language task instruction is first converted into a structured set of typed task constraints, from which \eris deterministically selects one of three generation branches: mesh, code-rigid, or code-articulated.
All three branches emit a common simulation-ready format and undergo the applicable verification stages.
Measured failures and diagnostics are returned through a branch-specific revision path.
Green marks the stages that apply only to articulated assets and the verified outputs.
}
\label{fig:eris-overview}
\end{figure}

Assets from 3D repositories~\citep{chang2015shapenet,deitke2023objaverse}
and generative models provide broad semantic
and visual coverage, but are not verified against a particular task.
\eris verifies each asset against its task
on the representation the simulator consumes and revises it from the measured violations before it enters a scene (\autoref{fig:eris-overview}).

\subsubsection{Constraints and routing}
\label{sec:eris-routing}

Given a natural-language task instruction or asset request, \eris first infers a structured set of task constraints $\widehat{\mathcal{R}}_{\mathrm{task}}$ using an LLM.
The constraints are drawn from a fixed set of constraint types covering dimensions and counts, wall thickness, clearance and access, articulation, functional behavior, and appearance (\S~\ref{app:eris-constraints}).
Each type specifies the fields needed to instantiate it and, when applicable, its corresponding executable geometry check, articulation intent, or functional simulation claim.

\eris has two generators that differ in how geometry is produced. A mesh generator reconstructs a surface from a generated image, and a code-native generator writes an executable construction program. Only the latter can declare joints, and together they provide three branches: mesh, code-rigid, and code-articulated.
Given $\widehat{\mathcal{R}}_{\mathrm{task}}$, branch selection is deterministic:

\begin{equation}
    b = \rho(\widehat{\mathcal{R}}_{\mathrm{task}}),
    \qquad
    b\in\{
        \textsc{Mesh},
        \textsc{Code-Rigid},
        \textsc{Code-Articulated}
    \},
    \label{eq:eris-route}
\end{equation}

where $\rho$ selects a generation branch from typed task constraints.
Requests prioritizing shape or appearance diversity without explicit geometric or interaction constraints route to mesh.
Requests specifying dimensions, counts, thicknesses, or clearances route to code-rigid.
Requests requiring relative motion or articulation route to code-articulated, regardless of other constraints.

Routing depends on task requirements rather than object category.
For example, the same container can route to mesh for visual diversity, code-rigid for prescribed fixture clearances, or code-articulated when movable parts are required.

\subsubsection{Mesh and code-native generation}
\label{sec:eris-generation}

The two generators trade visual detail for explicit control.
The mesh generator chains a text-to-image model, an image-conditioned 3D mesh generator~\citep{chen2026sam}, and post-processing~\citep{oh2025pamo}. It inherits the shape detail of generative models but exposes no construction parameters.
The code-native generator, shared by both code branches, writes an executable Blender program that defines geometry, named parts, and their hierarchy, so dimensions, counts, and clearances are set explicitly.
In the code-articulated branch, the program also declares joints over the part hierarchy with their type, axis, limits, and drive parameters.
All three branches ultimately produce assets with geometry, physical parameters, and verified collision representations, together with joint declarations where articulation is required.
The mesh generator keeps the texture of the generative model, whereas the code-native generator assigns part-aware materials during generation and a canonical PBR appearance after the asset passes verification.

\subsubsection{Verification and revision}
\label{sec:eris-verification}

\eris places verification inside the generation loop. Each round of the loop runs up to four verification stages on the asset in \autoref{tab:eris-stages}, and feeds every failure back to the generator before the next round.
The stages execute checks derived from the applicable task
constraints and the checks that \hdaynewton always requires,
such as construction validity, hierarchy, and collision representation.
Which stages apply depends on the branch and on the capabilities
the asset declares (\autoref{tab:eris-branch-stages}).

Within a round, \eris runs every stage the asset can reach rather than stopping at the first violation, so that all violations are revised together.
The generator receives each violation with its measured value, and the checks that already pass so that it does not break them.
Generation succeeds once a round passes every check,
and otherwise \eris reports the remaining violations after $K$ rounds.

\begin{table}[t]
    \centering
    \begin{minipage}[t]{0.60\linewidth}
        \vspace{0pt}
        \centering
        \caption{
            Verification stages in \eris with the checks most
            directly tied to task feasibility and downstream simulation.
        }
        \label{tab:eris-stages}
        \scriptsize
        \setlength{\tabcolsep}{3pt}
        \begin{tabularx}{\linewidth}{
            @{}>{\raggedright\arraybackslash}p{0.14\linewidth}YY@{}
        }
            \toprule
            Stage & Representative checks & Example failure \\
            \midrule
            Conformance
                & categorical attributes, dimensions, counts
                & rack has 3 shelves where 4 are requested \\
            Collider
                & resolution admissibility, SDF fidelity, layout preservation
                & thin tote wall vanishes after SDF baking \\
            Articulation
                & joint type, limits, motion range, binding
                & drawer binds before reaching its limit \\
            Interaction
                & insertion, rolling, transport, retention
                & peg jams on insertion \\
            \bottomrule
        \end{tabularx}
    \end{minipage}
    \hfill
    \begin{minipage}[t]{0.38\linewidth}
        \vspace{0pt}
        \centering
        \caption{
            Verification stages by generation branch.
            \checkmark denotes a standard stage, ``task''
            a task-dependent stage, and -- a non-applicable stage.
        }
        \label{tab:eris-branch-stages}
        \scriptsize
        \setlength{\tabcolsep}{2pt}
        \begin{tabularx}{\linewidth}{
            @{}l*{4}{>{\centering\arraybackslash}X}@{}
        }
            \toprule
            Branch
                & Conf.
                & Coll.
                & Artic.
                & Inter. \\
            \midrule
            Mesh
                & \checkmark & \checkmark & -- & -- \\
            Code-rigid
                & \checkmark & \checkmark & -- & task \\
            Code-articulated
                & \checkmark & \checkmark & \checkmark & task \\
            \bottomrule
        \end{tabularx}
    \end{minipage}
\end{table}

\paragraph{Conformance}
Conformance checks whether the generated asset satisfies the properties requested by the task instruction.
For the mesh branch, a VLM gives a discrete value for each categorical property, such as the presence of handles and openings, from turntable renders of the asset. When a value differs from the request, \eris updates the asset's
label to match instead of rejecting it.

Code-native assets expose named parts and explicit geometric parameters, allowing conformance to be measured directly against requested dimensions, counts, clearances, and other structured constraints.
Each check reports the requested and the measured value,
and the code-native generator revises the program from them.

\paragraph{Collider}
Assets in \hdayfactory are consumed by \hdaynewton, whose contact model represents colliders using sparse-grid SDFs~(\S~\ref{sec:contact}).
Collider construction must preserve task-relevant geometry while controlling collision memory.
Coarse SDFs can erase thin structures, introduce perforations, or fragment connected geometry, compromising downstream interactions.
Excessively fine SDFs increase field storage and collision-buffer requirements, particularly across parallel simulation instances.

Collider construction seeks to minimize predicted collision memory while preserving task-relevant geometry.
Let $\Pi$ partition an asset's parts into colliders, $\boldsymbol\Delta_{\mathrm{vox}}$ denote their SDF voxel sizes, and $M_{\mathrm{SDF}}(\Pi,\boldsymbol\Delta_{\mathrm{vox}})$ estimate collision memory.
We formulate the objective as
\begin{equation}
    \min_{(\Pi,\boldsymbol\Delta_{\mathrm{vox}})\in\mathcal{F}_{\mathrm{SDF}}}
    M_{\mathrm{SDF}}(\Pi,\boldsymbol\Delta_{\mathrm{vox}}).
    \label{eq:eris-collision-objective}
\end{equation}
where $\mathcal{F}_{\mathrm{SDF}}$ contains configurations that pass the collider checks (\autoref{tab:eris-collision-checks}).

\eris searches for feasible, low-memory representations through iterative validation.
For a fixed collider layout, geometric constraints bound the admissible voxel sizes.
\eris selects a resolution for each collider, bakes its SDF in \hdaynewton, and checks the resulting collision geometry.
Failed validation triggers a more conservative parameterization and rebaking until the checks pass or the admissible range is exhausted.
For multipart assets, \eris also searches for lower-cost groupings of eligible static parts, rejecting merges that fail per-part geometry checks.
Only when these searches fail does \eris return the measured violations for source-level revision.

\begin{table}[t]
\centering
\caption{
Checks of the collider verification stage.
}
\label{tab:eris-collision-checks}
\small
\begin{tabular}{lll}
\toprule
Check & Measurement & Action on failure \\
\midrule
Resolution admissibility
& geometric voxel-size bounds
& choose a finer admissible voxel size \\

SDF fidelity
& surface, boundary, and topology checks
& refine and rebake \\

Layout preservation
& per-part wall and surface retention
& keep failed part separate \\
\bottomrule
\end{tabular}
\end{table}

The three checks enter collider construction at different points (\autoref{tab:eris-collision-checks}).
Resolution admissibility is a pre-bake geometric check that constrains the initial SDF parameters using analytic bounds derived from collider scale, measured wall thickness, and declared functional features.
SDF fidelity is a post-bake quantitative check that evaluates surface retention, boundary validity, and topology changes in the resulting \hdaynewton collider.
Layout preservation is a merge-specific geometric check that measures per-part wall and surface retention after multiple source parts have been grouped into a shared collider.
We provide further details on this stage in
\S~\ref{app:eris-sdf}.

\paragraph{Articulation}
\eris uses the articulation intent derived from task constraints to guide the generation of joint declarations.
\eris verifies that the resulting joint declarations are valid and dynamically realizable.
It checks the joint type, axis, pivot, and limits, and tests the corresponding revolute or prismatic motion for geometric binding.
The articulated asset is then exercised in \hdaynewton.
Driven joints run under their declared drives, and passive joints receive a controlled test input.
\eris verifies that each joint achieves a prescribed fraction of its expected motion without exceeding joint limits or binding.

\paragraph{Interaction}
\eris evaluates task-relevant physical interactions using the baked SDF colliders in \hdaynewton.
Geometric clearance in the source meshes, for example, does not guarantee successful insertion after SDF conversion.
Each simulation probe specifies the interacting objects, their initial configuration, and applied inputs.
Generic probes test load support, transport, path following, and payload retention, while dedicated probes test interactions such as insertion, seating, and rolling.
For insertion, a part is placed above its mating fixture and allowed to settle under gravity, after which its final pose is checked against prescribed seating tolerances.
A constraint passes only when the measured outcome satisfies its acceptance criterion.

\paragraph{Branch-specific revision}
\eris revises failed assets through branch-specific operations guided by verification feedback.
Let $X_t$ denote the editable asset state at round $t$ for generation branch $b$, including geometry and, where applicable, joint declarations and physical parameters.
The revision step is
\begin{equation}
    X_{t+1}
    =
    \mathcal{F}_{b}
    \left(
        X_t,
        \widehat{\mathcal{R}}_{\mathrm{task}},
        \mathcal{C}^{-}_t,
        \mathcal{C}^{+}_t
    \right),
    \label{eq:eris-repair}
\end{equation}
where $\mathcal{F}_b$ is the branch-specific revision function, $\mathcal{C}^{-}_t$ contains measured violations and diagnostics, and $\mathcal{C}^{+}_t$ records satisfied constraints to help prevent regressions.

For code-native assets, \eris edits the executable construction program, including joint declarations and physical parameters where applicable.
For mesh assets, correctable collider defects, such as thin walls, are repaired through surface offsetting, remeshing, and SDF rebaking.
Conformance failures originating from image generation or 3D reconstruction, and collider failures without reliable local corrections, instead trigger regeneration.
\subsubsection{Appearance randomization}
Before finalizing an asset, \eris can optionally apply appearance randomization to produce multiple visual variants while preserving the verified geometry, articulation, and collision state.
The frozen geometry is UV-parameterized and packed into a shared atlas~\citep{young2018xatlas}, after which a texture-generation model~\citep{bao2026lumitex} generates multiple texture variants on the same asset.

\subsubsection{Experiments}
\label{sec:eris-experiments}

We conduct five controlled studies: branch selection, verification, revision feedback, LLM backbones, and generation efficiency.
Unless otherwise noted, the code-native generator uses claude-opus-5-5 as its LLM backbone at high reasoning effort, and \eris runs for at most $K=4$ rounds (\S~\ref{app:eris-config}).
The backbone study additionally evaluates claude-fable-5-1, gpt-6-astra, gpt-6.1-sol, and gpt-6-luna.
Each comparison varies one factor while holding other applicable settings fixed.

An asset succeeds by round $t$ when all applicable executable checks have passed; required checks that cannot be evaluated count as failures.
We report $\mathrm{Success}@t$ as the fraction of evaluation units passing verification by round $t$.
For revision experiments, $\mathrm{Recovery}@t$ measures the fraction of units unsuccessful after round $t$ that pass in round $t+1$.
Paired comparisons use matched evaluation units.
Generation efficiency is measured by generation rounds and estimated API-equivalent model cost per solved asset and per recovered first-round failure.
The verification study reports stage-wise rejection rates, unique rejections, and the frequency of post-bake collider re-authoring.

\paragraph{Branch selection}
We compare mesh and code-native generation across visual, geometric, and functional requirements matched to their intended capabilities (\autoref{tab:eris-routing}).

For visual fidelity, each generator produces five assets from each of six categories with complex, non-primitive geometry, yielding 30 assets per generator.
Both receive only the object category and its longest dimension.
A blinded VLM judge (claude-fable-5-1) evaluates four-view renders using four labels: \emph{detailed and realistic}, \emph{simplified}, \emph{distorted}, and \emph{cannot tell}.
Each asset is judged twice with reshuffled label order, yielding 60 judgments per generator.
Separate tote and rack probes assess dimensions, part structure, mating geometry, and functional seating.

On these complex everyday objects, the mesh generator is rated \emph{detailed and realistic} in 80\% of judgments, compared with 0\% for code-native generation, although objects from both generators remain recognizable.
Conversely, code-native generation satisfies all tested dimensional, structural, and mating-geometry requirements, including functional seating.
Across three end-to-end routing trials per request, \eris consistently selects the generator matched to the specified requirements.

\begin{table}[t]
\centering
\caption{
Branch selection under different downstream requirements.
Mesh and Code-native entries report success rates over the evaluation unit shown in the second column.
Selected generator denotes the generator to which \eris routes the corresponding request.
}
\label{tab:eris-routing}
\small
\begin{tabular}{llccc}
\toprule
Probe
& Evaluation unit
& Mesh
& Code-native
& Selected generator \\
\midrule

Visual shape fidelity
& VLM judgments
& \textbf{80\%}
& 0\%
& Mesh \\

Tote: vertical walls, no handles
& geometric checks
& 80\%
& \textbf{100\%}
& Code \\

Tote: wall $\geq 8$\,mm
& geometric checks
& 75\%
& \textbf{100\%}
& Code \\

Rack / peg: requested pocket count
& structural checks
& 37.5\%
& \textbf{100\%}
& Code \\

Rack / peg: diameter within $\pm3$\,mm
& geometric checks
& 0\%
& \textbf{100\%}
& Code \\

Rack / peg: all parts fully seated
& functional checks
& 0\%
& \textbf{100\%}
& Code \\

\bottomrule
\end{tabular}
\end{table}

\begin{table}[t]
\centering

\begin{minipage}[t]{0.5\linewidth}
\centering
\caption{First-round verification failures among 72 assets.
Unique rejections indicate failures at only one stage.}
\label{tab:eris-verification}
\footnotesize
\setlength{\tabcolsep}{3pt}
\renewcommand{\arraystretch}{1.15}
\begin{tabularx}{\linewidth}{@{}Xcc@{}}
\toprule
Verification stage & Rejected & Unique \\
\midrule
Any stage    & 34/72 (47\%) & -- \\
Conformance  & 28/72 (39\%) & 24/72 (33\%) \\
Collider     & 10/72 (14\%) & 4/72 (6\%) \\
Articulation & 2/72 (3\%)   & 0/72 \\
Interaction  & 2/72 (3\%)   & 0/72 \\
\bottomrule
\end{tabularx}
\end{minipage}
\hfill
\begin{minipage}[t]{0.48\linewidth}
\centering
\caption{Post-bake SDF collider re-authoring.
Re-authoring does not imply asset rejection.}
\label{tab:eris-sdf-refinement}
\footnotesize
\setlength{\tabcolsep}{3pt}
\renewcommand{\arraystretch}{1.15}
\begin{tabularx}{\linewidth}{@{}Xr@{}}
\toprule
Simulator-side intervention & Incidence \\
\midrule
Colliders requiring re-authoring
& 346/1597 (22\%) \\
Assets with re-authored colliders
& 39/72 (54\%) \\
\bottomrule
\end{tabularx}
\end{minipage}

\end{table}

\begin{wraptable}{r}{0.45\textwidth}
\vspace{-10pt}
\centering
\caption{Revision feedback ablation.
All variants share the same first-round asset.}
\label{tab:eris-repair-ablation}
\scriptsize
\setlength{\tabcolsep}{2pt}
\renewcommand{\arraystretch}{1.15}
\begin{tabularx}{\linewidth}{@{}Xlll@{}}
\toprule
Metric
& \shortstack{Independent\\regeneration}
& \shortstack{Coarse\\feedback}
& \eris \\
\midrule
$\mathrm{Success}@1$
& 0.63 & 0.63 & 0.63 \\

$\mathrm{Recovery}@1$
& 0.33 (4/12)
& 0.17 (2/12)
& \textbf{0.92} (11/12) \\

$\mathrm{Success}@2$
& 0.75 & 0.69 & \textbf{0.97} \\

$\mathrm{Recovery}@2$
& 0.38 (3/8)
& 0.40 (4/10)
& \textbf{1.00} (1/1) \\

$\mathrm{Success}@3$
& 0.84 & 0.81 & \textbf{1.00} \\
\bottomrule
\end{tabularx}
\end{wraptable}

\paragraph{Verification}
We evaluate first-round verification on 72 assets, including 32 from the revision feedback study and 40 from the claude-opus-5-5 backbone study.
We examine failures across four stages: conformance, collider, articulation, and interaction.
We also measure SDF collider re-authoring following simulator-side validation.

Verification rejects 47\% of first-round assets, with conformance as the primary source of rejection (\autoref{tab:eris-verification}).
Separately, 22\% of SDF colliders require re-authoring after their initial bake (\autoref{tab:eris-sdf-refinement}).
These results distinguish verification that rejects an asset from simulator-side validation that revises its collision representation before acceptance.

\paragraph{Revision feedback}
We compare three revision strategies on 32 tasks with matched first-round assets.
Independent regeneration discards the failed asset and starts anew.
Coarse feedback retains the editable program but receives only a binary pass/fail signal.
\eris additionally uses measured constraint violations and diagnostics.

Of the 12 first-round failures, \eris repairs 11 after one revision and the remaining one after two, reaching 100\% final success (\autoref{tab:eris-repair-ablation}).
Independent regeneration reaches 84\% final success, while coarse feedback achieves 81\%.
Retaining the editable program without measured feedback therefore does not improve success over independent regeneration.

\begin{wraptable}{r}{0.4\textwidth}
\centering
\vspace{-15pt}
\caption{Revision across LLM backbones.
All values report $\mathrm{Success}@4$ with $K=4$ rounds;
$\Delta$ is the absolute improvement over independent regeneration.}
\label{tab:eris-backbone}
\scriptsize
\setlength{\tabcolsep}{3pt}
\renewcommand{\arraystretch}{1.12}
\begin{tabular}{@{}lccc@{}}
\toprule
LLM backbone & Raw & \eris & $\Delta$ \\
\midrule
claude-opus-5-5 (high)
& 0.65 (26/40) & \textbf{1.00 (40/40)} & +0.35 \\
claude-fable-5-1
& 0.70 (28/40) & 0.95 (38/40) & +0.25 \\
gpt-6-astra
& 0.65 (26/40) & 0.95 (38/40) & +0.30 \\
gpt-6.1-sol
& 0.58 (23/40) & 0.93 (37/40) & +0.35 \\
gpt-6-luna
& 0.43 (17/40) & 0.75 (30/40) & +0.33 \\
\bottomrule
\end{tabular}
\vspace{-10pt}
\end{wraptable}
\paragraph{LLM backbones}
To test whether the benefit of measurement-grounded revision generalizes across different LLM backbones, we compare \eris with independent regeneration under the same verifier and round budget for each backbone (\autoref{tab:eris-backbone}).
We evaluate 20 tasks with two seeds each, for 40 paired units per backbone.
Each pair shares the same first-round asset; after a failure, the raw baseline
generates a fresh independent sample from the original request, whereas \eris
revises the existing asset using measured verification feedback.
Across the evaluated LLM backbones, \eris consistently improves $\mathrm{Success}@4$ by 25--35 percentage points over independent regeneration.
This consistent gain despite substantial differences in baseline success rates suggests that the benefit of measurement-grounded revision is not specific to a particular LLM backbone.

\begin{wraptable}{r}{0.4\textwidth}
\vspace{-12pt}
\centering
\caption{Generation efficiency.
Overall measures are normalized by solved assets, and repair measures by recovered first-round failures.
Costs are estimated API-equivalent model costs.}
\label{tab:eris-efficiency}
\scriptsize
\setlength{\tabcolsep}{3pt}
\renewcommand{\arraystretch}{1.15}
\begin{tabularx}{\linewidth}{@{}Xccc@{}}
\toprule
Metric
& One-shot
& \shortstack{Independent\\regeneration}
& \eris \\
\midrule
$\mathrm{Success}@4$
& 18/40 & 26/40 & \textbf{40/40} \\
\addlinespace
\multicolumn{4}{@{}l}{\textit{Overall efficiency}} \\
Rounds / solved
& 2.22 & 3.81 & \textbf{1.70} \\
Cost / solved
& \$2.59 & \$5.00 & \textbf{\$2.48} \\
\addlinespace
\multicolumn{4}{@{}l}{\textit{Repair efficiency}} \\
Rounds / recovered
& -- & 7.4 & \textbf{1.3} \\
Cost / recovered
& -- & \$10.43 & \textbf{\$2.39} \\
\bottomrule
\end{tabularx}
\end{wraptable}

\paragraph{Generation efficiency}
Closed-loop repair need not increase generation cost if measurements reduce the number of unsuccessful generation rounds.
We compare \eris with two baselines (\autoref{tab:eris-efficiency}).
One-shot performs exactly one generation round per evaluation unit and makes no further attempt after a failure, while independent regeneration repeatedly resubmits the original request without using feedback from previous rounds.
We evaluate both overall generation efficiency and the marginal efficiency of recovering failures after the first round.
Compared with independent regeneration, \eris achieves
higher final success while requiring fewer generation rounds
and lower model cost per solved asset, with larger efficiency
gains when considering only the recovery of first-round failures.

\subsection{Scenario generation}
\label{sec:world-generation}

We introduce a pipeline for generating task-conditioned scenarios and evaluate its scene-generation and background components.
Building a scene manually requires placing the robot and objects and checking whether the configuration satisfies the task requirements.
Repeating this process across the many variants needed for a benchmark is difficult to scale.
Inverse scene generation (ISG) therefore uses staged verification and feedback to construct scene variants that satisfy task-relevant constraints.

ISG generates task-conditioned scene variants from a task $\mathcal T$ and its workplace $\mathcal W$ (\S~\ref{sec:scene-generation}).
A separate pipeline adds photorealistic backgrounds (\S~\ref{sec:photoreal-background}).
A scenario-authoring agent can also write the process logic governing workplace operations (\S~\ref{sec:process-logic}).

\begin{figure}[t]
\centering
\includegraphics[width=\linewidth]{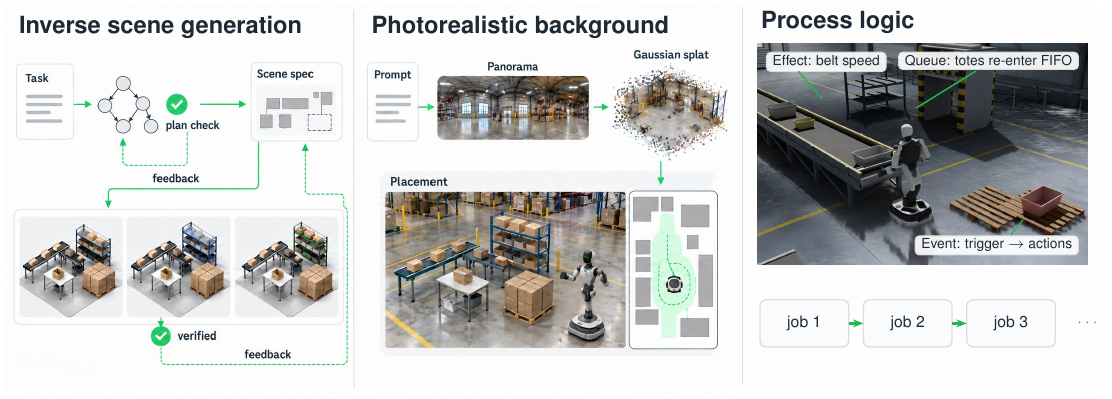}
\caption{Overview of scenario generation. Left, inverse scene generation revises the task plan and then the scene specification until each passes its checks and critics. Middle, a Gaussian-splat background is generated from a prompt and the verified work cell is placed on its free floor. Right, process logic keeps the work going across jobs without a reset. 
}
\label{fig:worldgen-overview}
\end{figure}

\autoref{fig:worldgen-overview} summarizes the scenario-generation pipeline, and \autoref{tab:worldgen-overview} distinguishes its outputs and verification methods.
We evaluate scene generation and background construction without a robot policy, and each experiment is reported with its component.
For the background study, the agent generated the packaging scene from a supplied scene description and applied static verification. A person reviewed the result (\autoref{fig:gs-worlds}). 

\begin{table}[t]
\centering\footnotesize
\caption{Components of scenario generation, their outputs, verification, and reported measures. Process logic is schema-validated when loaded.}
\label{tab:worldgen-overview}
\renewcommand{\arraystretch}{1.3}
\renewcommand{\tabularxcolumn}[1]{m{#1}}
\begin{tabularx}{\linewidth}{@{}>{\raggedright\arraybackslash}m{0.22\linewidth}YYY@{}}
\toprule
\tablehead{Component} & \tablehead{Output} & \tablehead{Verification} & \tablehead{Reported measures}\\
\midrule
Task planning & Task plan, a predicate-based graph of work steps & Deterministic plan checks, then plan critic & Acceptance, cost\\[4pt]
\cmidrule(lr){1-4}
Scene generation & Scene specification and resolved variants & Deterministic scene checks, then scene critics & Acceptance, validity, fidelity, cost\\[4pt]
\cmidrule(lr){1-4}
Background generation & Panorama, Gaussian splat, work-cell placement & Image, view, and stability checks; model assessment & Rubric, stability\\[4pt]
\cmidrule(lr){1-4}
Process logic authoring & Logic with effects, events, queues, and idle joint targets & Schema validation when the scenario is loaded & - \\
\bottomrule
\end{tabularx}
\end{table}

\subsubsection{Inverse scene generation}
\label{sec:scene-generation}

Each generated scene variant must satisfy a set of hard constraints $\mathcal C_{\mathrm{scene}}$, including catalog asset availability, non-interpenetrating placements, valid support relations, and the presence of task-required objects.
The admissible set $\mathcal S_{\mathrm{adm}}$ consists of scene configurations satisfying every constraint in $\mathcal C_{\mathrm{scene}}$.

We use \emph{generate-then-verify} as a baseline for scene-specification generation.
Given a scene description, the agent generates a complete scene specification, which is resolved into scene variants and checked against the requirements.
If the candidate fails verification, it is discarded and the agent starts a new attempt from the original description, without access to the previous candidate or its failure feedback.

In contrast, ISG generates a task plan and a scene specification sequentially, verifying each before proceeding to the next.
The generation agent first summarizes the task steps, measurable success signals, required objects, spatial relations, and variation axes.
The agent uses these preparatory descriptions to generate a task plan $z_1=\Gamma$ and a scene specification $z_2=\hat S$.
The task plan specifies symbolic work steps rather than robot motion trajectories. The scene specification defines the physical configurations needed by those steps and is resolved into scene variants $s_{1:N_s}$.

At each stage, deterministic checks, which run as code without a language model, test the candidate. For the task plan, this is the plan check described below. For the scene specification, the checks verify required objects and relations and static validity across all resolved variants.
Language models then serve as a \emph{plan critic} or \emph{scene critics}, corresponding to \textsc{PlanCritic} and \textsc{SceneCritics} in the algorithm.
These critics run in separate calls after the deterministic checks pass and assess whether the candidate meets the task intent and the criteria for that stage.
A candidate is accepted only when the checks pass and the critics approve it.
Otherwise, the agent revises the candidate using the failed checks, measured violations where available, and the critics' concerns.

The task, workplace, and preparatory descriptions remain fixed throughout revision.
At each stage, revision retains the accepted outputs of earlier stages.
Each stage has a fixed attempt budget.
This use of feedback is related to iterative refinement with model feedback~\citep{madaan2023selfrefine} and coding agents that verify and roll back scene programs~\citep{li2026legoanything}.

Procedural generators sample layouts from hand-written rules or constraint programs~\citep{deitke2022procthor,raistrick2024infinigenindoors}, language models map descriptions to layouts~\citep{feng2023layoutgpt,yang2024holodeck}, and recent agentic systems add simulation-ready assets and physical checks, sometimes conditioned on a task or language specification~\citep{xia2026sage,pfaff2026scenesmith,wang2026embodiedgenv2}.
Task-generation systems propose a task and then build a scene for it~\citep{wang2024robogen,hua2024gensim2}.

\begin{algorithm}[t]
\caption{Inverse scene generation}
\label{alg:inverse-scene}
\small
\begin{algorithmic}[1]
\Require Task $\mathcal T$ of workplace $\mathcal W$, catalog $\mathcal C$, positive budgets $R_{\mathrm{plan}},R_{\mathrm{scene}}$, variants $N_s$
\Ensure Scene specification $\hat S$ with variants $s_{1:N_s}$, or a typed stop
\Statex Calls retain prior candidates and feedback in context; typed stops terminate the procedure.
\State $O\gets\Call{Outline}{\mathcal T,\mathcal W}$; \quad $P\gets\Call{ScenePlan}{O}$
\State $\Gamma\gets\varnothing$; \quad $\hat S\gets\varnothing$; \quad $e\gets\varnothing$
\For{$r=1,\ldots,R_{\mathrm{plan}}$}
  \State $\Gamma\gets\Call{TaskPlan}{P,\Gamma,e}$; \quad $e\gets\Call{PlanCheck}{\Gamma,P}$ \Comment{deterministic checks}
  \State \textbf{if} $e=\varnothing$ \textbf{then} $e\gets\Call{PlanCritic}{\Gamma,P}$
  \If{$e=\varnothing$} \textbf{break} \EndIf
\EndFor
\If{$e\neq\varnothing$} \Return \textsc{blocked} \EndIf
\For{$r=1,\ldots,R_{\mathrm{scene}}$}
  \State $\hat S\gets\Call{Specify}{P,\Gamma,\mathcal C,\hat S,e}$ \Comment{asset selection and authoring}
  \State $(s_{1:N_s},e)\gets\Call{SceneCheck}{\hat S,P,\Gamma,N_s}$ \Comment{static, resolve, variant checks}
  \State \textbf{if} $e=\varnothing$ \textbf{then} $e\gets\Call{SceneCritics}{\hat S,s_{1:N_s},\Gamma}$
  \If{$e=\varnothing$} \Return $\hat S,s_{1:N_s}$ \EndIf
  \State \Call{Escalate}{$e$} \Comment{repair, restart with context, or stop}
\EndFor
\State \Return \textsc{budget}
\end{algorithmic}
\end{algorithm}

\autoref{alg:inverse-scene} gives the procedure.
The agent first attempts to fix the failure using the feedback. If the same failure recurs, the stage restarts in a fresh context. A third occurrence ends the run.

\paragraph{Task plan}
\label{par:scene-task-plan}
Given a person-supplied task $\mathcal T$ and its workplace $\mathcal W$ (\S~\ref{sec:overview-preliminaries}), the agent authors a task plan $\Gamma$ as a directed graph of work steps.
Nodes specify predicate preconditions and effects, and edges specify success, failure, or recovery transitions.
Its initial predicate state is $\sigma_{\mathrm{init}}$, its goal is $g$, and its invariants are $\mathcal I$.
We denote the states reachable under the plan by $\mathrm{Reach}_\Gamma(\sigma_{\mathrm{init}})$ and the objects and relations required by the plan by $\mathrm{Req}(\Gamma)$.
A valid task plan requires that the goal does not already hold in the initial state, that at least one reachable state satisfies the goal, and that every reachable state preserves the invariants:
\begin{align}
  &\sigma_{\mathrm{init}}\not\models g, \qquad
   \exists\,\sigma\in\mathrm{Reach}_\Gamma(\sigma_{\mathrm{init}}):\ \sigma\models g, \qquad
   \forall\,\sigma\in\mathrm{Reach}_\Gamma(\sigma_{\mathrm{init}}):\ \sigma\models\mathcal I.
   \label{eq:plan-valid}
\end{align}
Before any geometry exists, the plan check explores up to $10^4$ reachable predicate states to check goal reachability, invariant violations, dead ends, and cycle closure. It also checks representability in the scene specification.
This bounded search does not establish the universal condition in Eq.~\eqref{eq:plan-valid} when the reachable state space is not fully explored.
Each hard requirement is assigned to a deterministic check that must pass. A requirement that no check can verify, such as a hard visibility or reachability relation, is rejected at this stage.
The plan critic evaluates the task plan in a fresh context and accepts it only with a score of at least 0.8 and no open concern.
The agent revises the task plan until its deterministic checks pass and the plan critic approves it, or the budget is exhausted.

\paragraph{Scene specification}
After task-plan acceptance, the agent selects catalog assets and authors a scene specification $\hat S$, a typed representation of the workspace, objects and poses, optional background, and variation axes.
If a required asset is missing from the catalog, scene generation stops unless the run allows the agent to generate new assets (\autoref{tab:plan-critic}).
A compiler emits USD. The resolver $s_k=\textsc{Resolve}(\hat S,\mathrm{seed}_k)$ samples only declared variations within the task plan's pose bounds, producing the scene variants $s_{1:N_s}$ from one specification.
Every resolved variant must contain the objects and relations required by the task plan and pass static verification. Static checks cover asset and reference validity, spawn collisions, support relations, and geometric tolerances. 
We allow at most 5 mm of overlap or ground penetration and 1 cm of support clearance. 
The resolved variants must also differ only along the declared variation axes.
After the scene specification and its resolved variants pass all deterministic scene checks, the scene critics assess them against the task plan.

A scenario stores collision-buffer sizes for the contacts it can produce at once, and the batch size is chosen from the resulting memory estimate before allocation (\S~\ref{app:collision-memory}).

The following CNC request illustrates the task steps and success signals used to construct a task plan.
\begin{promptbox}{Example task and generated steps}
\textbf{Task.} Generate a long-horizon CNC tending workcell that repeatedly loads two shaft families, unloads finished parts, routes failed gauge results to quarantine, replenishes an empty infeed tray, and performs a fixture changeover between batches. Carry machine occupancy, tool-wear state, and completed-part count across cycles without resetting the whole cell.\\[0.4em]
\textbf{Step} \texttt{load\_blank}. Pick a shaft blank from the active infeed tray and load it into the lathe chuck. \emph{Success:} blank seated in chuck within tolerance, door closed, machine occupancy=true.\\[0.3em]
\textbf{Step} \texttt{gauge\_route}. Gauge part and route to finished tray or quarantine bin. \emph{Success:} passed part in finished tray and count incremented, or failed part in quarantine bin.\\[0.3em]
\textbf{Step} \texttt{fixture\_changeover}. Between batches, swap chuck fixture set for the other shaft family. \emph{Success:} new fixture mounted, old fixture returned to rack, active family switched.
\end{promptbox}

In this example, the task plan uses predicates such as \texttt{chuck\_empty} and carries tool wear and part counts across eight cycles.

\begin{figure}[t]
\centering
\begin{minipage}[t]{0.485\linewidth}
\centering
\includegraphics[width=\linewidth]{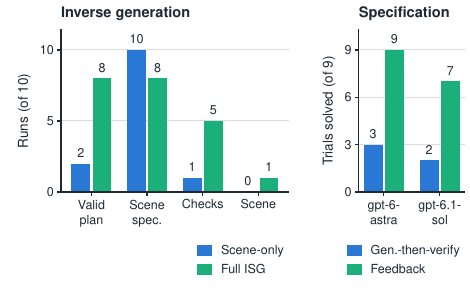}
\caption{Ablations of staged verification and feedback. Left, full ISG with gpt-6.1-sol on the ten tasks of \autoref{tab:model-ablation}, with and without the plan check and plan critic. Valid plan applies the plan check retrospectively when it was skipped. Right, specification-generation trials solved within five attempts, with and without feedback.}
\label{fig:paradigms}
\end{minipage}\hfill
\begin{minipage}[t]{0.485\linewidth}
\centering
\includegraphics[width=\linewidth]{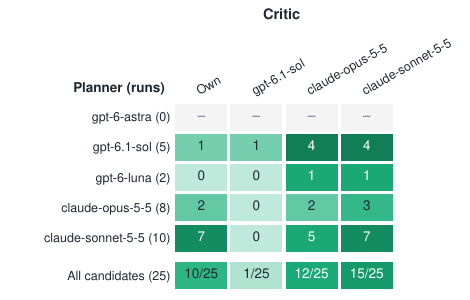}
\caption{Accepted scenes when the fully checked candidates of \autoref{tab:model-ablation} are re-judged by a fixed critic. Parentheses give the number of such candidates, Own is the planner's own critic, and color shows the accepted fraction.}
\label{fig:critic-swap}
\end{minipage}
\end{figure}

\paragraph{Metrics}
We evaluate generation without a robot policy using acceptance, static validity, specification fidelity, and cost.
\begin{enumerate}[leftmargin=*,itemsep=0.15em,topsep=0.3em]
  \item \emph{Acceptance}, the number of runs that reach each stage. A candidate is \emph{fully checked} if it passes every deterministic check and \emph{accepted} if the critics of its stage also approve it.
  \item \emph{Static validity}, the number of candidates whose specification passes static verification.
  \item \emph{Specification fidelity}, agreement with a reference specification on three tiers (two boxes and one item on a table, three items on each of three shelves, sixteen items in two trays), 
  scored on nine criteria (Appendix~\ref{app:ablation-protocol}).
  \item \emph{Cost}, wall time and model calls per accepted scene, and mean wall time per run.
\end{enumerate}
The specification-generation trials isolate scene-specification generation by starting from supplied scene descriptions and omitting task-plan generation (\autoref{tab:scene-modes}).
Success in these trials means meeting the reference-based fidelity criteria. It is distinct from acceptance by the plan and scene critics in full ISG.
When planners are compared, acceptance is also reported under a fixed critic, because a critic that shares the planner's model can be lenient (\autoref{fig:critic-swap}, \autoref{tab:fixed-critic}).

\paragraph{Ablations of staged verification and feedback}
We separately test the effects of task-plan verification in the full pipeline and feedback-based repair within scene-specification generation.
The \emph{scene-only verification} ablation still generates a task plan but omits its deterministic checks and the plan critic. It therefore differs from the specification-generation trials, which omit task-plan generation entirely.
The \emph{generate-then-verify} baseline independently resamples scene specifications without feedback from previous failures.
Scene-only verification reaches a scene specification in all ten runs and three times faster. However, only 1 run yields a candidate that passes every deterministic check, against 5 with full ISG, and 8 of its 10 unchecked plans fail the plan check when tested afterwards (\autoref{fig:paradigms}, left).
Within scene-specification generation, removing failure feedback reduces the number of successful trials.
Generate-then-verify solves 3 and 2 of 9 trials within five attempts with gpt-6-astra and gpt-6.1-sol, against 9 and 7 with feedback (\autoref{fig:paradigms}, right).

\paragraph{Ablation over frontier models}
Scene acceptance depends substantially on the model critic.
Under a shared claude-opus-5-5 critic, gpt-6.1-sol and claude-sonnet-5-5 achieve identical acceptance rates despite substantial differences under their own critics (\autoref{tab:fixed-critic}).
Re-evaluating the same candidates reveals disagreements over stand-in assets, primitive fixtures, and support gaps (\autoref{fig:critic-swap}).
Consequently, the cost per accepted scene also depends on critic selection (\autoref{tab:scene-cost}). In separate specification-generation trials, verification feedback increases success from 4/54 first attempts to 40/54 within five attempts (\autoref{tab:model-ablation}).
However, gemini-4-argon fails the easiest tier on object positioning. These experiments use ten tasks from a separate 30-task industrial corpus (\S~\ref{app:ablation-protocol}), rather than the seven benchmark tasks (\autoref{tab:bench-tasks}). 

\begin{table}[t]
\centering\small
\caption{Planning with a critic shared by both planners. Each row runs the planner on the ten tasks of \autoref{tab:model-ablation} with the same budgets, and Critic names the model that serves every plan and scene critic call. Rows with the planner's own critic repeat \autoref{tab:model-ablation}. Min is the mean wall time per run in minutes.}
\label{tab:fixed-critic}
\begin{tabular*}{\linewidth}{@{\extracolsep{\fill}}llccccc@{}}
\toprule
\tablehead{Planner} & \tablehead{Critic} & \tablehead{Plan} & \tablehead{Scene spec.} & \tablehead{Checks} & \tablehead{Scene} & \tablehead{Min}\\
\midrule
gpt-6.1-sol & Own & 8/10 & 8/10 & 5/10 & 1/10 & 27\\
gpt-6.1-sol & claude-opus-5-5 & 7/10 & 7/10 & 4/10 & 3/10 & 28\\
\addlinespace
claude-sonnet-5-5 & Own & 10/10 & 10/10 & 10/10 & 7/10 & 12\\
claude-sonnet-5-5 & claude-opus-5-5 & 9/10 & 9/10 & 9/10 & 3/10 & 22\\
\bottomrule
\end{tabular*}
\end{table}

\begin{table}[t]
\centering\small
\caption{Frontier-model ablation for scene generation. Planning counts, over ten tasks, runs with an accepted plan, runs with at least one candidate that passes every deterministic check, and runs that end in an accepted scene, judged by critics of the planner's model, and the mean wall time per run in minutes. Wall times include provider latency.
}
\label{tab:model-ablation}
\begin{tabular*}{\linewidth}{@{\extracolsep{\fill}}lccccccccc@{}}
\toprule
 & \multicolumn{5}{c}{\tablehead{Specification generation}} & \multicolumn{4}{c}{\tablehead{Full ISG}}\\
\cmidrule(lr){2-6}\cmidrule(l){7-10}
\tablehead{Model (effort)} & Easy & Medium & Hard & 1st & Attempts & Plan & Checks & Scene & Min\\
\midrule
gpt-6-astra (high) & 3/3 & 3/3 & 3/3 & 2/9 & \textbf{2.44} & 5/10 & 0/10 & 0/10 & 29\\
gpt-6.1-sol (medium) & 3/3 & 3/3 & 1/3 & 0/9 & 2.57 & 8/10 & 5/10 & 1/10 & 27\\
gpt-6-luna (medium) & 2/3 & 0/3 & 1/3 & 0/9 & 3.33 & 3/10 & 2/10 & 0/10 & 13\\
claude-opus-5-5 (high) & 1/3 & 3/3 & 3/3 & 0/9 & 3.29 & 9/10 & 8/10 & 2/10 & 29\\
claude-sonnet-5-5 (medium) & 3/3 & 2/3 & 3/3 & 2/9 & 3.00 & \textbf{10/10} & \textbf{10/10} & \textbf{7/10} & \textbf{12}\\
gemini-4-argon (high) & 0/3 & 3/3 & 3/3 & 0/9 & 2.67 & \textbf{10/10} & 8/10 & 6/10 & 22\\
\bottomrule
\end{tabular*}
\end{table}

\begin{figure}[t]
\centering

\begin{subfigure}[t]{0.24\linewidth}
  \includegraphics[width=\linewidth]{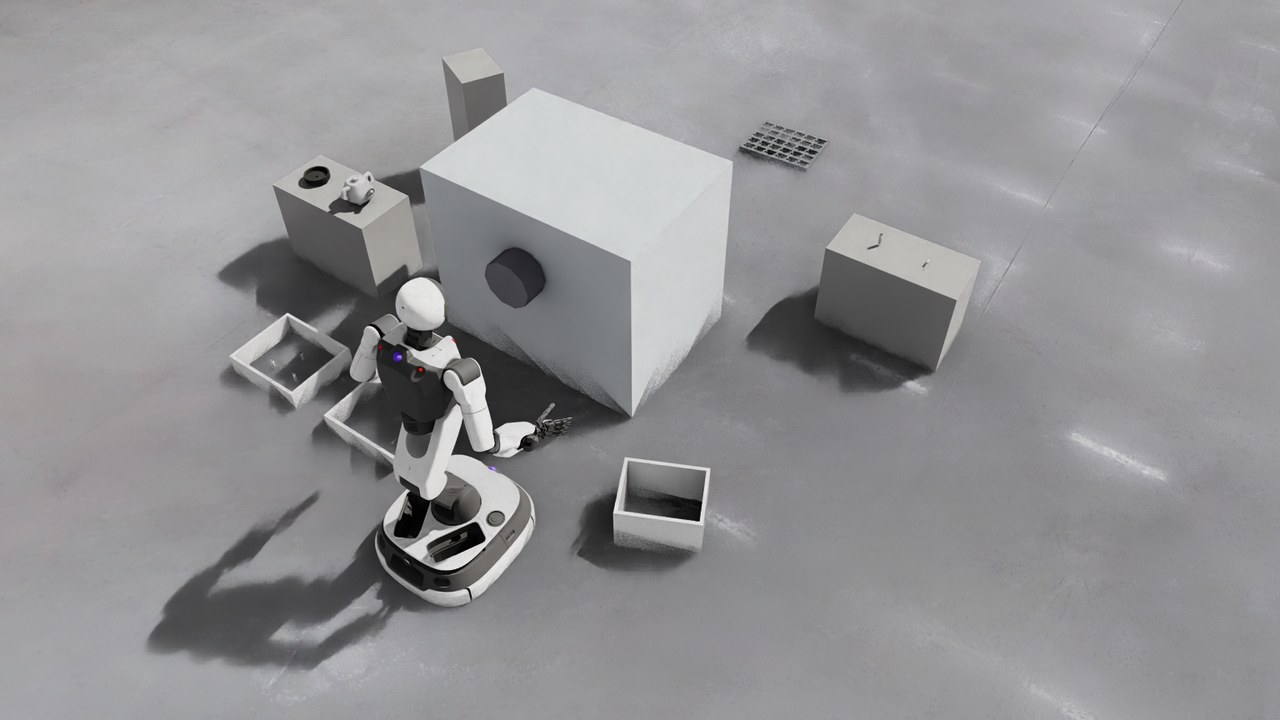}
  \caption{CNC tending, variant 1}
\end{subfigure}\hfill
\begin{subfigure}[t]{0.24\linewidth}
  \includegraphics[width=\linewidth]{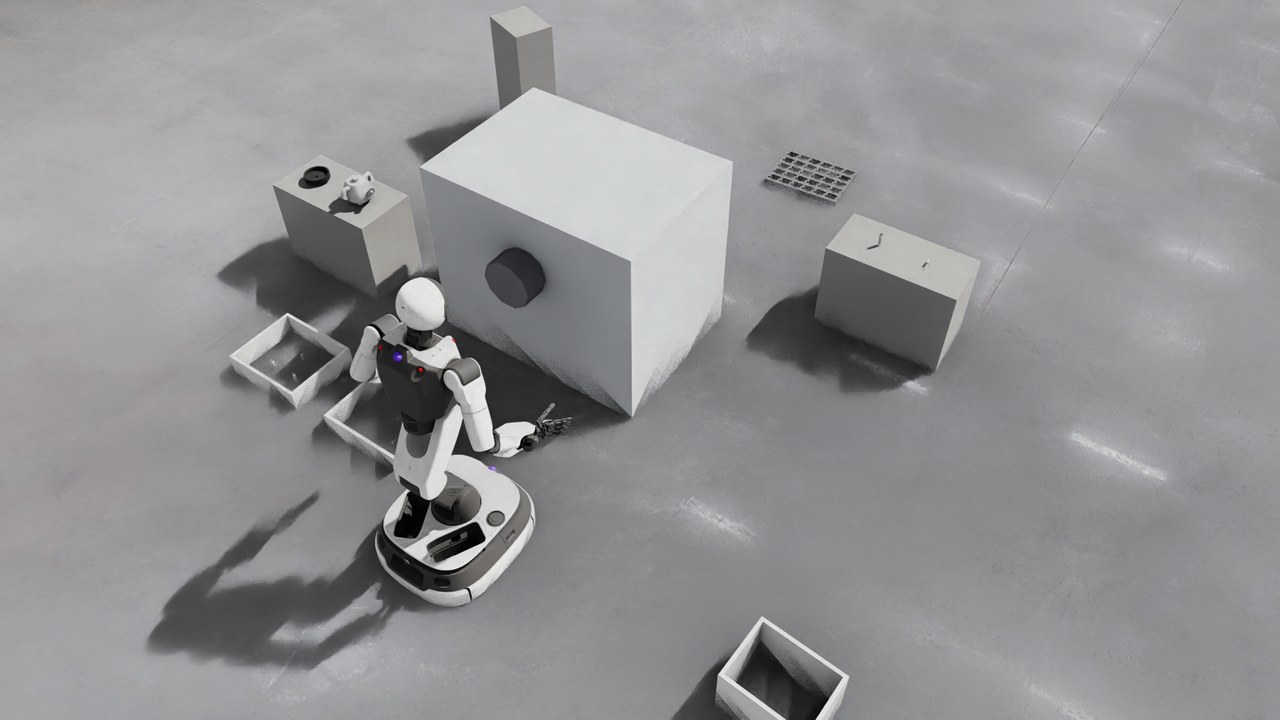}
  \caption{CNC tending, variant 2}
\end{subfigure}\hfill
\begin{subfigure}[t]{0.24\linewidth}
  \includegraphics[width=\linewidth]{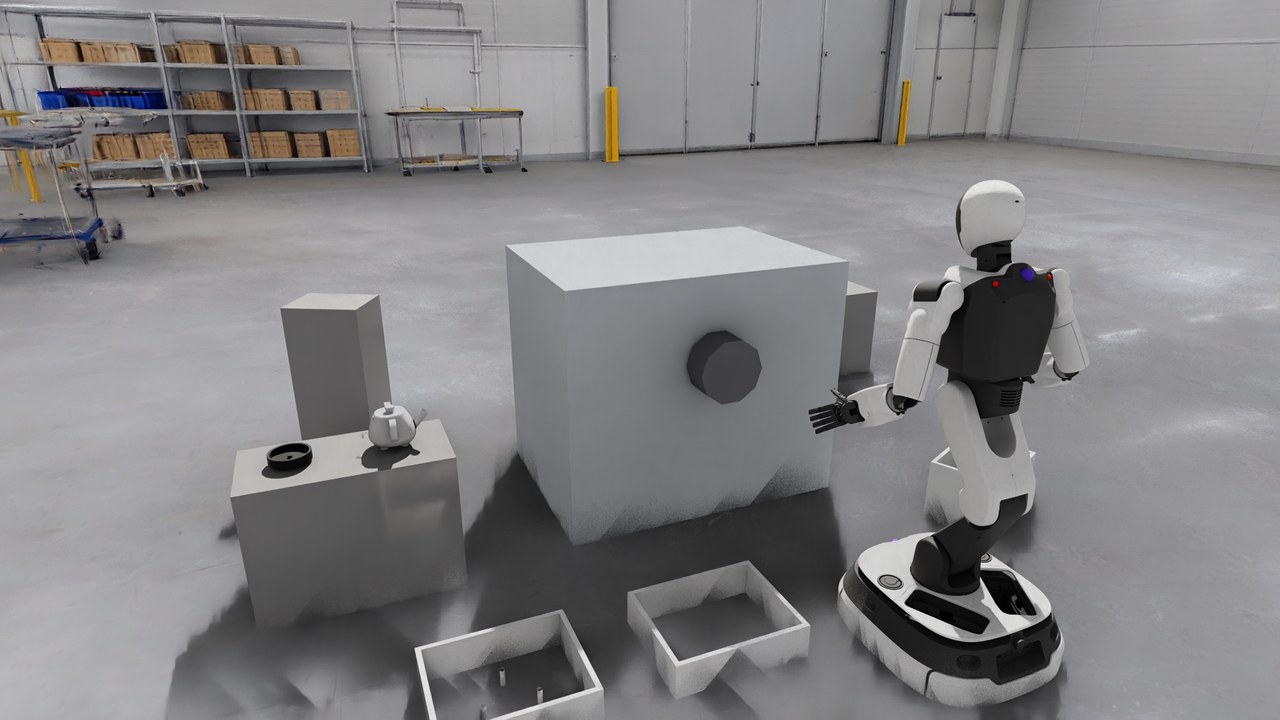}
  \caption{CNC cell detail}
\end{subfigure}\hfill
\begin{subfigure}[t]{0.24\linewidth}
  \includegraphics[width=\linewidth]{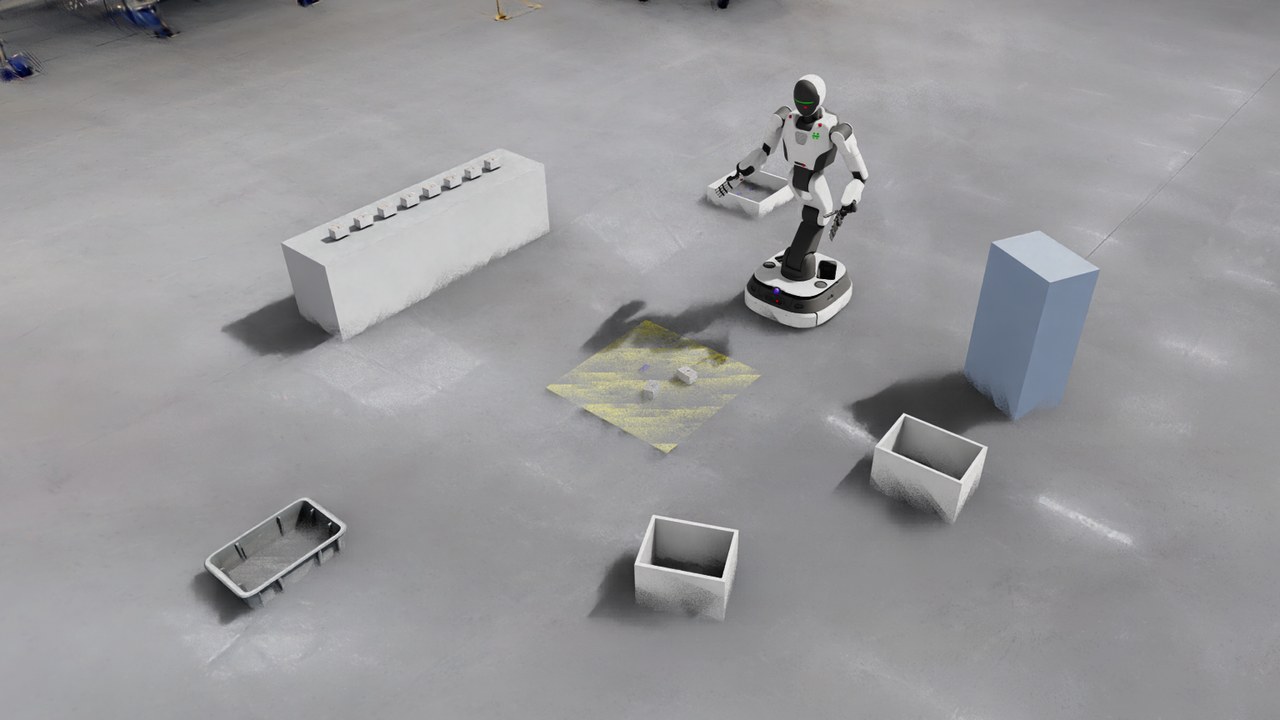}
  \caption{Secondary packaging}
\end{subfigure}

\vspace{2mm}

\begin{subfigure}[t]{0.24\linewidth}
  \includegraphics[width=\linewidth]{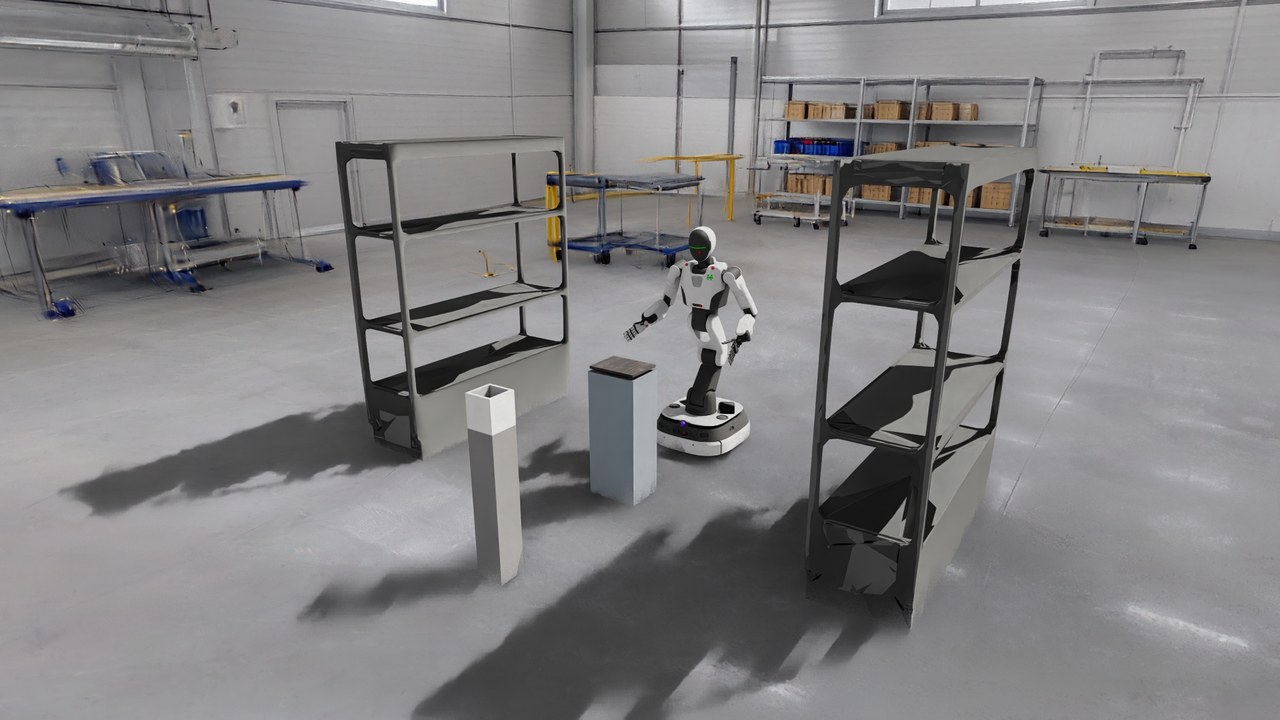}
  \caption{Adhesive application}
\end{subfigure}\hfill
\begin{subfigure}[t]{0.24\linewidth}
  \includegraphics[width=\linewidth]{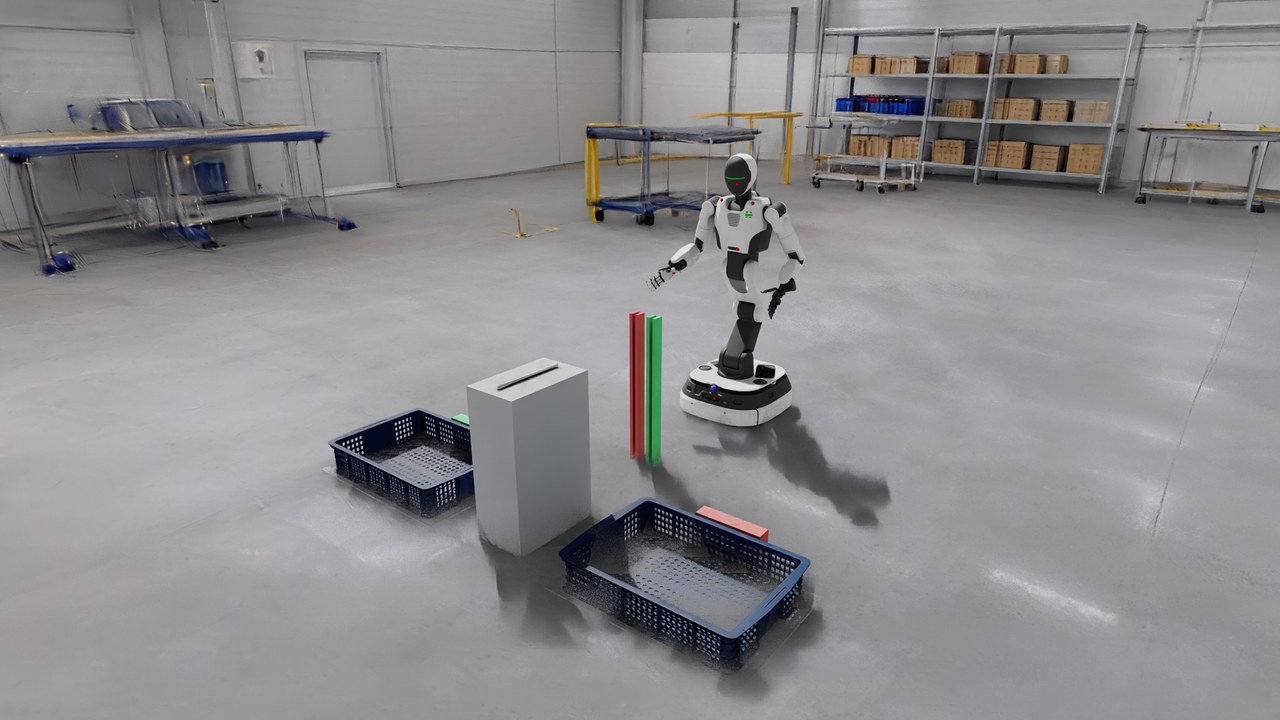}
  \caption{Go/no-go gauging}
\end{subfigure}\hfill
\begin{subfigure}[t]{0.24\linewidth}
  \includegraphics[width=\linewidth]{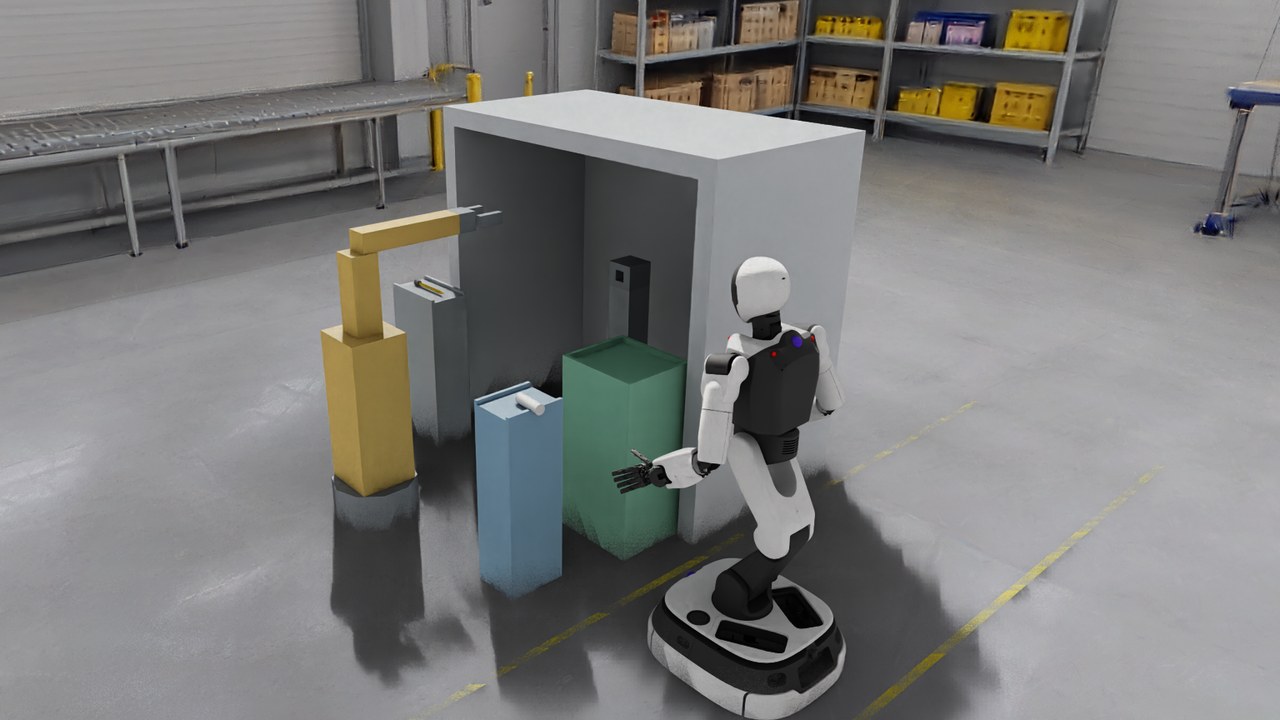}
  \caption{CNC chuck exchange}
\end{subfigure}\hfill
\begin{subfigure}[t]{0.24\linewidth}
  \includegraphics[width=\linewidth]{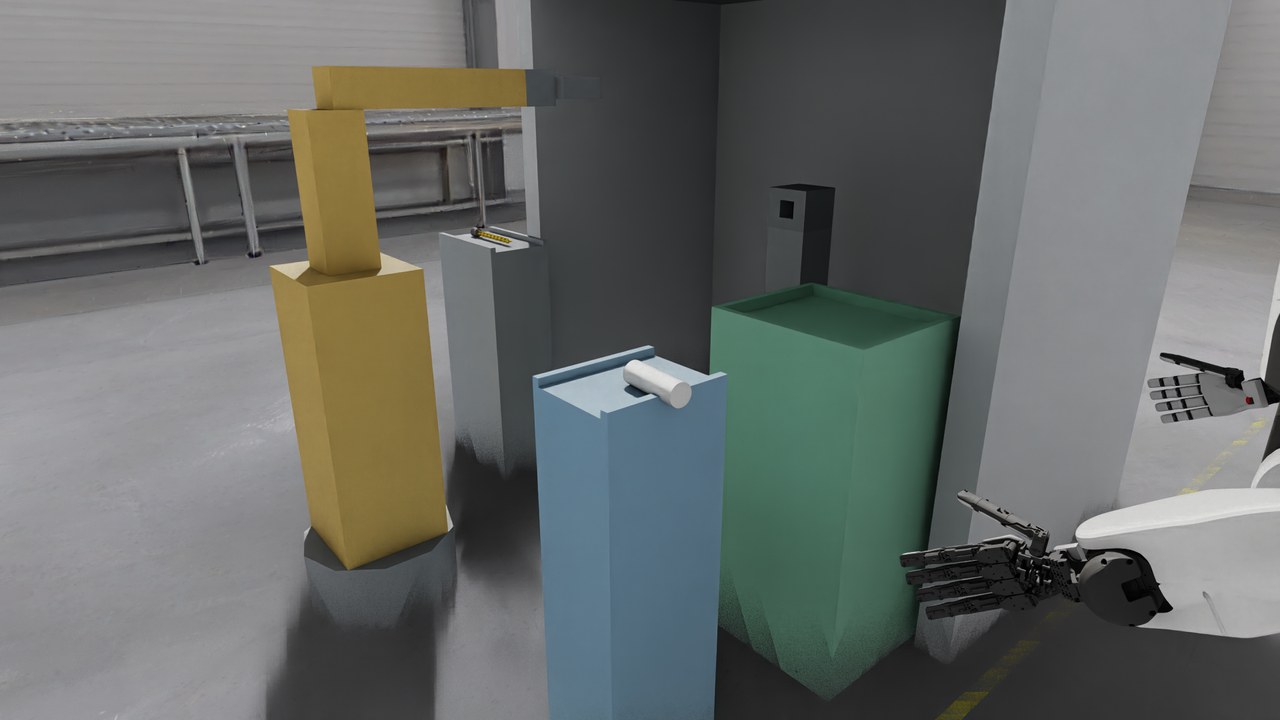}
  \caption{Generated billet}
\end{subfigure}

\caption{Scenes accepted by inverse scene generation under each planner's own critics.
(a--c) CNC tending with two scene variants and a six-step plan generated by claude-sonnet-5-5.
(d) Secondary packaging with a six-step plan generated by claude-sonnet-5-5.
(e, f) Scenes generated by gpt-6.1-sol.
(g, h) CNC chuck exchange with an asset generated during scene construction (\autoref{tab:plan-critic}).
A stricter critic rejects (a--d) (\autoref{fig:critic-swap}).
All panels show static renders at $t=0$ with generated Gaussian-splat backgrounds (\S~\ref{sec:photoreal-background}) and manually placed \friday.}
\label{fig:scene-generation}
\end{figure}

\paragraph{Bottlenecks}
Missing assets and limited scene-specification expressivity constrain inverse scene generation.
With gpt-6.1-sol, 8/10 runs yield accepted plans, but only 1/10 produces an accepted scene (\autoref{tab:model-ablation}).
Increasing the generation budget does not resolve asset availability, which blocks most runs (\autoref{tab:plan-critic}).

Accepted scenes often rely on primitive substitutes for missing fixtures (\autoref{fig:scene-generation}).
Because scene critics inspect catalog descriptions rather than renders, they can overlook mismatches such as a vise-like fixture represented by a teapot.
Explicitly specifying the required affordances of generated assets enables one additional accepted scene (\autoref{fig:scene-generation}).
The remaining failures expose requirements that the current scene specification cannot express (\autoref{tab:plan-critic}).

\begin{table}[t]
\centering\small
\caption{Removing bottlenecks of inverse scene generation one at a time. Each row runs gpt-6.1-sol on the same ten industrial tasks and counts runs with an accepted plan, runs that reach a scene specification, runs with at least one candidate that passes every deterministic check, and runs that end in an accepted scene.}
\label{tab:plan-critic}
\begin{tabular*}{\linewidth}{@{\extracolsep{\fill}}lcccc@{}}
\toprule
\tablehead{Condition} & \tablehead{Plan} & \tablehead{Scene spec.} & \tablehead{Checks} & \tablehead{Scene}\\
\midrule
Default budget (40\,min, 40 calls) & 8/10 & 8/10 & 5/10 & 1/10\\
+ Larger budget (2\,h, 80 calls) & 9/10 & 9/10 & 2/10 & 1/10\\
+ Asset generation & 9/10 & 9/10 & 2/10 & 0/10\\
+ Affordance format in prompt & 9/10 & 9/10 & 3/10 & 2/10\\
\bottomrule
\end{tabular*}
\end{table}

Rubric judges score the candidates after generation, separately from the plan and scene critics. Asset fit largely drives their scores, which cluster in a narrow range (\S~\ref{app:rubric-prompts}).

\paragraph{Threats to validity}
Specification fidelity is measured against one reference per tier derived from the evaluation rubric and does not establish broader scene quality.
In \autoref{tab:model-ablation}, each planner uses critics from its own model, so scene acceptance reflects both generation quality and critic decisions.
The critic-swap and shared-critic experiments partially separate these effects (\autoref{fig:critic-swap}, \autoref{tab:fixed-critic}), although the shared-critic comparison uses only one run per task and model.
The model families also differ in their agent interfaces.
Without an independent human audit, critic agreement does not establish the suitability of accepted scenes for the requested tasks.
Finally, plan checks explore at most $10^4$ predicate states, and static geometric checks are approximate.

\subsubsection{Photorealistic background}
\label{sec:photoreal-background}

Background generation runs separately from ISG.
After scene verification, we place each simulated cell inside generated 3D surroundings rendered as Gaussian splats.
They form the background of a scene variant and therefore enter only its observation model $O_s$ (\S~\ref{sec:overview-preliminaries}).
Before a background is added, the scene contains task-relevant geometry but leaves the surrounding space empty.
Real-to-sim methods reconstruct each scene from captured data~\citep{torne2024rialto,han2025re3sim,jiang2025gsworld,wang2026vlk}, which for a facility requires physical access and a capture session.
We therefore generate the surroundings with 3D world generators~\citep{worldlabs2025marble,hunyuanworld2025,yang2025matrix3d} and render them as 3D Gaussian splats~\citep{kerbl2023gaussian}, as in simulators that pair splat rendering with rigid-body physics~\citep{qureshi2024splatsim,jia2026gsplayground,kim2026legs}.
Unlike systems that align a collision mesh to the splat~\citep{kim2026legs}, the generated background is used for rendering only.
A generated interior is not a measured facility, even when the generator is conditioned on a reference photograph.

\paragraph{Background generation}
Prompts are sampled from eight facility presets, such as electronics, food, automotive, and logistics plants, over 17 design factors such as ceiling, lighting, racks, and density (\autoref{tab:bg-design-space}).
They require free floor space for the separately simulated work cell, since a plausible hall can still leave no room for it (\S~\ref{app:bg-config}, \autoref{fig:background-failure}).
We assess the panorama before reconstructing it in 3D, as DreamScene360 does during iterative visual assessment and prompt refinement~\citep{zhou2024dreamscene360}.
Panorama-to-3D reconstruction produces a Gaussian splat~\citep{zhou2024dreamscene360,xiong2025panodreamer} through one of the following generator paths.
The proprietary path sends a panorama or reference image to a hosted world-generation service~\citep{worldlabs2025marble} and returns a splat.
The open-source path chains an image-to-panorama diffusion model~\citep{yang2025matrix3d}, a multi-view generator that renders camera paths through the panorama, and 3DGS optimization on those views~\citep{hyworld2026hyworld2}, in the spirit of PanoDreamer~\citep{xiong2025panodreamer}. A lighter variant lifts the panorama with monocular panoramic depth~\citep{jiang2025da360}, as DreamScene360 does with a different depth model~\citep{zhou2024dreamscene360}.
The provider records little beyond model and seed, whereas the open-source chain records every seed, checkpoint, and rejected view.
Both hallucinate geometry the panorama does not show.

We estimate a six-plane room enclosure to remove out-of-bounds Gaussians, align the floor and walls, and set metric scale.
We then search background transforms that place the work-cell footprint on free floor, ranking them by minimum clearance with a safety margin while keeping the simulated scene coordinates fixed.
Distinct translations and yaw angles provide placements for scenario variants.
Appendix~\ref{app:bg-config} gives the procedure and parameters, with an example in \autoref{fig:background-preparation}.

\paragraph{Verification}
Verification first checks the panorama and background views, then evaluates the background with the work cell in place.
The checks cover alignment and scale, footprint clearance, and overlap with task-relevant geometry.
A recorded rollout is re-rendered with the background to check missing-background coverage and the temporal stability of fixed regions along the camera trajectories.
Where a stage uses a VLM judge, deterministic checks run first and every judged criterion must pass (\autoref{tab:app-thresholds}).
Generation and diagnostic renders can be inspected before qualification. Installation requires both panorama and background-view QA to pass, followed by qualification on integrated scene views and rollout videos.
Skipping VLM evaluation leaves the result unverified and does not permit installation.
We re-render recorded trajectories under different backgrounds without rerunning the policy, processing variants in chunks to limit peak memory usage (\S~\ref{app:rendering}).

\begin{figure}[t]
\centering
\begin{subfigure}[t]{0.327\linewidth}\includegraphics[width=\linewidth]{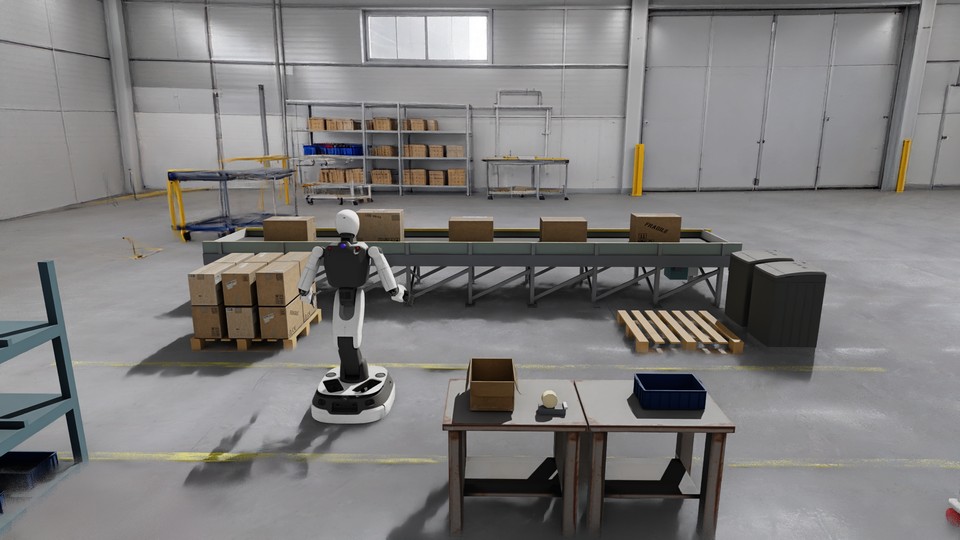}\caption{}\end{subfigure}\hfill
\begin{subfigure}[t]{0.327\linewidth}\includegraphics[width=\linewidth]{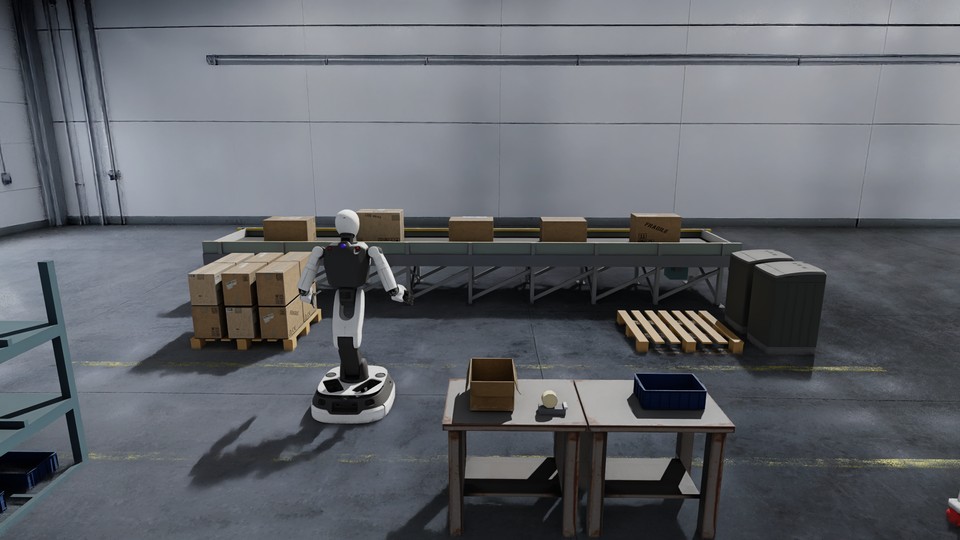}\caption{}\end{subfigure}\hfill
\begin{subfigure}[t]{0.327\linewidth}\includegraphics[width=\linewidth]{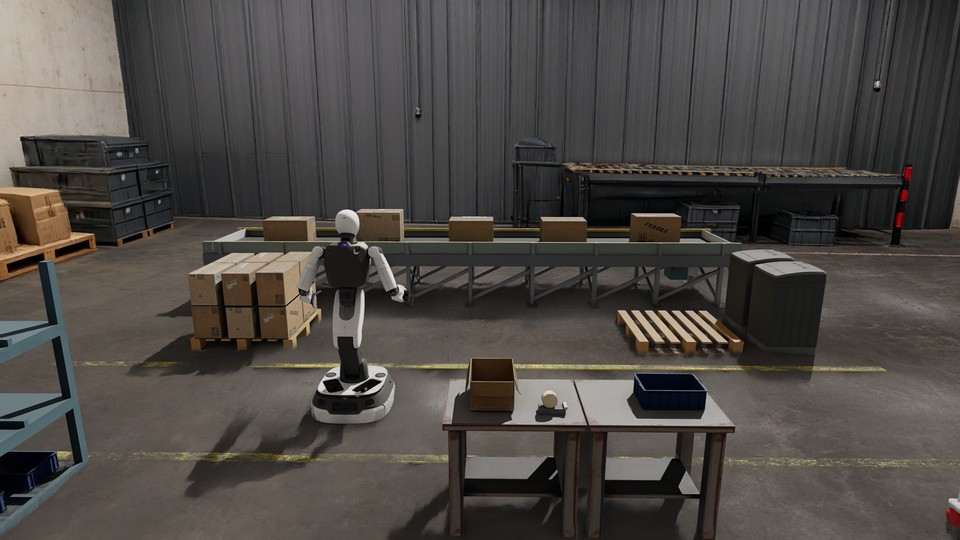}\caption{}\end{subfigure}

\vspace{1mm}
\begin{subfigure}[t]{0.327\linewidth}\includegraphics[width=\linewidth]{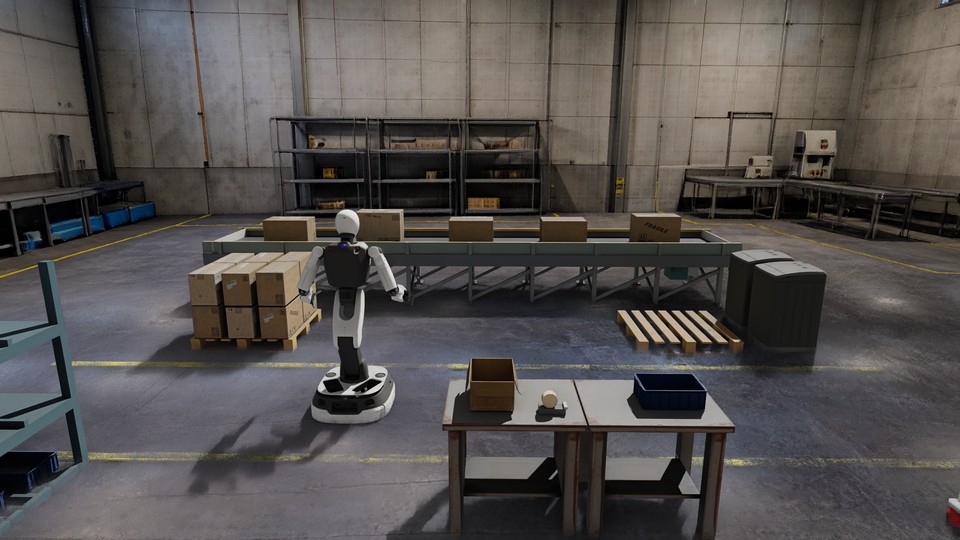}\caption{}\end{subfigure}\hfill
\begin{subfigure}[t]{0.327\linewidth}\includegraphics[width=\linewidth]{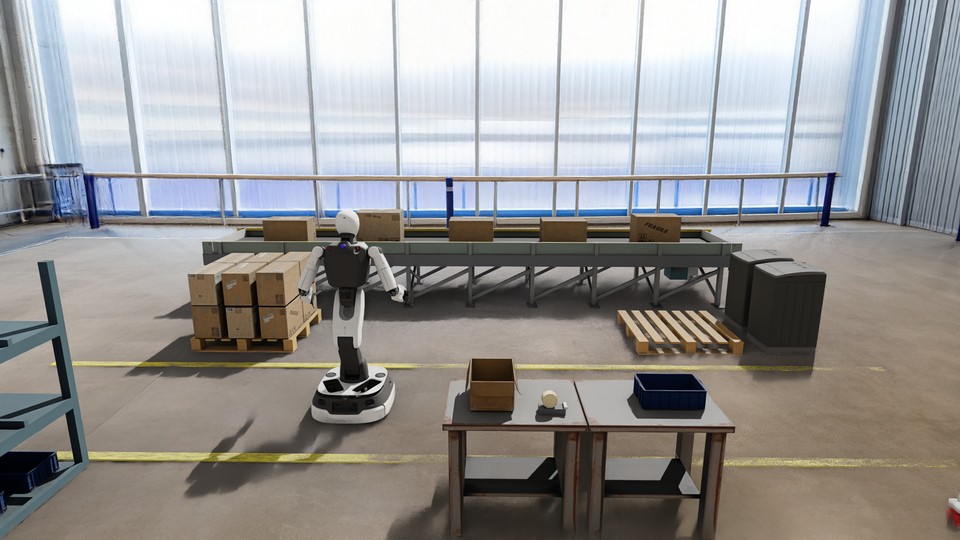}\caption{}\end{subfigure}\hfill
\begin{subfigure}[t]{0.327\linewidth}\includegraphics[width=\linewidth]{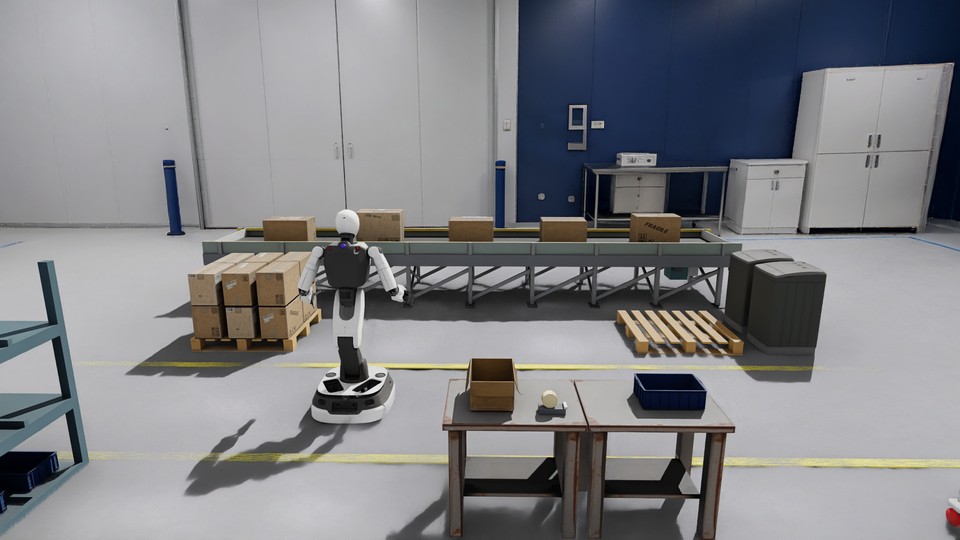}\caption{}\end{subfigure}

\vspace{1mm}
\begin{subfigure}[t]{0.327\linewidth}\includegraphics[width=\linewidth]{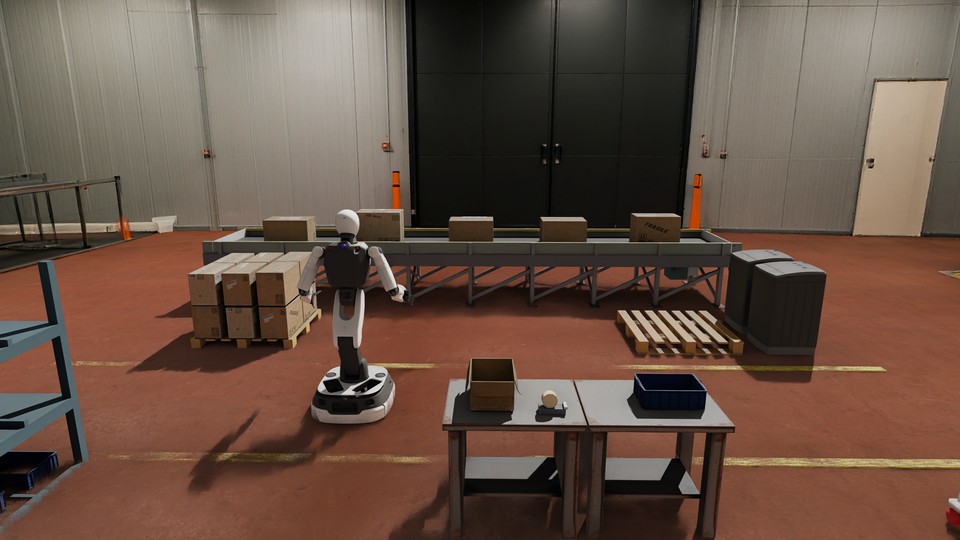}\caption{}\end{subfigure}\hfill
\begin{subfigure}[t]{0.327\linewidth}\includegraphics[width=\linewidth]{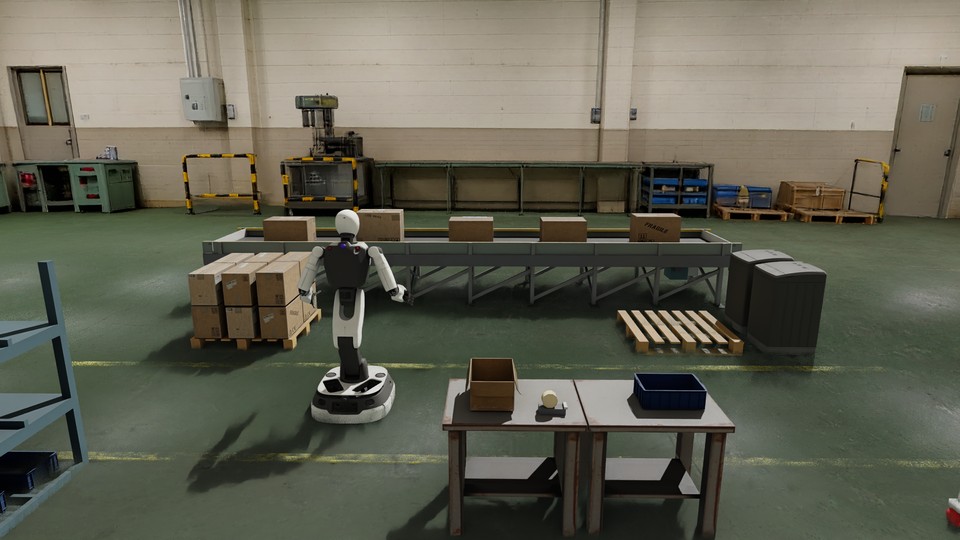}\caption{}\end{subfigure}\hfill
\begin{subfigure}[t]{0.327\linewidth}\includegraphics[width=\linewidth]{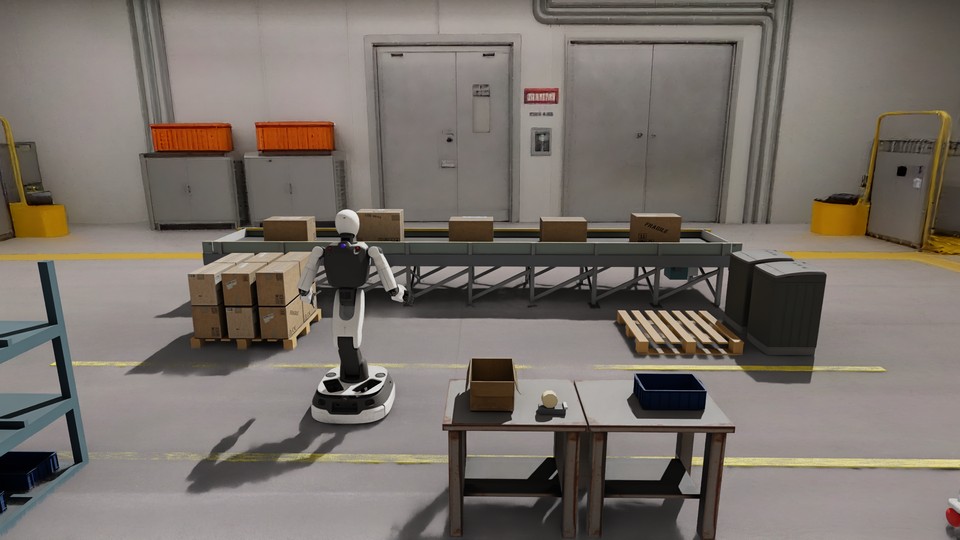}\caption{}\end{subfigure}
\caption{One cell under nine backgrounds, nine scene variants with identical dynamics.
The packaging cell  is placed by clearance search (\S~\ref{app:bg-config}) in nine generated Gaussian-splat warehouses.
Geometry, assets, robot pose, and camera are identical across panels, so the variants share $F_{s,\ell}$ and differ only in $O_s$ (Eq.~\eqref{eq:run-execution}).
}
\label{fig:gs-worlds}
\end{figure}

\paragraph{Results}
We evaluate the backgrounds using the packaging scene generated with human review (\autoref{fig:gs-worlds}).
A blind VLM judge compares the proprietary and open-source backgrounds on nine matched frames (\S~\ref{par:bg-comparison}, \autoref{tab:bg-paths}).
It prefers the proprietary splat in all 18 pairwise comparisons and scores it 2.85 against 1.78 on a five-point rubric.
The judge rates neither background fully usable, and the comparison rests on one reference image and one judge model.
These post-hoc rubric judgments are separate from the pipeline verification gates. The preference results show a relative advantage for the proprietary background in this comparison, but do not establish that either background is ready for use. Retention in the background catalog does not change that assessment.

We automatically place the packaging cell in the free floor space.
\autoref{fig:gs-worlds} shows the cell in nine of the eleven generated backgrounds.

\subsubsection{Process logic generation}
\label{sec:process-logic}
Process logic $\ell$ governs workplace operations within a scenario $S=(s_{1:N_s},\ell)$, including equipment behavior and work supply during execution.
The scenario-authoring agent can specify $\ell$ using four typed components: effects, events, queues, and idle targets (\S~\ref{app:process-logic}).
Of the seven benchmark tasks, conveyor loading uses one belt and one queue, flow-rack loading three belts and one queue, press tending five events, and barcode routing one belt and two events.

Schema validation checks the representation when a scenario is loaded.
Finalization additionally checks event construction, queue configuration, and effect-parameter bindings for each scene variant, rejecting unsupported effect types.
These checks establish structural validity but not whether the logic implements the intended industrial process.

\subsection{Evaluation generation}
\label{sec:eval-authoring}

\begin{figure}[t]
\centering
\includegraphics[width=0.97\linewidth]{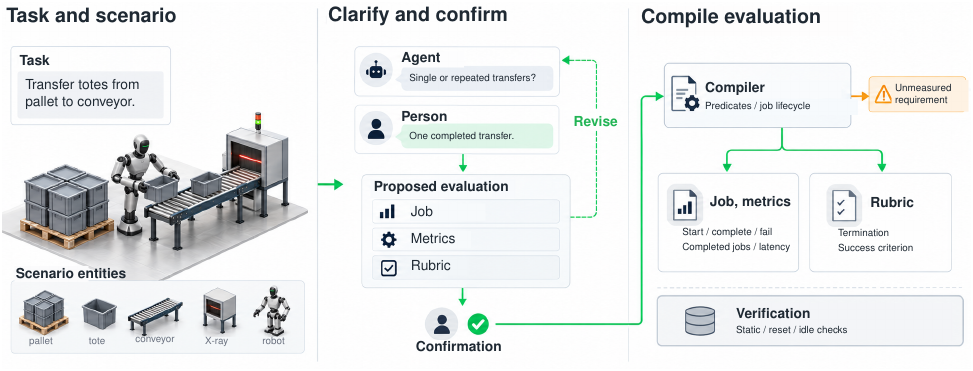}
\caption{Evaluation generation. The agent asks what the task and scenario leave undecided and proposes a job, metrics, and a rubric, which the person who wrote the task confirms. A compiler without a model turns the confirmed proposal into an evaluation, rejects what it cannot measure, and checks the result against the scenario.}
\label{fig:evaluation-generation}
\end{figure}

From the same task that drives scenario generation, a coding agent also generates the evaluation $E=(j,m_{1:N_m},c)$, so that each world $\xi=(S,E)$ of Eq.~\eqref{eq:agentic-world} is produced as a whole.
We split this generation into two stages (\autoref{fig:evaluation-generation}).
The agent proposes an interpretation of the task that the person who wrote it confirms, and a compiler with no model in the loop turns the confirmed proposal into executable checks that are tested against the scenario before the world enters the benchmark.

The agent composes $E$ from the predicates and metrics of \S~\ref{sec:evaluation} and cannot add new ones.

A generated reward can be checked against a separately defined performance metric~\citep{ma2024eureka,xie2024text2reward,yu2023l2r}.
A generated evaluation defines how performance is assessed, so it can compile and run while applying criteria that do not match the intended task.

We distinguish two outcomes.
A generated evaluation passes verification when it loads and composes as an executable evaluation and then passes the three checks of \S~\ref{sec:eval-compilation} against its scenario.
Its verdicts agree with the intended task when they match labeled recordings, which verification cannot show.
\S~\ref{sec:eval-study} measures both for evaluations generated with and without questions.

\subsubsection{Proposal and confirmation}
\label{sec:eval-proposal}

The agent first reads the task and the scenario and asks about what they leave undecided.
A question is allowed only if its answer changes what is measured or what counts as passing, and only if the scenario does not already answer it.
Typical questions concern what completes a job, which tolerance applies, which outcomes count as failure, how long a job may take, and whether the work repeats.
The agent then reviews the answers and may ask follow-up questions.

The agent proposes an evaluation consisting of a job, its metrics, and a rubric. The rubric specifies when a run ends and what it reports. It may also define a success criterion.

Tasks within a workplace can share a job definition and metrics while using different rubrics.
At the inbound conveyor station, single tote transfer specifies a run-level success criterion, whereas conveyor loading reports metrics without assigning a verdict.
Both count jobs consistently, allowing comparison across runs using their metric distributions.
\autoref{tab:eval-example} illustrates the questions, answers, and evaluation proposals for both tasks.

\begin{table}[t]
\centering\footnotesize
\caption{Evaluation generation for two tasks at the inbound conveyor station.
Both tasks share a job and metrics but use different rubrics.
Answers are italicized.}
\label{tab:eval-example}
\setlength{\tabcolsep}{4pt}
\renewcommand{\arraystretch}{1.15}
\begin{tabularx}{\linewidth}{@{}>{\raggedright\arraybackslash}p{0.18\linewidth}X@{}}
\toprule
\tablehead{Step} & \tablehead{Content}\\
\midrule
Task 1
& \emph{Single tote transfer}: ``Pick up the tote presented on the pallet and place it upright on the running conveyor.''\\
Task 2
& \emph{Conveyor loading}: ``Pick up each tote presented on the pallet and place it upright on the running conveyor.''\\
\midrule
Question 1
& Does a job complete upon placement on the belt or entry into the X-ray inlet?
\emph{Entry into the inlet.}\\
Question 2
& What tilt is allowed?
\emph{At most $10^\circ$.}\\
Question 3
& When does a run end?
\emph{After one job or 1\,min for Task 1, and after 5\,min for Task 2.}\\
\midrule
Shared job
& Starts when a tote is presented in the pick region, requires upright placement supported by the conveyor, and completes upon entry into the X-ray inlet.
All conditions use simulator state.\\
Shared metrics
& Completed and failed jobs, throughput, latency, stalls, protocol violations, upright placements, and inlet entries.\\
\midrule
Rubric 1
& Ends after one job or 1\,min.
Passes with one completed job and no protocol violations.\\
Rubric 2
& Runs for 5\,min without a success criterion.
Reports completed jobs and throughput.\\
\bottomrule
\end{tabularx}
\end{table}

The evaluation reads the full privileged state, but a task can require more than that state expresses.
Each condition drawn from the task is therefore labeled by what can judge it. It is \emph{physics} when a predicate on privileged state measures it, \emph{visual} when only a camera can show it, and \emph{unmeasured} when neither can.
These labels show the person where the evaluation falls short of the task, instead of hiding the gap behind a plausible check.
The proposal also marks which names of the scenario the policy may be given. Names that appear in the task, such as a tote, are visible to the policy, and all others, such as a target region, are reserved for the evaluator.

The person confirms the proposal as a whole for one scenario variant, and a change to the scenario requires a new confirmation.

\subsubsection{Compilation and verification}
\label{sec:eval-compilation}

A deterministic compiler converts the confirmed proposal into an executable evaluation without model calls or simulator rollouts.
It resolves scenario references, instantiates supported predicates and events, binds them to job start, completion, and failure conditions, and assembles the specified metrics and rubric.
The confirmed proposal and model outputs are retained for recompilation.
Only simulator-state predicates can determine success.
Visual requirements produce event-bound evidence records for post-run assessment (\S~\ref{app:visual-evidence}), but cannot enter success criteria.
Unsupported or unmeasured requirements block compilation rather than being replaced by proxies (\S~\ref{app:eval-rejections}).

Compiled evaluations undergo three checks against their scenarios without requiring a policy or human intervention (\autoref{tab:eval-checks}).
Prerequisite guards check loading, scenario references, and time limits.
The static check examines the executable definitions, including those edited by hand.
The reset check tests initial job conditions over 4\,s of uncontrolled physics in each scene variant.
The idle check runs the evaluation without robot control to detect jobs completed by the scenario alone.
For example, widening the conveyor-loading start region produces four completed jobs in an idle run, which the other checks do not reject.
However, all three checks can accept an incorrect upright tolerance because they do not assess task intent (\S~\ref{app:eval-verification}).

Failures return diagnostic feedback to the agent for revision, with at most three candidates including the initial attempt.
In question-guided generation, the agent revises the proposal for renewed human confirmation, or simulated-user confirmation in our study, while retaining previous answers.
In direct generation, it revises the executable evaluation.
Because the loop receives no labeled trajectories or verdict feedback, passing verification does not establish agreement with task intent.
Unresolved failures terminate the loop without an accepted evaluation.

\begin{table}[t]
\centering
\caption{Three policy-independent checks for compiled evaluations.
Reset and idle checks run in every scene variant without robot control.}
\label{tab:eval-checks}
\footnotesize
\setlength{\tabcolsep}{4pt}
\renewcommand{\arraystretch}{1.2}
\begin{tabularx}{\linewidth}{@{}l p{0.19\linewidth}YY@{}}
\toprule
\tablehead{Check}
& \tablehead{Execution}
& \tablehead{Rejects when}
& \tablehead{Does not detect} \\
\midrule
Static
& No simulation
& Invalid definitions, unjustified event rules, or time limits beyond the run
& Incorrect regions or tolerances \\
\addlinespace
Reset
& 4\,s after reset
& Expected job units fail to satisfy or maintain their start conditions
& Overly broad start regions or later state changes \\
\addlinespace
Idle
& Rollout to termination
& Jobs complete without robot action, the run passes, or start events recur during open jobs
& Errors requiring policy actions or task-intent labels \\
\bottomrule
\end{tabularx}
\end{table}

The three checks establish consistency with the scenario while they do not determine whether an evaluation captures task intent. The following study assesses this agreement using labeled recordings (\S~\ref{sec:discussion-eval-feedback}).

\subsubsection{Questions and verification feedback}
\label{sec:eval-study}

We evaluate single tote transfer (\autoref{tab:eval-example}) under four conditions, comparing direct evaluation generation with question-guided generation, each with and without verification feedback.
In question-guided generation, a simulated user answers clarification questions and confirms proposals using a fixed task-intent specification.
Four models perform three trials per condition, with up to three candidates when feedback is enabled. Evaluations are assessed against 15 labeled recordings withheld from generation and revision (\S~\ref{app:eval-generation-protocol}).

\begin{figure}[t]
\centering
\includegraphics[width=\linewidth]{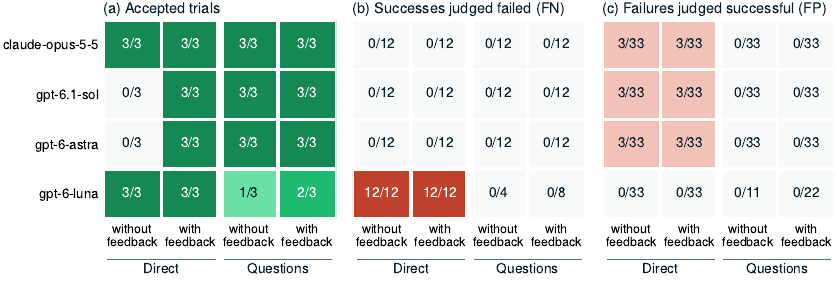}
\caption{Evaluations generated for single tote transfer, three trials per model and condition. (a) Trials whose evaluation passes the guards and the three checks on the first scene variant. (b) Successful recordings judged failed, out of four per executable evaluation. (c) Failed recordings judged successful, out of eleven per executable evaluation. Panels (b) and (c) include evaluations that verification rejects.}
\label{fig:eval-generation-study}
\end{figure}

Verification feedback increases direct evaluation acceptance from 6/12 to 12/12 but does not correct any of the 21 misclassified recording pairs (\autoref{fig:eval-generation-study}).
These errors arise from ambiguous completion criteria and additional conditions imposed by the models.
Question-guided generation recovers the intended completion event and uprightness tolerance, achieving agreement with all assessed labels.
It produces 10/12 accepted evaluations without feedback and 11/12 with feedback.

These results distinguish executable correctness from agreement with task intent.
However, the study covers one task, one scene variant, and 15 reused recordings without isolated violations of individual requirements.
The simulated user does not establish the reliability of human confirmation, and question-guided generation changes elicitation, review, and compilation together, preventing attribution of the improvement to any single component.

\FloatBarrier
\Needspace{5\baselineskip}
\section{\benchmark}
\label{sec:tasks}

\subsection{Tasks}
\label{sec:task-list}

The seven tasks cover the industrial work that robots already do and that employs many people (\autoref{tab:bench-tasks}).
BEHAVIOR-1K chooses its household activities by surveying people~\citep{li2024behavior1k}, but no such survey exists for factory work, so the selection follows robot deployment and employment.
Handling and assembling are among the largest applications of industrial robots~\citep{ifr2024pressconference}, the matching occupations employ millions of people in each major robot market (\S~\ref{app:task-selection}) and are highly exposed to automation~\citep{frey2017future}, and robotics research targets the same work~\citep{pantano2022skills,correll2018apc,nist2025assembly}.

Conveyor loading and flow-rack loading are repeated pick and place, and the other tasks add whole-body pushing of a loaded cart, filling and exchange of a fixture, a press started through a two-hand control~\citep{iso2019twohand}, process equipment that the robot operates and waits on, as mobile robots do in chemistry laboratories~\citep{burger2020mobile}, and routing by a label that the robot reads.
Welding, dispensing, and processing are established robot applications~\citep{ifr2024pressconference} but are left for a future benchmark.
The generators build further workplaces and tasks, such as the CNC tending cell of \S~\ref{sec:scene-generation}, and these sources decide which tasks enter the benchmark.

\begin{table}[t]
\centering\footnotesize
\setlength{\tabcolsep}{4pt}
\caption{The seven tasks in \benchmark. A job is the unit of work that the evaluation counts, and the carried state is what earlier jobs leave for later ones. Robot application is the matching class of industrial robot applications~\citep{ifr2020classification}, and occupation is the matching group of workers.
The task instructions are in \S~\ref{app:task-specs}.
}
\label{tab:bench-tasks}
\begin{tabular*}{\linewidth}{@{\extracolsep{\fill}}llllll@{}}
\toprule
\tablehead{Family} & \tablehead{Task} & \tablehead{One job} & \tablehead{Carried state} & \tablehead{Robot application} & \tablehead{Occupation}\\
\midrule
Material flow & Conveyor loading & Tote, pallet $\to$ belt & Totes on pallet, belt & Picking, placing & Material movers\\
 & Flow-rack loading & Container, stack $\to$ lane & Filled lanes, stack & Picking, placing & Material movers\\
 & Cart push & Carton, table $\to$ dock & Cart loads, docks & Transport$^{\dagger}$ & Material movers\\
Assembly & Jig loading & \shortstack[l]{Part in stored jig;\\empty jig seated} & Filled jigs & Assembling & Assemblers\\
Machine tending & Press tending & \shortstack[l]{Stamped part\\on outfeed} & Press, stamped parts & Press tending & Press operators\\
Process operation & Valve operation & Beaker, spout $\to$ output & Valve, filled beakers & -- & Plant operators\\
Inspection & Barcode routing & Parcel $\to$ coded tote & Routed parcels & Inspection, triage & Sorters\\
\bottomrule
\end{tabular*}
\par\smallskip
{\scriptsize\raggedright $^{\dagger}$Service robots for transport and logistics~\citep{ifr2026servicerobots}.\par}
\end{table}

Every task runs on persistent state.
No scenario restores its scene between jobs, and finished units stay in it unless the process logic removes them.
Process logic acts in four tasks: belts move units in conveyor loading, flow-rack loading, and barcode routing, queues return or restock units in the first two, and events drive the press in press tending and stop and restart the belt in barcode routing, while cart push, jig loading, and valve operation start from a fixed set of units.
A run has a maximum duration $T_{\mathrm{run}}$ and may end earlier under its rubric, for example when a target number of jobs is completed or an inactivity limit is reached. Completed-job counts and elapsed time are measured outcomes. Evaluation uses the resulting trajectory regardless of the policy's internal structure.

\begin{figure}[t]
\centering
\begin{subfigure}[t]{0.243\linewidth}
  \includegraphics[width=\linewidth]{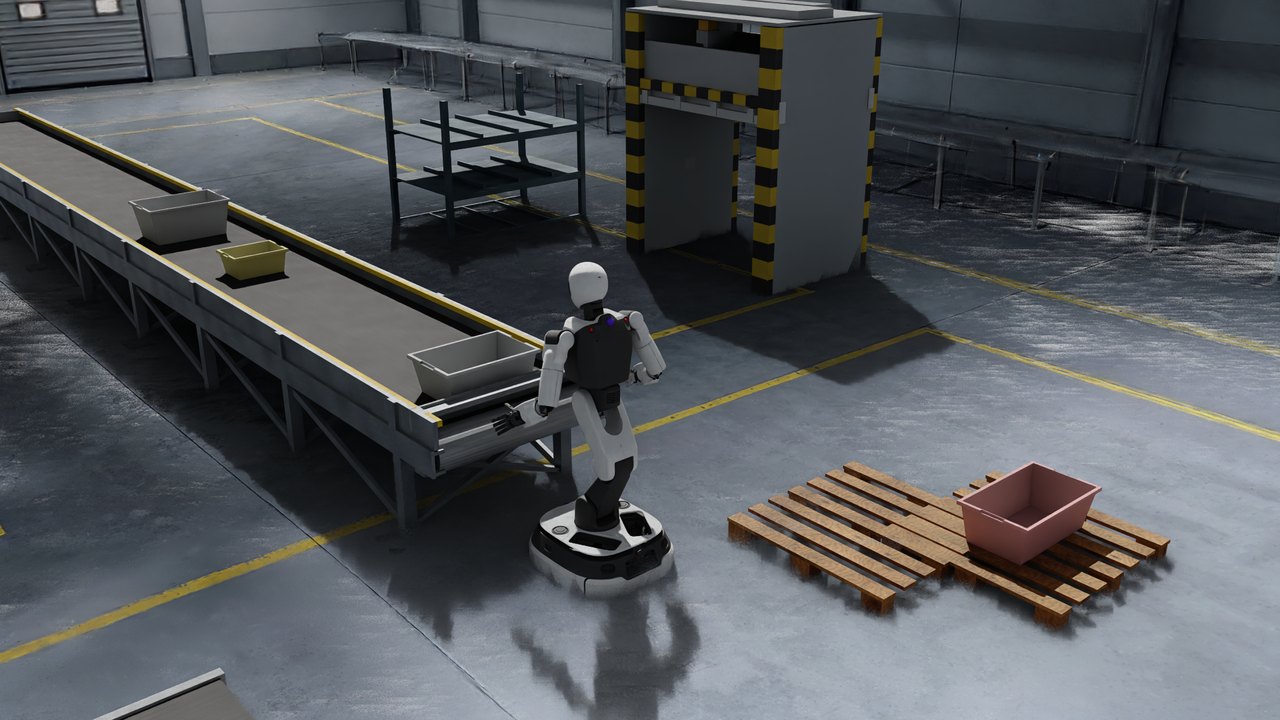}
  \caption{Conveyor loading}
\end{subfigure}\hfill
\begin{subfigure}[t]{0.243\linewidth}
  \includegraphics[width=\linewidth]{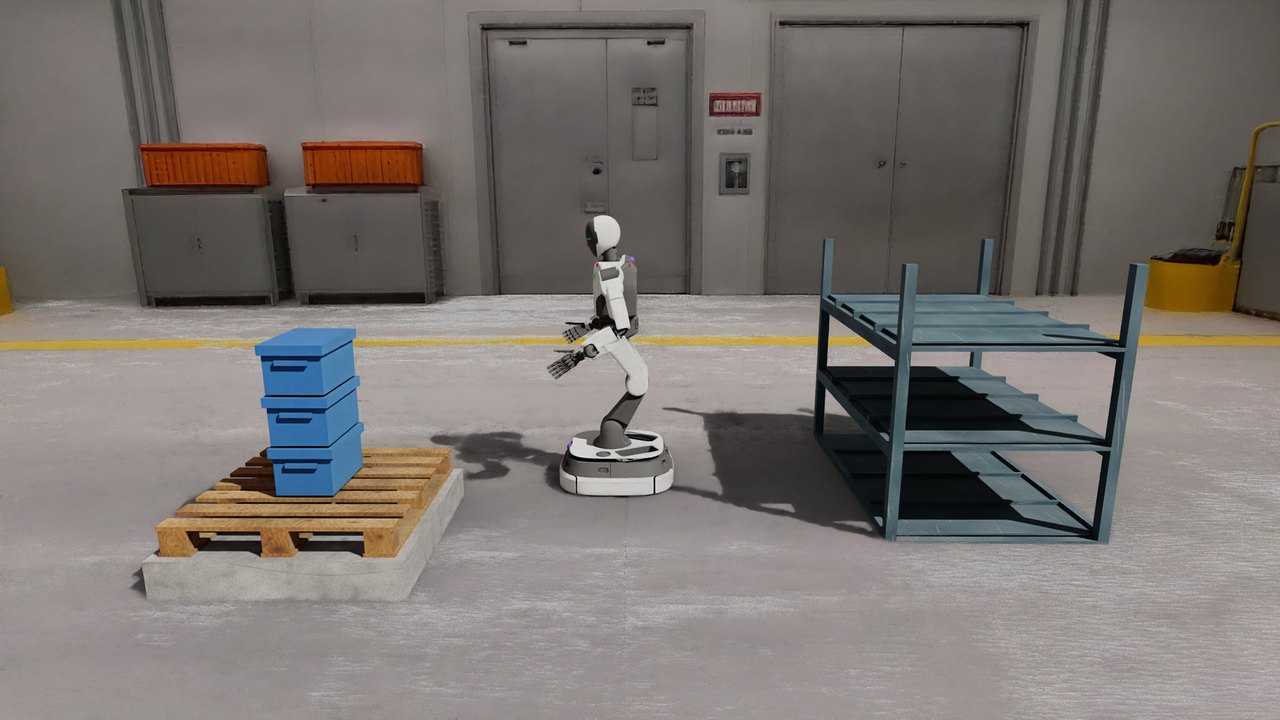}
  \caption{Flow-rack loading}
\end{subfigure}\hfill
\begin{subfigure}[t]{0.243\linewidth}
  \includegraphics[width=\linewidth]{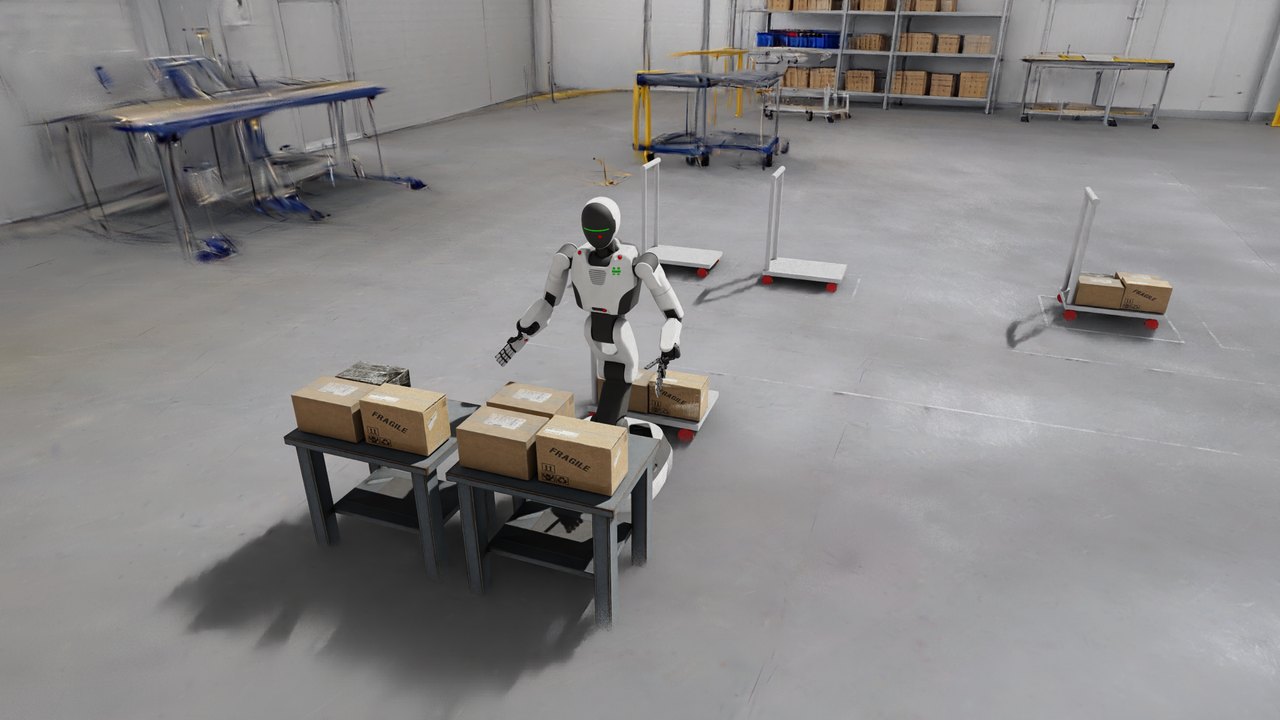}
  \caption{Cart push}
\end{subfigure}\hfill
\begin{subfigure}[t]{0.243\linewidth}
  \includegraphics[width=\linewidth]{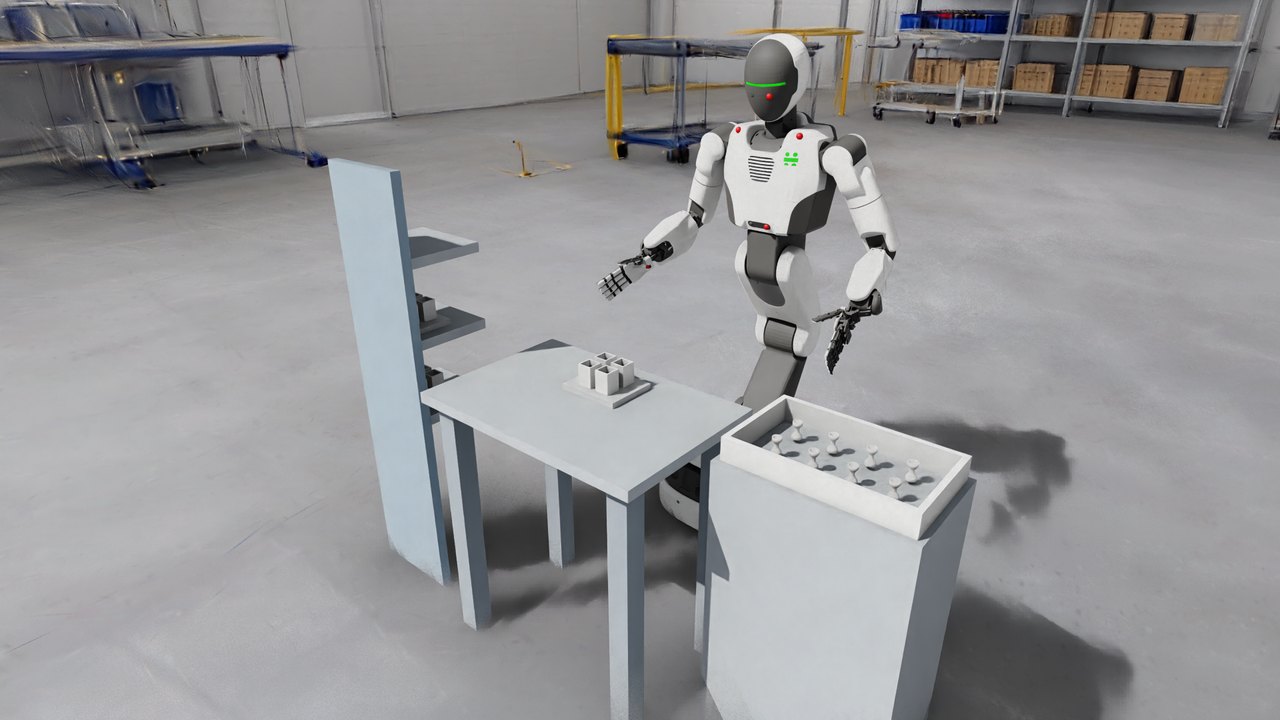}
  \caption{Jig loading}
\end{subfigure}

\vspace{2mm}
\begin{subfigure}[t]{0.243\linewidth}
  \includegraphics[width=\linewidth]{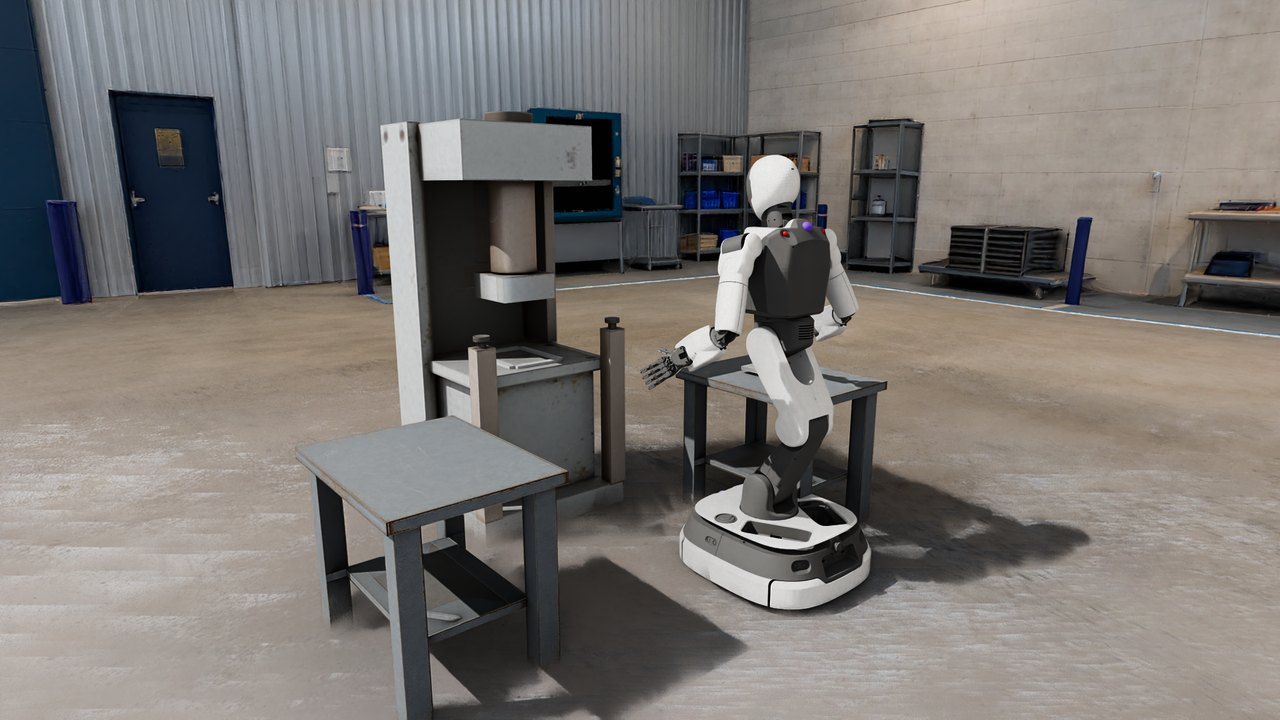}
  \caption{Press tending}
\end{subfigure}\hspace{0.0093\linewidth}
\begin{subfigure}[t]{0.243\linewidth}
  \includegraphics[width=\linewidth]{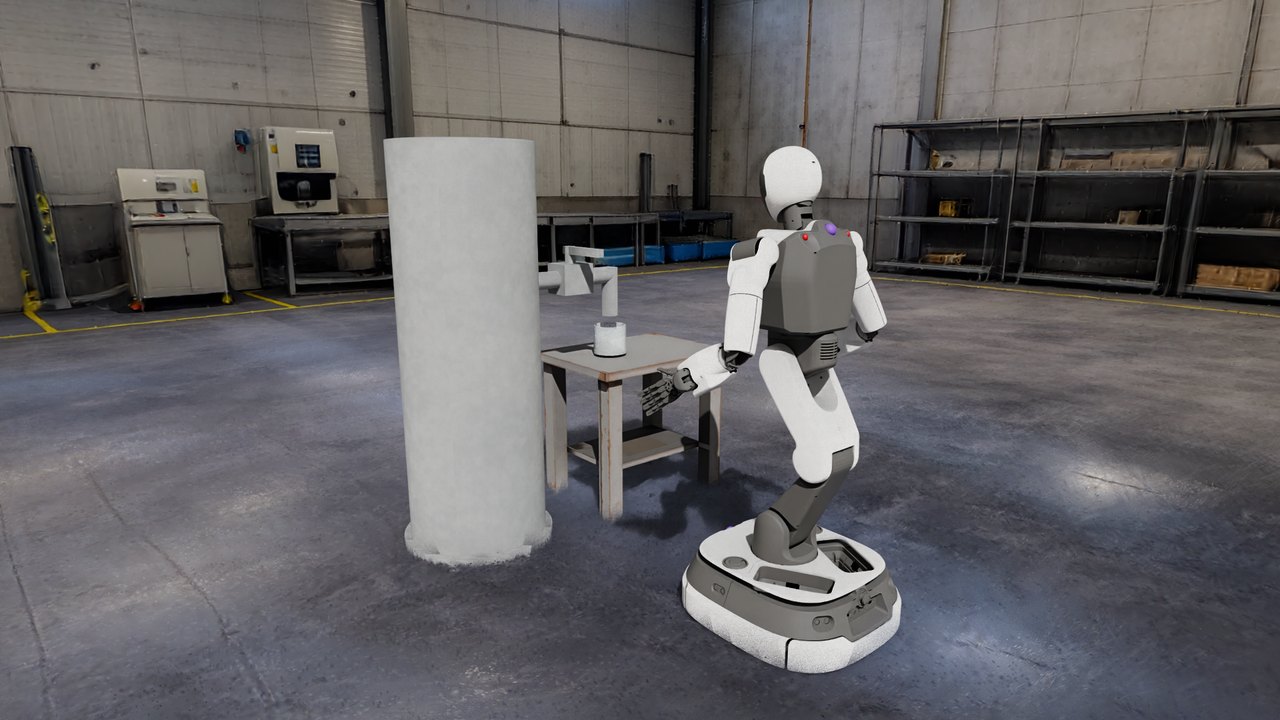}
  \caption{Valve operation}
\end{subfigure}\hspace{0.0093\linewidth}
\begin{subfigure}[t]{0.243\linewidth}
  \includegraphics[width=\linewidth]{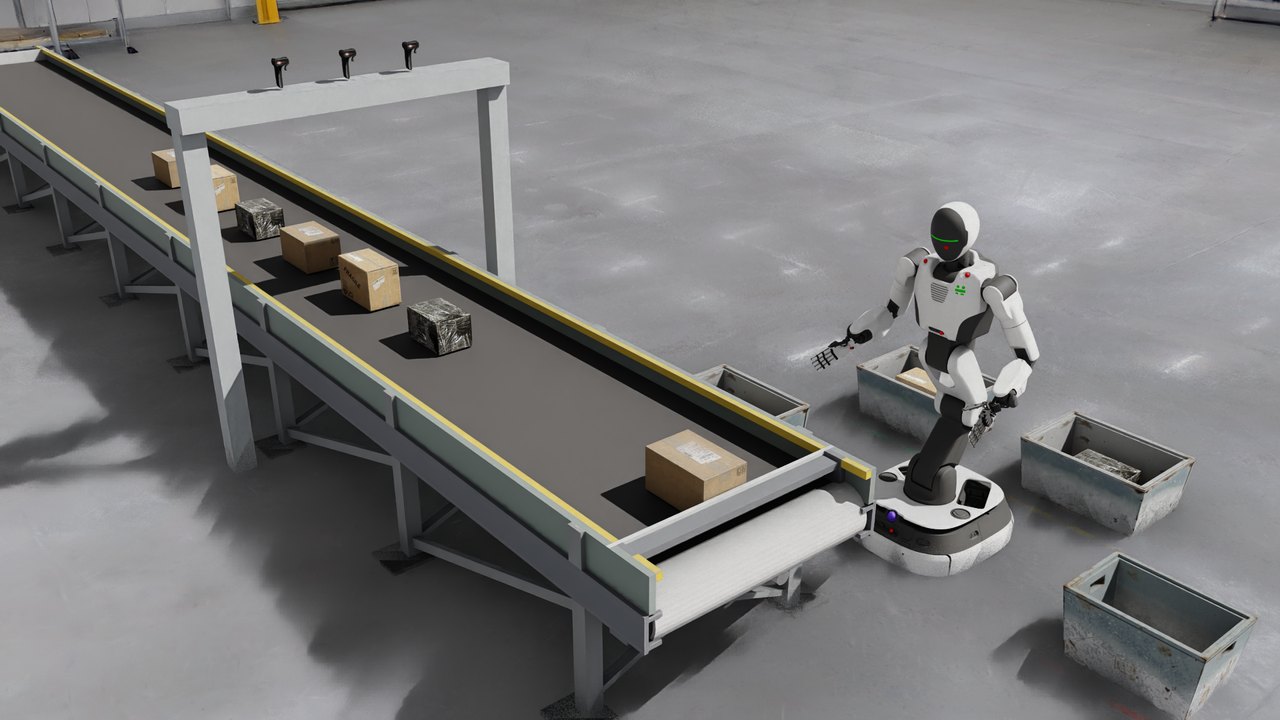}
  \caption{Barcode routing}
\end{subfigure}
\caption{The seven benchmark tasks in photoreal factory halls with \friday. (a--c) Material flow. (d) Assembly. (e) Machine tending. (f) Process operation. (g) Inspection.}
\label{fig:bench-tasks}
\end{figure}

The scenario representation supports scene variants that differ in the mass, placement, or appearance of their units and fixtures (\S~\ref{sec:world-generation}).
The scenes are generated from each task and its workplace in photorealistic factory halls (\autoref{fig:bench-tasks}), those of cart push, jig loading, and barcode routing by inverse scene generation (\S~\ref{sec:scene-generation}). Process logic specifies equipment behavior and unit supply (\S~\ref{sec:process-logic}).

\subsection{Evaluation}
\label{sec:evaluation}

Policies act on observations $o_t$, whereas \benchmark determines job outcomes and any run-level verdicts from privileged simulator state $x_t$.
Observation-based assessments, such as visual judgments, provide supplementary evidence but do not affect the current benchmark outcomes.
An evaluation $E=(j,m_{1:N_m},c)$ defines the job $j$ by events on privileged state, the metrics $m_{1:N_m}$ aggregated from job records, and the rubric $c$, which sets when a run ends, what it reports, and, optionally, which outcomes count as success (\autoref{tab:nomenclature}).
Evaluations of the same task can share a job and metrics while differing in the rubric, for example one that requires a single transfer within a time limit and one without a success criterion that counts the transfers of a fixed interval.

\begin{figure}[t]
\centering
\includegraphics[width=\linewidth]{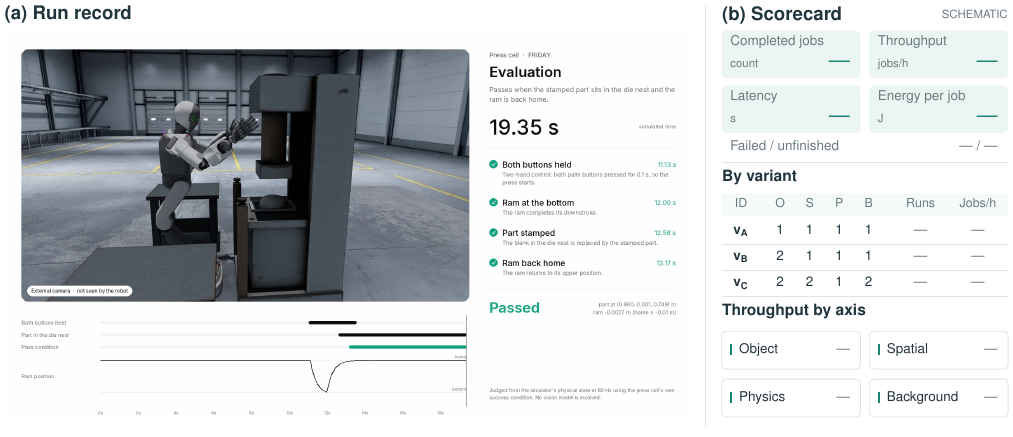}
\caption{Run evidence and variant-based scorecard.(Left) Press-cycle record with state-based events at 60\,Hz, showing a cycle-level verdict rather than the full outfeed-task outcome. (Right) Schematic scorecard with aggregate metrics, per-variant results, and condition-axis summaries.
O, S, P, and B denote object, spatial, physics, and background conditions.
}
\label{fig:eval-interface}
\end{figure}

\subsubsection{Jobs}
\label{sec:eval-jobs}

A job is an occurrence of work.
In conveyor loading, a tote that arrives on the pallet starts a job, which completes when the tote, upright on the belt, enters the X-ray tunnel, and because scanned totes return to the pallet, the same tote takes part in successive jobs.
Jobs are delimited by events on privileged state.
A predicate is a true-or-false test of the instantaneous state, such as a unit inside a region, upright within a tilt bound, at rest, or supported by another body, or a joint within a range (\autoref{tab:predicates}), and its truth value is determined by thresholding a satisfaction margin at zero, following the robustness semantics of temporal logic~\citep{fainekos2009robustness,donze2010robust}.
An event occurs when a predicate has held for $K_{\mathrm{hold}}$ consecutive steps, the bounded \emph{always} operator of signal temporal logic~\citep{maler2004stl}, so that contact noise does not trigger it,
\begin{equation}
  r_t=b_t\,\big(r_{t-1}+1\big),\qquad e_t=\mathbf{1}\big[r_t=K_{\mathrm{hold}}\big],\qquad r_{-1}=0,
  \label{eq:debounce}
\end{equation}
for the predicate value $b_t=b(x_t)\in\{0,1\}$, evaluated for all $N_{\mathrm{env}}$ simulation instances in parallel.

A job completes at its completion event once its required events have occurred since its start (\autoref{tab:task-events}).
It fails at a failure event, which overrides a completion in the same step, times out at its limit $\tau_{\mathrm{job}}$, or is censored if the run ends first.
A start event for a unit whose job is still open is a violation, as is an event outside any job, such as a hand inside the die while the ram is away from home.

\begin{table}[t]
\centering\footnotesize
\caption{Job events over the seven tasks. 
}
\label{tab:task-events}
\begin{tabular*}{\linewidth}{@{\extracolsep{\fill}}llll@{}}
\toprule
\tablehead{Task} & \tablehead{Start} & \tablehead{Required} & \tablehead{Completion}\\
\midrule
Conveyor loading & On pallet & Upright on belt & Entered tunnel\\
Flow-rack loading & In stack & In aisle & Upright in lane\\
Cart push & Staged & Left table, crossed aisle & In dock, loading deck clear\\
Jig loading & In tray & Left tray, in bench jig & Jig stored, empty jig seated\\
Press tending & On infeed & Nested, two-hand, stamped & Part on outfeed\\
Valve operation & On input side & Filled & On output side, valve shut\\
Barcode routing & Scanned & Off belt end & In destination tote\\
\bottomrule
\end{tabular*}
\end{table}

\subsubsection{Metrics}
\label{sec:eval-metrics}

We measure sustained performance through throughput, job outcomes, latency, and motor energy.
Throughput is the number of completed jobs per simulated hour, $3600\,N_{\mathrm{done}}/t_{\mathrm{run}}$, where $t_{\mathrm{run}}$ is measured in seconds, as in bin picking~\citep{mahler2019dexnet4}.
Policy comparisons hold the evaluation and work supply fixed.

Completion latency is reported alongside failed, timed-out, and censored jobs.
Jobs still open at the end of a run are right-censored~\citep{kaplan1958nonparametric}, while failures are competing outcomes~\citep{fine1999proportional}.
Reporting latency only for completed jobs can favor policies that finish easier jobs.

\hdaynewton measures electrical power at the motor terminals of each actuator.
We report drawn and regenerated energy, peak drawn power, and drawn energy per completed job.
The latter includes energy spent on failed attempts and is undefined when no job completes.
These measures exclude computing and sensing loads and are reported separately from throughput.

\subsubsection{Rubrics}
\label{sec:eval-rubrics}

A rubric specifies run termination, reporting, and optional success criteria.
Conveyor loading reports performance over a fixed interval without a success criterion.
For rubrics requiring $N_{\mathrm{goal}}$ completed jobs, a run terminates upon reaching the target, the time limit $T_{\mathrm{run}}$, or a specified inactivity limit.
A run passes only if it reaches the target without failures, timeouts, stalls, or violations (Eq.~\eqref{eq:run-acceptance}).
Passing reflects only the encoded predicates, which may miss placement defects outside their specified tolerances.

\paragraph{Verification with scripted trajectories}
We further test the evaluations of six tasks, excluding conveyor loading, using 185 region checks, 20 start-state probes, and 32 scripted trajectories with predefined outcomes.
All tests match their expected results after correcting spurious start events during object transport (\S~\ref{par:eval-scripted}).
These results establish agreement on the tested cases, not complete task coverage or policy performance.

\subsection{Evaluation across variants}
\label{sec:eval-regimes}

Evaluation is organized around scene variants, each specifying a complete task configuration.
The placements within a variant encode relationships established during scene construction. Moving a fixture may require corresponding changes to part placement, support, access space, and target regions.
Using the variant as the evaluation unit retains these coordinated choices, subject to the verification limits of \S~\ref{sec:scene-generation}.

An evaluation regime declares the variation axes and conditions represented by its variants, such as object identity, spatial configuration, physical properties, and visual surroundings~\citep{xie2024generalization,pumacay2024colosseum}.
Outcomes are grouped by these axes and conditions and aggregated across the evaluated set.
Background-only variants share the dynamics $F_{s,\ell}$ and differ in the observation model $O_s$ (\S~\ref{sec:photoreal-background}).
Isolating one axis requires matched variants that hold the other conditions fixed.

Policies should be compared on the same declared variants, with initial conditions and randomization settings drawn from the same distribution~\citep{snyder2026beyond}.
Repeated runs provide observations within each variant, and the comparison specifies how variants are sampled and weighted in aggregate results.
Jobs within a run share the state left by earlier jobs, so the run is the unit of statistical analysis~\citep{agarwal2021precipice,cameron2015practitioner}.
Repeat counts and uncertainty estimates are reported alongside job outcomes, throughput, latency, and motor-terminal energy.
For a fixed evaluation, physical outcomes can be recomputed from a stored trajectory (\autoref{fig:eval-interface}).

\Needspace{5\baselineskip}
\section{Discussion and open questions}
\label{sec:discussion}

\subsection{Value of richer physical simulation}

Richer physical models allow a benchmark to capture contact, actuator, and sensor phenomena that simpler models omit. 
However, their value depends on how well they reproduce real hardware behavior, whether they improve policy comparisons, and whether the additional physical information can be used to evaluate the policy.

\paragraph{How accurately do the models reproduce real hardware?}
This paper examines the contact, actuator, and sensor models against their specified behavior and does not validate them against hardware measurements~\citep{oberkampf2010verification}.
The fingertip studies compare with quadrature of the same pressure law (\S~\ref{par:fingertip-patch-study}), and the wheel study with ideal differential-drive kinematics (\S~\ref{par:wheel-ground-study}).
The battery comparison and the finger-rolling and tactile examples show behavior under the stated settings and have no independent reference (\S~\ref{sec:simulation}).
No result has been compared with measurements from a physical robot.

\paragraph{Which physical details change rankings?}
Contact models, actuator dynamics, time steps, and solver resolution can affect task outcomes, but their influence on policy rankings remains unmeasured and under-explored.
The central question is therefore which physical phenomena and levels of numerical accuracy are necessary to reproduce policy rankings observed on real hardware, and which can be simplified without changing those rankings.

\paragraph{What distinguishes gentle from damaging handling?}
Job completion and motor energy do not fully capture how objects are handled.
A part may be placed gently or dropped into a fixture while satisfying the same success criteria.
Contact-derived measures such as impact impulse, approach speed, peak pressure, accumulated slip, and transverse insertion force could distinguish these behaviors when measured for individual parts and job phases.
Force and pressure limits are used in human--robot contact safety~\citep{haddadin2009requirements,iso2016ts15066}, while damage-aware manipulation considers object integrity alongside grasp success~\citep{xie2024deligrasp}.
However, these quantities do not predict damage without material-specific thresholds, and their values can depend on numerical resolution.
Which measures can reliably distinguish acceptable from damaging handling without penalizing forces required by the task?

\subsection{Limits of automated verification}
\paragraph{What should decide scene acceptance?}
Asset-level verification establishes whether an asset satisfies its specified requirements, but not whether it supports the intended interaction once placed in a scene. Scene-level checks introduce additional constraints on placement and spatial relations, yet they do not establish whether the intended interactions are feasible, including whether the robot can reach and grasp the objects.
Model critics provide additional judgments, yet the same candidates that pass deterministic checks receive different acceptance decisions depending on the critic (\autoref{fig:critic-swap}), a known weakness of model judges~\citep{zheng2023judging}.
It remains an open question to determine which aspects of scene suitability require additional evidence such as reachability analysis or rendered observations.

\paragraph{How can evaluations be validated with limited labels?}
A related limitation arises in evaluation generation. Even when an evaluation passes verification, its criteria may not faithfully reflect the intended operation.
In the evaluation generation study of \S~\ref{sec:eval-study}, feedback from verification raises acceptance without changing a single verdict, because the loop receives no labels, and the errors that remain come from what the instruction leaves unstated. These results rest on one task and fifteen recordings.
More generally, model judges can share errors because language-model errors are correlated across models~\citep{kim2025correlated}.
Success detection and dense progress estimates from vision-language models~\citep{du2023successvqa,ma2024gvl} face the same limit, because agreement among them is not evidence of correctness.

One possibility is to compare different forms of verifiers generated from the same instruction.
Beyond the predicates on simulator state in \benchmark, instructions can be expressed as temporal logic~\citep{cosler2023nl2spec,chen2023nl2tl,liu2023lang2ltl}, reward code~\citep{yu2023l2r,ma2024eureka,xie2024text2reward}, or a learned success detector or reward model~\citep{du2023successvqa,lee2026roboreward,liang2026robometer}.
These verifiers may fail in different ways, and disagreements among them could reveal misinterpreted requirements. However, disagreement alone does not establish which judgment is correct, and different verifiers may share the same errors. This suggests using verifier disagreement to identify executions worth labeling, and combining a limited number of independent labels with model judgments to assess the reliability of generated evaluations~\citep{angelopoulos2023prediction,kossen2021active}.

\subsection{Expressiveness and diversity of generated scenes}

\paragraph{Do generated scenes remain diverse?}
A generator that passes every check can still return the same few scenes.
The checks of \S~\ref{sec:scene-generation} require distinct layouts, while its family diversity is defined but not quantified here. Moreover, distinct layouts do not necessarily demand different capabilities from a policy.
Repeated revision may also favor layouts and assets that are easier to verify, potentially reducing diversity, although this effect has not yet been measured.
Quality-diversity search keeps an archive of good solutions that differ along chosen descriptors~\citep{mouret2015illuminating}, open-ended and unsupervised environment design generate environments that stay novel and solvable~\citep{wang2019poet,dennis2020paired}, and similarity kernels measure diversity without a reference set~\citep{friedman2023vendi}.
The challenge is to identify which variations in geometry, physical properties, and appearance meaningfully change what a policy must do, rather than merely how a scene looks.

\paragraph{Do scene variations predict real-world performance?}
The value of generated scene variants for policy evaluation depends in part on whether performance across them predicts performance in the real world~\citep{li2024simpler}.
This has not been exhaustively tested.
Scenes reconstructed from real cells~\citep{torne2024rialto,han2025re3sim} offer one way to test it, because scores on a reconstruction can be compared with trials in the cell it was built from.
Such comparisons could help determine which physical and visual variations are needed for evaluations on generated scenes to predict real-world policy performance.

\subsection{Evolving role of coding agents}

\paragraph{Which parts of the harness remain necessary as coding agents improve?}
The staged checks, typed specifications, and model critics in our pipeline are designed to guide coding agents through world generation, but their value may change as agents become more capable.
Prior work shows that agent interfaces and scaffolds affect performance on software tasks~\citep{yang2024sweagent} and in robot control~\citep{fu2026capx,berman2026embody}, and that a simple pipeline without an agent loop can match elaborate software agents~\citep{xia2025agentless}.
In inverse scene generation, a change in how the prompt states affordances helps more than a larger budget (\autoref{tab:plan-critic}).
These results establish the importance of pipeline design under the tested conditions, but not whether its individual components will remain necessary as coding agents improve. Some may continue to provide essential guarantees, while others may become unnecessary overhead or limit what stronger agents can achieve.

A related question is how much of the harness itself can be delegated to coding agents. For example, \eris infers the requirements of an asset from the task, but the language that states them and the check behind each of its terms are designed by hand.
Letting the generation agent add terms, measurements, and probes from a fixed set of primitives would let verification grow with the benchmark, but a requirement written by the agent that also generates the asset is not an independent check of it, much as language models favor their own outputs when they judge them~\citep{panickssery2024llm}.

\paragraph{How should the pipeline be tested?}
As discussed above, some parts of the harness may become unnecessary as coding agents improve, so its value must be shown again with every change rather than assumed.
The comparison of staged verification against its baselines (\autoref{fig:paradigms}) can serve as a fixed internal benchmark that every change of agent, prompt, or harness must pass, as software agents are compared on fixed task suites~\citep{jimenez2024swebench}.
Agents also change between versions, so a generated world is reproducible only with the agent and harness that generated it.
A small, fixed test suite could make these comparisons practical, provided it can distinguish improvements in the agent from improvements in the harness.

\paragraph{Can world generation evaluate coding agents?}
Coding agents are now evaluated on embodied work, as controllers that write programs for a robot~\citep{fu2026capx} or act through a visual interface~\citep{hu2026via}.
\benchmark adds two roles for them.
Its tasks can score a coding agent as the policy under test, with the same evaluations as any other policy, and generating a world from a task is itself a programming problem whose output passes or fails checks that need no labels (\S~\ref{sec:world-generation}).
Verified world generation could provide a separate measure of coding-agent capability, although its relationship to robot-control performance remains unexplored.

\subsection{Physics-grounded supervision for learned evaluators}
\label{sec:discussion-eval-feedback}
Reset-free execution provides information that isolated episodes do not capture. Job events reveal how work progresses across repeated operations, including stalls, failures, and their effects on subsequent jobs. These records can provide supervision for learned progress estimators, but they also complicate policy comparison because successive jobs share the same evolving physical state.

\paragraph{Can physics calibrate value functions?}
Learned models now serve as value functions that estimate progress from frames~\citep{ma2024gvl,budzianowski2025opengvl}, as reward models for reinforcement learning~\citep{zhang2025rewind,zhai2025vlac,lee2026roboreward,liang2026robometer}, and as general models that label progress for robot learning~(Gemini 4 Argon; \cite{rao2026argon}).
Most are trained or scored against progress taken from frame order or from human labels, but frame order assumes that progress grows with time, which fails for runs that stall or regress, and human labels are costly at the length of a shift.
Physics gives the progress of every job without labels, over runs long enough for stalls, regressions, and compounding errors to occur, so \benchmark can measure where an estimate drifts from completed work, as most whole-run judges of \S~\ref{par:progress-judges} do when they accept a tote that lies upside down.
Physics-defined progress could therefore provide supervision for learned value functions, although its usefulness in real cells without privileged simulator state remains to be established.

\subsection{Statistical reliability of reset-free evaluation}
\label{sec:discussion-statistics}

\paragraph{How many runs separate two policies?}
How reliably the benchmark can distinguish between policies remains unknown, because no policies have yet been compared on the seven tasks.
Jobs within one run depend on the state left by earlier jobs, so treating them as independent observations would underestimate uncertainty.
The interval estimates recommended for policy evaluation~\citep{agarwal2021precipice,kressgazit2024empirical} must therefore use runs as the statistical unit or account for dependencies among jobs within each run, although cluster-robust uncertainty estimates can be unreliable with few runs~\citep{cameron2015practitioner}.
Sequential tests can reduce evaluation cost by stopping once sufficient evidence is collected, for binary success~\citep{snyder2025better} and continuous outcomes~\citep{snyder2026beyond}, but must also account for this dependence.
Establishing reliable policy comparisons will require measuring run-to-run variability and how the precision of estimated policy differences depends on the number and duration of runs.

\Needspace{5\baselineskip}
\section{Conclusion}
\label{sec:conclusion}

We introduce \hdayfactory, a framework for constructing task-conditioned industrial worlds and evaluating repeated robot jobs under persistent scene state.
\hdaynewton extends Newton's SDF-based hydroelastic contact with constitutive laws and a semi-implicit primal-dual solver, couples stateful actuators to articulated dynamics, and provides simulated tactile and joint observations.
Verification-guided world construction generates and revises assets and scene variants, incorporates photorealistic backgrounds and process logic, and compiles human-confirmed evaluation proposals into job definitions, metrics, and rubrics.
\benchmark defines seven industrial tasks that measure job outcomes, throughput, latency, and motor energy without resetting the scene between jobs.

Controlled studies characterize physical and sensor responses, demonstrate the benefits of measured feedback in asset generation, and reveal the dependence of scene acceptance on model critics.
Evaluation-generation experiments further show that passing executable checks does not establish agreement with task intent.
Scripted trajectories test the benchmark's job-level evaluations, but policy comparisons over persistent runs remain untested.

These findings identify several open challenges.
The physical models require hardware validation, and their effects on policy rankings remain unknown.
Generated worlds must be assessed for task feasibility and meaningful diversity beyond executable validity.
Verification harnesses must also be evaluated independently of the coding agents they support.
Physics-derived job records offer potential supervision for learned evaluators, but their usefulness beyond simulation remains to be established.
Most importantly, reliable policy comparison requires accounting for dependencies among successive jobs and measuring uncertainty across repeated runs.

We believe simulation benchmarks should extend beyond household tasks to capture the repeated, state-dependent work of industrial robots.
We hope the methods, empirical findings, and open questions presented in this paper contribute to the open-source robotics ecosystem by informing the construction of industrial simulation environments, task-conditioned world generation, and persistent evaluation protocols.

\FloatBarrier
\clearpage
\section{Contributors and acknowledgments}
\label{sec:contributors}

\subsection{Contributors}
\label{sec:contributors-list}

Within each role, contributors are listed alphabetically by given name and then by family name, and the order does not indicate relative contribution.
\medskip

\contributors{Project Leads}{Jeongmin Lee, Taehyeon Kim}
\contributors{Core Contributors}{Dohyun Jang, Minji Lee, Seonghun Oh}
\contributors{Contributors}{%
Dongwoo Park, Geonhyup Lee, Jaehyung Kim, Jaemin Yoon, Jinseop Bok, Jungwoo Kim, Minseok Yang, Philsik Chang, Soomin Lee}

\subsection{Acknowledgments}
\label{sec:acknowledgments}

We thank ChanHee ``Luke'' Song (NVIDIA), Justin Yu (UC Berkeley), Minyoung Hwang (MIT), and Seungjae Lee (University of Maryland) for careful discussions and for their insights into recent developments in the field.

\begingroup
\small
\bibliographystyle{abbrvnat}
\bibliography{report/references}
\endgroup
\clearpage
\appendix
\small
\section{SDF-hydroelastic contact details}
\label{app:contact-implementation}
This appendix gives numerical details of the contact model and solver of \S~\ref{sec:contact-modeling} and~\ref{sec:contact-solver}, followed by the protocols of the contact studies and the response kernel of the magnetic tactile sensor of \S~\ref{sec:tactile}. Choices of collision query, field storage, and surface discretization affect cost and numerical approximation, but do not by themselves introduce additional material phenomena.

\subsection{Collision geometry and memory}
\label{app:sdf-queries}

Newton's sparse-grid SDF representation uses a coarse grid and packed high-resolution subgrids near the surface. Concentrating high-resolution storage near the surface avoids allocating a uniformly fine grid throughout the surrounding volume.

For SDF hydroelastic contact, hierarchical pruning identifies candidate cells, and marching cubes extracts the zero level set of the pressure difference. Surface quadrature and contact reduction then provide the contact constraints described in the main text. The cost of this path depends on overlap volume, grid resolution, and the number of generated surface elements.

SDF resolution trades geometric fidelity, including surface features and narrow clearances, against memory consumption.
\label{app:collision-memory}
Four mechanisms reduce collision memory in batched simulation.

\paragraph{Per-scenario collision budget}
A generated scenario also fixes how much collision memory one simulation instance needs.
Each asset already carries an SDF resolution chosen under a memory model when it is generated (\S~\ref{sec:eris-verification}), and the remaining memory requirements depend on the scenario. GPU simulation gains throughput by stepping many instances at once~\citep{taomaniskill3}, and collision buffers grow with the number of instances and of contact pairs.
We therefore size the collision buffers for the contacts that a scenario can produce at once rather than for every possible pair and store the sizes with the scenario. The batch size is then chosen from the resulting memory estimate before allocation, instead of shrinking the batch after an out-of-memory failure.
The stored sizes hold only if no buffer overflows during a rollout.
This per-scenario budget further reduces memory without losing contacts in rollouts that reach the grasp.

\paragraph{Shared structure across simulation instances}
Simulation instances that share the robot's joint structure may hold different scene variants. Identical meshes and catalog assets are stored once per batch, but a signed-distance field is shared only when instances reuse the same loaded asset.

\paragraph{Asset-level SDF budget}
Each part's signed-distance resolution is chosen when the asset enters the catalog, so that no single detailed asset dominates collision memory.
\hdaynewton sizes the buffer of every hydroelastic contact pair~\citep{elandt2019pressure} by the most finely resolved part in the scene, so one detailed part inflates the memory of all pairs. Compact field encodings trade memory against accuracy~\citep{ji2026tesdf}.
We keep the encoding, take the coarsest resolution between 32 and 1\,mm that represents the part, and refine further only while the part adds at most 1\,MB per instance. Parts above 500\,MB per instance are rejected. For assets generated by \eris, collision construction follows \S~\ref{app:eris-sdf}.

\paragraph{Instance-chunked collision}
Collision buffers are sized for a chunk of simulation instances rather than the whole batch, while the solver still steps all instances together.
In deterministic mode the contact set equals that of a single pass.

\subsection{Pressure law and solver discretization}
\label{app:solver-discretization}
This subsection gives the evaluation of the nonlinear pressure law of Eq.~\eqref{eq:nonlinear-pad-pressure} and the discretization behind the contact solver of \S~\ref{sec:contact-solver}.

\paragraph{Nonlinear pressure evaluation}
\label{par:nonlinear-pressure}
For contact-surface extraction, we extend the nonnegative interior pressure defined by Eq.~\eqref{eq:nonlinear-pad-pressure} outside the body as $p_A=-k_{h,A}\phi_A$ for $\phi_A\geq0$. 
This signed extension is continuously differentiable at the surface and remains monotone in signed distance. It is used to locate pressure-equilibrium crossings and does not introduce tensile contact forces. The nonlinear law is evaluated at voxel corners and quadrature locations, while marching-cubes edge interpolation still approximates the contact surface and requires adequate spatial resolution.

The compression-hardening preset of Eq.~\eqref{eq:nonlinear-pad-pressure} targets the SDF-SDF and SDF-rigid-plane paths. The reduced force-derived stiffness is held fixed within each dynamics step, so introducing nonlinear pressure does not make the time integration fully implicit in the nonlinear constitutive law.

\paragraph{Semi-implicit Hunt-Crossley damping}
To fold Hunt-Crossley damping into the implicit solve, we predict penetration as $\delta^{k+1}\approx\delta^k-hv_n^{k+1}$, with normal relative velocity $v_n$ positive on separation, and freeze only the penetration factor in the damping coefficient at its current value. For an active penetrating contact, with patch-derived stiffness $K$ fixed during the substep and before near-rigid regularization, the update is
\begin{equation}
\begin{aligned}
f_n^{k+1}
&= \left[K(\delta^k-hv_n^{k+1})-K\alpha\delta^k v_n^{k+1}\right]_+ \\
&= K\left[\delta^k-\tau_{\mathrm{contact}}^k v_n^{k+1}\right]_+,
\qquad \tau_{\mathrm{contact}}^k=h+\alpha\delta^k.
\end{aligned}
\label{eq:semi-implicit-hunt-crossley}
\end{equation}
Here, $h$ is the substep duration and $[x]_+=\max(x,0)$ enforces a nonnegative normal force. Freezing the penetration factor removes the quadratic velocity term that would arise from a fully implicit Hunt-Crossley product, yielding a monotone, piecewise-affine normal force law for fixed contact geometry. This retains penetration-dependent dissipation while allowing the normal response to be solved jointly with friction and articulated dynamics. Contacts with $\delta^k\leq0$ are inactive, so the velocity-dependent term is not applied before penetration.

\paragraph{Discrete friction law}
Tangential reactions and the normal forces that bound them remain coupled unknowns. At convergence, the discrete contact law satisfies
\begin{equation}
\|\mathbf{f}_{t,i}^{k+1}\|\leq\mu_i^k f_{n,i}^{k+1},
\qquad
\mathbf{f}_{t,i}^{k+1}\cdot\mathbf{v}_{t,i}^{k+1}
=-\mu_i^k f_{n,i}^{k+1}\|\mathbf{v}_{t,i}^{k+1}\|,
\label{eq:discrete-contact-friction}
\end{equation}
where $i$ denotes the contact index.
These conditions allow a nonzero sticking force at zero slip velocity and recover friction opposing the next-step motion during sliding. Because each contact samples the local tangential velocity, the same solve resolves both net sliding resistance and the frictional moment generated by twisting across the patch. Freezing the geometry and coefficient preserves the existing local Coulomb update without freezing the friction direction or the normal-force-dependent bound. The coefficient is refreshed at each substep, so rapid transitions between sticking and sliding remain sensitive to the time step. This is a lagged approximation to the continuous speed-dependent law, not a fully implicit treatment of speed weakening.

\paragraph{Near-rigid regularization}
A contact with material stiffness $K_i$ has a mass metric $m_{\mathrm{contact},i}$, a numerical estimate of the inertia behind the contact propagated along the load path, not its exact effective mass. The reference stiffness $K_{\mathrm{ref},i}=m_{\mathrm{contact},i}(\vartheta/h)^2$ relates the correction to a dimensionless threshold $\vartheta$. When $K_i>K_{\mathrm{ref},i}$, the effective stiffness is
\begin{equation}
K_{\mathrm{eff},i}=K_{\mathrm{ref},i}\left(\frac{K_i}{K_{\mathrm{ref},i}}\right)^{\chi_i},
\qquad \chi_i=\min\!\left(\frac{\delta_i}{\delta_{\mathrm{allow}}},1\right).
\label{eq:near-rigid-stiffness}
\end{equation}
Here, $\delta_{\mathrm{allow}}$ is the penetration allowance, and contacts below the reference stiffness remain unchanged. This geometric interpolation softens shallow contact while recovering the original stiffness for deeper compression, with additional damping accompanying the softening. Its mass metric is independent of the preconditioner settings, so numerical scaling does not redefine the contact response. Unlike preconditioning, this correction intentionally modifies the discrete material response near contact onset. 

Regularization modifies both the stiffness and the contact time scale in Eq.~\eqref{eq:semi-implicit-hunt-crossley}. In this paragraph, $\delta_i$ denotes the current penetration $\delta_i^k$, and the time index on the contact coefficients is omitted for brevity. We denote the regularized counterpart of $\tau_{\mathrm{contact},i}^k=h+\alpha\delta_i^k$ by $\tau_{\mathrm{eff},i}$ and define it, for a penetrating contact, as
\begin{equation}
\tau_{\mathrm{eff},i}=
\begin{cases}
\displaystyle
\max\!\left(h+\alpha\delta_i,\,
 h+2\left[\sqrt{\frac{m_{\mathrm{contact},i}}{K_{\mathrm{eff},i}}}
          -\sqrt{\frac{m_{\mathrm{contact},i}}{K_i}}\right]\right),
& K_i>K_{\mathrm{ref},i},\\[6pt]
h+\alpha\delta_i, & K_i\le K_{\mathrm{ref},i}.
\end{cases}
\label{eq:near-rigid-timescale}
\end{equation}
The second argument of the maximum imposes a damping floor associated with softening and increases the time scale only when it exceeds $h+\alpha\delta_i$. With these regularized coefficients, the normal contact law becomes
\begin{equation}
\begin{aligned}
f_{n,i}^{k+1}
&=\left[K_{\mathrm{eff},i}\left(\delta_i^k-\tau_{\mathrm{eff},i}v_{n,i}^{k+1}\right)\right]_+\\
&=\left[K_{\mathrm{eff},i}\left(\delta_i^k-hv_{n,i}^{k+1}\right)
-K_{\mathrm{eff},i}(\tau_{\mathrm{eff},i}-h)v_{n,i}^{k+1}\right]_+.
\end{aligned}
\label{eq:regularized-normal-contact}
\end{equation}
The first term is the elastic response evaluated at the predicted next-step penetration, and the second is the damping term. Its coefficient is therefore $K_{\mathrm{eff},i}(\tau_{\mathrm{eff},i}-h)$. When regularization is disabled, $K_{\mathrm{eff},i}=K_i$ and $\tau_{\mathrm{eff},i}=h+\alpha\delta_i^k$, recovering Eq.~\eqref{eq:semi-implicit-hunt-crossley}. Nonpenetrating contacts remain inactive.

\subsection{Contact and solver studies}
\label{app:contact-study-protocols}

Four studies follow. Each compares the engine with a numerical reference of the same law or with another backend as configured, and none compares with hardware measurements.

\paragraph{Fingertip patch friction}
\label{par:fingertip-patch-study}
\autoref{fig:fingertip-patch-friction}(a,d) uses the right-index distal pad of the \friday hand model, isolated from the articulated hand and rigidly reoriented toward a plane without rescaling. Only the pad participates in contact, and the distal housing is visual context. The pad uses $k_h=10^8\,\mathrm{N}/\mathrm{m}^3$, $\mu=0.6$, a 0.5\,mm SDF voxel against the asset default of 2\,mm, 0.25\,mm quadrature spacing, moment-matching contact reduction, $\alpha=10^{-6}\,\mathrm{s}/\mathrm{m}$, deterministic contact generation, and 128 solver iterations without warm starting or near-rigid regularization. The pressure law is linear, $p=k_h[-\phi]_+$, and the Coulomb coefficient is constant. At indentations of 0.25, 0.5, and 1\,mm, direct signed-distance queries against the original triangles followed by 0.05\,mm planar quadrature give normal loads of 3.39, 7.84, and 18.81\,N. Pure rotation at 0.1\,rad/s is prescribed about the normal axis through each reference pressure center. Integrating the local sliding tractions gives reference effective radii of 4.71, 5.07, and 5.57\,mm.

\hdaynewton evaluates one dynamics step of 1\,ms from each prescribed state. The test body of $10^4$\,kg with diagonal inertia $10^4\,\mathrm{kg}\,\mathrm{m}^2$ in zero gravity keeps the pad from moving during the force probe. At the three indentations, normal-load errors are 0.58--1.26\% and effective-radius errors are 0.23--0.52\%. With default reduction at the same resolution, effective-radius errors are 6.33--10.74\%. Moment matching targets the friction moment, so this agreement checks that design rather than grasp fidelity in general.

The point-contact configuration of panel (d) uses the same pad mesh through a convex collision representation with a few contact points per pair, matched to each reference normal load by pose bisection, and its effective-radius errors are 32.1--57.6\%. An additional nonnegative torsional-friction coefficient fitted at the middle load reaches its zero lower bound because the manifold already produces more torque than the reference. This configuration does not represent every point-contact setup.

\paragraph{Fingertip compression and dissipation}
\label{par:fingertip-compression-study}
\autoref{fig:fingertip-patch-friction}(b,e) uses the same isolated pad, test body, stiffness, voxel, quadrature spacing, and solver settings, with default contact reduction and $\alpha=20\,\mathrm{s}/\mathrm{m}$. Each prescribed pose and velocity is reset before one solver step, so these are constitutive probes rather than a continuously actuated indentation rollout.

For panel (b), the prescribed cycle is $\delta=\delta_{\max}(1-\cos\theta)/2$ with $\delta_{\max}=1$\,mm and $\dot\delta=v_{\max}\sin\theta$, sampled at 129 phases. Peak speeds of 5, 10, and 20\,mm/s give loop work of 1.429, 2.857, and 5.712\,mJ, respectively, 10.8\% above direct pressure-law quadrature with the same velocity factor. The excess comes from the implicit treatment of the elastic force, which is evaluated at the end-of-step penetration and therefore adds $khv$ to the loading branch and subtracts it from the unloading branch, where $k$ is the local force gradient, $h$ the step, and $v$ the indentation speed. The negligible-dissipation baseline ($\alpha=10^{-6}\,\mathrm{s}/\mathrm{m}$) at 20\,mm/s shows the same term as a residual loop work of 0.596\,mJ, and at a 0.1\,ms step the fastest-cycle excess falls to 0.4\%.

Panel (e) holds approach and release speeds at matched magnitudes for each of 31 indentations from 0.25 to 1\,mm and extracts
\begin{equation}
c_{\mathrm{eff}}(\delta)=\frac{F_{\mathrm{load}}(\delta,+v)-F_{\mathrm{unload}}(\delta,-v)}{2v},
\end{equation}
where $F_{\mathrm{load}}$ and $F_{\mathrm{unload}}$ are the normal forces on the loading and unloading branches and positive $v$ denotes increasing indentation. At $v=10$\,mm/s, computed coefficients are 80.4, 175.9, and 396.7$\,\mathrm{N}\,\mathrm{s}/\mathrm{m}$ at indentations of 0.25, 0.5, and 1\,mm, 18.7, 12.2, and 5.4\% above $\alpha F_{\mathrm{el}}$, where $F_{\mathrm{el}}$ is the elastic normal force. The offset is the implicit term of the previous paragraph, which shifts the coefficient by $kh$, and it falls below 1.2\% at a 0.1\,ms step. Repeating at 5 and 20\,mm/s changes the extracted coefficient by at most 1.5\%. The analytical constant-damping comparator has $c=156.82\,\mathrm{N}\,\mathrm{s}/\mathrm{m}$, calibrated to the reference at 0.5\,mm. The comparator is a lumped constant-$c$ Kelvin-Voigt damper in parallel with the same elastic response, not a separate engine measurement or every distributed viscoelastic model.

Panels (c,f) are separate predictions of Eq.~\eqref{eq:nonlinear-pad-pressure}, using $\beta\in\{0,0.5,1,2\}$, fixed $k_h=10^8\,\mathrm{N}/\mathrm{m}^3$, and $\ell=1$\,mm. These illustrative parameters are not fitted to a physical pad. For 41 indentations from 0 to 1\,mm, original-mesh signed-distance queries provide local interior depth, and planar quadrature integrates each pressure law. These curves show equal initial local pressure gradients and increasing resistance at greater compression, and are computed offline, independently of the solver preset described in \S~\ref{par:nonlinear-pressure}.

\paragraph{Wheel-ground contact}
\label{par:wheel-ground-study}
\autoref{fig:tire-ground-contact} uses a simplified differential-drive rig with the base dimensions of \friday rather than the complete robot. Two cylindrical wheels hang from a box chassis on preloaded vertical springs, and four freely rotating swivel casters, two of them on a rear pivot, share the load, for a total mass of about 135\,kg. Wheel radius is 70\,mm and half-track $b$ is 262.2\,mm. Wheel density is held at $1200\,\mathrm{kg}/\mathrm{m}^3$ as width varies, so wheel mass varies with width. Each rollout lasts 4\,s at a 2\,ms step with at most 32 solver iterations and warm starting, $k_h=10^{10}\,\mathrm{N}/\mathrm{m}^3$, a 1\,mm SDF voxel target, 0.5\,mm quadrature spacing, and $\mu=0.6$ for the wheels, casters, and ground. Both contact methods use the same rig and velocity-servo settings at each width. The rim-point baseline removes the axial midpoint from the current analytic contact manifold, so it is not a comparison against the current midpoint-enhanced manifold.

After 1\,s of settling, opposite wheel-velocity targets command an ideal in-place yaw rate of 1\,rad/s. Over the final second, unwrapped chassis yaw gives the computed yaw rate, normalized by the mean kinematic rate computed from achieved wheel speeds. Widths are 10, 30, 53, 70, 88, and 120\,mm, with one rollout per method and width.

The pressure illustration in panel (b) is a separate static probe on the authored profile of the \friday drive wheel at the indentation estimated for a 300\,N load, and its samples before contact reduction integrate to 266.6\,N. The visual mesh is narrower than the 88\,mm collision cylinder of the robot asset, which the width sweep in panel (c) uses.

\paragraph{Comparison with MuJoCo}
\label{par:solver-comparison}

We compare CDMM with Newton's MuJoCo backend on \friday lifting its own battery in two configurations. The first keeps each backend's default contact treatment across time steps (\autoref{tab:solver-comparison-rollouts}). The second is the stiff-contact case of \autoref{fig:contact-solver-comparison}.
In the first configuration, both backends start from the same grasp state and follow the same joint position targets through the same controller, with MuJoCo's native collision detection disabled and contacts generated at each substep from the current state by the shared reduced SDF-hydroelastic pipeline.
The 8.2\,s path raises the wrist by about 25\,cm and is followed by a 2\,s hold at the final target.
The hand and battery use $k_h=10^{12}\,\mathrm{N}/\mathrm{m}^3$, $\mu=1.0$, SDF voxels of 1.61--2.00\,mm for the fingers and 1.91--1.92\,mm for the battery, default contact reduction, and warm starting, with CDMM using at most 32 iterations and $\alpha=20\,\mathrm{s}/\mathrm{m}$ and MuJoCo using at most 15 or 100 iterations, and all recorded model parameters match between backends.

CDMM uses near-rigid regularization and Hunt-Crossley damping. MuJoCo uses positive-format contact parameters, whose time constants are floored at twice the time step, with an elliptic friction cone and an impedance ratio of 1000.
The two contact laws therefore differ, so these results compare the backends as configured and do not isolate the numerical solver.
We count a rollout as lifted when the final battery rise exceeds 10\,cm and as jittering when the root-mean-square third difference of the battery position, sampled at 60\,Hz over the final 1.8\,s, exceeds 2\,mm. We run five rollouts per setting.
Contact generation is not forced to be deterministic, and repeated rollouts differ.

\begin{table}[!htb]
\centering\footnotesize
\caption{Lifted and jittering rollouts out of five per setting, written as lifted / jittering. All 60 rollouts remain finite.}
\label{tab:solver-comparison-rollouts}
\begin{tabular}{@{}llll@{}}
\toprule
\tablehead{Time step} & \tablehead{CDMM, at most 32} & \tablehead{MuJoCo, at most 15} & \tablehead{MuJoCo, at most 100}\\
\midrule
$1/240$\,s & 5 / 0 & 0 / 0 & 0 / 0\\
$1/480$\,s & 5 / 0 & 0 / 0 & 0 / 0\\
$1/960$\,s & 5 / 0 & 0 / 0 & 0 / 0\\
$1/1920$\,s & 5 / 0 & 1 / 1 & 0 / 0\\
\bottomrule
\end{tabular}
\end{table}

CDMM lifts the battery in all 20 rollouts, and none of them jitters.
MuJoCo lifts it in one of 40 rollouts, at $h=1/1920$\,s, with a cap of 15 iterations, and that battery jitters with a value of 5.5\,mm. The largest value among the CDMM rollouts is 1.3\,mm, and the threshold is set after inspecting these results.
Raising the MuJoCo iteration cap from 15 to 100 does not produce a lift in these rollouts.

\autoref{fig:contact-solver-comparison} uses a separate configuration at $h=1/2400$\,s with an iteration cap of 32 in both backends. It starts from a different input with a 1\,s hold, a commanded 15\,cm wrist lift over 0.35\,s, and a 3\,s final hold.
To reduce differences in the normal-contact model, we disable near-rigid regularization in both backends and assign each patch contact to MuJoCo through direct-format parameters built from its patch stiffness and Hunt-Crossley damping. These parameters bypass the time-constant floor, and the mapping relies on a reference-mass approximation, so friction, integration, and the articulated contact response remain backend-specific.
CDMM completes the 4.35\,s rollout with a final battery rise of about 13.3\,cm. MuJoCo raises the battery by about 5.8\,cm before the hand and battery poses change abruptly near $t=1.185$\,s, and the first nonfinite state occurs at $t=1.18875$\,s. This is one rollout per backend, chosen after a parameter search, and the outcome depends on stiffness, contact mapping, and time step.
These observations characterize the stated configurations rather than general solver superiority, and we make no runtime claim.

\subsection{Magnetic tactile response kernel}
\label{app:tactile-kernel}

The tactile model adopts spatially weighted displacement superposition from HydroShear~\citep{dang2026hydroshear}.
We adapt this formulation to the discrete magnets of the \friday hand using loads reconstructed from hydroelastic contact patches. This subsection specifies the resulting pressure-to-displacement mapping and describes the sensor-side shear history and coupling assumptions.

\paragraph{Pressure-induced displacement}

All positions and normals are expressed in the sensor frame.
Let $\mathbf{q}_m$ denote the undeformed position of magnet $m$, and let $\mathbf{x}_i$, $\mathbf{n}_i$, and $A_i$ denote the position, inward unit normal, and area of contact contribution $i$.
Each contribution carries an elastic pressure $p_i$
on the extracted pressure-equilibrium surface, reconstructed from its depth and stiffness with the linear law of Eq.~\eqref{eq:sdf-pressure-field},
giving a normal load $F_i=p_iA_i$.

The tangential offset from contact $i$ to magnet $m$ and its Gaussian weight are
\begin{equation}
\mathbf{r}_{mi} = (I-\mathbf{n}_i\mathbf{n}_i^{\mathsf T})(\mathbf{q}_m-\mathbf{x}_i),\qquad w_{mi} = \exp\!\left(-\frac{\|\mathbf{r}_{mi}\|^2}{2\ell^2}\right),
\label{eq:tactile-kernel-weight}
\end{equation}
where $I$ is the three-dimensional identity matrix and $\ell>0$ sets the lateral response scale. The vector response per unit normal load is
\begin{equation}
\begin{aligned}
\mathbf{G}_{mi}
&= e^{-d_m/\lambda}\Bigg[c_n w_{mi}\mathbf{n}_i +c_t\left(
\varrho e^{1/2}\frac{\mathbf{r}_{mi}}{\ell}w_{mi}
+(1-\varrho)\frac{2\ell\mathbf{r}_{mi}}{\|\mathbf{r}_{mi}\|^2+\ell^2}
\right)\Bigg],\\
\mathbf{u}^{\mathrm{press}}_m &= \sum_i \mathbf{G}_{mi}F_i,
\end{aligned}
\label{eq:tactile-kernel-explicit}
\end{equation}
where $c_n$ and $c_t$, in $\mathrm{m/N}$, control normal and radial compliance, and the magnet depth $d_m$ and attenuation length $\lambda$ determine depth attenuation. The parameter $\varrho\in[0,1]$ blends the local Gaussian and longer-range radial responses.
This phenomenological kernel does not resolve volumetric elasticity or deformation caused by rigid magnet inclusions.

\paragraph{Sensor-side queries and shear history}
\label{par:tactile-shear-history}
Shear deformation is tracked at query points on the
sensor surface. At contact onset, each query stores its
position in the contacted object's local frame as a reference.
As the object moves, the reference's displacement relative to
the sensor surface is projected onto the local tangent plane
and spatially weighted to estimate magnet displacement.

Each query maintains its own contact history, which resets
upon contact loss. An optional Coulomb limit with its own friction coefficient
clips the accumulated tangential displacement and moves the
reference accordingly. It is off by default, so a query sticks
until it loses contact. 

\paragraph{Coupling to contact dynamics}

\label{par:tactile-coupling}

The tactile model reconstructs elastic loading from
collision geometry rather than using the solver's
final contact reactions. Damping, solver friction,
near-rigid regularization, and nonlinear pressure
stiffening therefore do not directly enter the
sensor response, although they can affect it through
the resulting motion and contact geometry.

Sensor deformation is not fed back into the dynamics.

\paragraph{Parameters and calibration scope}

\autoref{tab:tactile-kernel-parameters} lists the default parameters of Eq.~\eqref{eq:tactile-kernel-explicit}. These are initialization values, not experimentally calibrated benchmark settings. No calibration result or accuracy claim is implied by the defaults.

\begin{table}[t]
\centering\small
\caption{Pressure-response kernel parameters and their default values. Lengths are reported in SI units.}
\label{tab:tactile-kernel-parameters}
\begin{tabularx}{\linewidth}{@{}lXl@{}}
\toprule
\tablehead{Parameter} & \tablehead{Meaning} & \tablehead{Default}\\
\midrule
$c_n$ & Normal compliance & $10^{-5}\,\mathrm{m}/\mathrm{N}$\\
$c_t$ & Radial compliance & $10^{-5}\,\mathrm{m}/\mathrm{N}$\\
$\varrho$ & Local/global radial blend & $0.5$\\
$\ell$ & Lateral response scale & $2\times10^{-3}\,\mathrm{m}$\\
$\lambda$ & Depth attenuation length & $10^{-3}\,\mathrm{m}$\\
$d_m$ & Configured magnet depth & $1.5\times10^{-3}\,\mathrm{m}$\\
\bottomrule
\end{tabularx}
\end{table}

\paragraph{Same-load probes}
\label{par:tactile-probes}
\autoref{fig:tactile-probes} uses the right-index distal pad of Appendix~\ref{par:fingertip-patch-study}, isolated from the articulated hand, with the linear pressure law at $k_h=10^8\,\mathrm{N}/\mathrm{m}^3$ and the default parameters of \autoref{tab:tactile-kernel-parameters}. The magnet array is centered across the pad width as in the \friday hand design, with a 5\,mm pitch. Each probe is a rigid body with $k_h=10^{12}\,\mathrm{N}/\mathrm{m}^3$ and a flat face. The first is a 12 by 10\,mm face. The second is a 3 by 10\,mm ridge, offset 4.5\,mm toward the fingertip and 1.5\,mm across the pad. The third is a pair of 4 by 4\,mm posts, 9\,mm apart and offset 1.5\,mm across the pad.

Probes are posed kinematically and no solver step is taken. Collision and sensor evaluation use a 0.125\,mm SDF grid without contact reduction, and bisection on the normal offset brings the net normal load to within 0.2\% of 1\,N. The resulting indentations are 0.14, 0.44, and 0.38\,mm, below the 1.5\,mm magnet depth. Direct quadrature of the pad height field agrees with these loads within 2\%. Faces on the probe sides raise the total face load above the net normal load by about 10\% for the ridge and the posts.

The cosine similarity of the Hall readings between probe pairs is 0.67 to 0.80, against 1 for a model driven by net force alone. It changes by at most 0.01 on a 0.25\,mm grid. These values depend on probe placement, and moving the ridge 4.5\,mm changes them by more than 0.1, so they illustrate the response and do not measure the discrimination of the sensor. The field magnitudes are not meaningful under default parameters. The study does not involve hardware measurements.

\clearpage
\section{Actuator model and integration details}
\label{app:actuator-implementation}

This appendix gives the local state updates and parameter definitions of the actuator model in \S~\ref{sec:actuator}. The equations describe reduced actuator models, not resolved electromagnetic fields or hydraulic circuits. Unless stated otherwise, geometry and driving quantities are sampled at the beginning of a substep of duration $h$.

\subsection{Drive and transmission models}
\label{app:actuator-models}

\paragraph{Electrical and thermal states}
\label{par:actuator-electrical}
The winding resistance is evaluated from the current temperature as $R^k=\max(R_0[1+\alpha_R(T^k-T_0)],R_{\min})$, where $R_0$ is the resistance at the reference temperature $T_0$, $\alpha_R$ is its temperature coefficient, and $R_{\min}>0$ prevents a degenerate circuit. After control and voltage limiting determine $V^k$, the current update is
\begin{equation}
\begin{aligned}
i_\infty^k &= \frac{V^k-k_e\omega_m^k}{R^k},\\
\widetilde{i}^{k+1} &= i_\infty^k+(i^k-i_\infty^k)e^{-R^kh/L},\\
i^{k+1} &= \operatorname{clip}(\widetilde{i}^{k+1},i_{\min},i_{\max}).
\end{aligned}
\label{eq:actuator-current-update}
\end{equation}
Here, $i_{\min}$ and $i_{\max}$ are the current limits. For $L=0$, the pre-limit current is $i_\infty^k$. The exponential update is exact for the frozen linear circuit before current clipping, not for the complete coupled motion and controller response. In the current-commanded path, the requested voltage includes back-EMF compensation, whereas the direct-voltage controller does not cancel back-EMF. Both paths remain subject to the available voltage range.

With copper loss $P_{\mathrm{Cu}}^k=(i^{k+1})^2R^k$ held fixed, the lumped winding temperature advances as
\begin{equation}
T^{k+1}=T_\infty^k+(T^k-T_\infty^k)e^{-h/(R_{\mathrm{th}}C_{\mathrm{th}})},
\qquad T_\infty^k=T_{\mathrm{amb}}+R_{\mathrm{th}}P_{\mathrm{Cu}}^k.
\label{eq:actuator-temperature-update}
\end{equation}
Here, $R_{\mathrm{th}}$ and $C_{\mathrm{th}}$ are thermal resistance and capacity, and $T_{\mathrm{amb}}$ is the ambient temperature. The new temperature affects resistance on the following substep. Reported terminal power is the signed quantity $V^ki^{k+1}$. It is not a battery-energy model or a substep-integrated energy balance.

The control loops advance at each simulation substep. Separate firmware sampling rates and communication jitter are not modeled. Battery discharge, automatic bus-voltage sag, and thermal protection logic are outside the electrical model.

\paragraph{Transmission geometry and reflected inertia}
\label{par:actuator-transmission}
Constant gearing uses $\phi(q)=nq$ with gear ratio $n$. For the offset slider-crank, let $a$ be crank radius, $b$ connecting-rod length, $e$ slider-axis offset, $c$ motor rotation per slider travel, and $\epsilon\in\{-1,1\}$ the selected geometric branch. With $\theta=q+q_0$ for a joint offset $q_0$, the motor position is
\begin{equation}
\phi(q)=c\left[a\sin\theta+\epsilon\sqrt{b^2-(e-a\cos\theta)^2}\right].
\label{eq:actuator-slider-crank}
\end{equation}
The local ratio $r=\phi'(q)$ maps velocity, while the forward-drive efficiency $\eta$ scales the transmitted motor effort $\tau_m$ to the joint effort $\tau_q=\eta r\tau_m$ before any unilateral output restriction. This scalar efficiency is not a detailed direction-dependent loss model for forward and backdriven operation.

The coupled rotor kinetic energy is $\tfrac12 I_r r(q)^2\dot q^2$. Its contribution to the joint equation therefore contains both $I_r r^2\ddot q$ and $I_r r r'\dot q^2$. The first enters the mass matrix and the second enters the inertial bias, without electrical clipping or efficiency scaling. Invalid slider-crank geometry disables transmission output, and the validity check is not a regularization of near-singular transmission ratios. Joint ranges should avoid geometric singularities.

The transmission is kinematic, so backlash and elastic deformation of gears or linkages are not resolved.

\paragraph{LuGre friction}
\label{par:actuator-friction}
For motor speed $\omega_m$, the deterministic LuGre model uses a bristle deflection $z$ and a Stribeck torque envelope $g$:
\begin{equation}
\begin{aligned}
g(\omega_m)&=\tau_c+(\tau_s-\tau_c)e^{-(\omega_m/\omega_s)^2},\\
\dot z&=\omega_m-\lambda z,\qquad \lambda=\frac{\sigma_0|\omega_m|}{g(\omega_m)},\\
\tau_f&=\sigma_0z+\sigma_1\dot z+B\omega_m.
\end{aligned}
\label{eq:actuator-lugre}
\end{equation}
Here, $\sigma_0$ and $\sigma_1$ are bristle stiffness and damping, $B$ is viscous damping, and $\tau_s$, $\tau_c$, and $\omega_s$ set the static level, sliding level, and transition speed. A positive floor is applied to $g$. At fixed speed, the state update is
\begin{equation}
z^{k+1}=e^{-\lambda h}z^k+\frac{1-e^{-\lambda h}}{\lambda}\omega_m,
\label{eq:actuator-bristle-update}
\end{equation}
with the continuous limit $z^{k+1}=z^k+h\omega_m$ as $\lambda\to0$.

\paragraph{Low-speed variability}
\label{par:actuator-variability}
The optional variability models use independently seeded, temporally correlated processes. Each normalized Ornstein-Uhlenbeck state follows
\begin{equation}
x^{k+1}=a_{\mathrm{OU}} x^k+\sqrt{1-a_{\mathrm{OU}}^2}\,\epsilon_{\mathrm{OU}}^k,
\qquad a_{\mathrm{OU}}=e^{-h/\tau_{\mathrm{noise}}},\qquad \epsilon_{\mathrm{OU}}^k\sim\mathcal N(0,1),
\label{eq:actuator-ou-update}
\end{equation}
where $\tau_{\mathrm{noise}}$ is the correlation time. The normalized process has unit variance at stationarity, and resetting it to zero does not start it in its stationary distribution.

For friction variation, the log-scale amplitude depends on joint speed $v=\dot q$ through $\ell(v)=\ell_0[\rho+(1-\rho)e^{-(v/v_f)^2}]$. The scale applied during friction-force and implicit-coefficient evaluation is
\begin{equation}
s_f=\operatorname{clip}\!\left(\exp\!\left[\ell(v)x^{k+1}-\tfrac12\ell(v)^2\right],s_{\min},s_{\max}\right).
\label{eq:actuator-friction-variation}
\end{equation}
It jointly multiplies $\sigma_0$, $\sigma_1$, $\tau_c$, and $\tau_s$, preserving their ratios. The parameters $\ell_0$, $v_f$, and $\rho$ control the low-speed amplitude, velocity scale, and high-speed fraction, and $s_{\min}$ and $s_{\max}$ bound the scale. The exponential correction gives unit mean before clipping for a stationary normalized Gaussian state and a fixed envelope, but clipping and velocity-state dependence need not preserve this mean.

A separate process $y$ produces a joint-effort disturbance $d_{\mathrm{effort}}=\sigma_d e^{-(v/v_d)^2}y^{k+1}$ with amplitude $\sigma_d$ and velocity scale $v_d$, which is subtracted from the actuator output. Its amplitude, correlation time, and seed are set independently. Unlike passive friction, this additive disturbance can inject mechanical energy. Both extensions are empirical descriptions of low-speed variability rather than microscopic contact models.

\paragraph{Reduced single-acting response}
\label{par:actuator-single-acting}
Let $s_{\mathrm{drive}}\in\{-1,1\}$ specify the permitted drive direction and $u^k$ be the joint-side drive effort before the unilateral hydraulic response. A nonnegative chamber-effort state $H$ is updated and applied as
\begin{equation}
\begin{aligned}
H^{k+1}&=a_hH^k+(1-a_h)[s_{\mathrm{drive}}u^k]_+,\qquad a_h=e^{-h/\tau_h},\\
\tau_{\mathrm{drive}}&=s_{\mathrm{drive}}[H^{k+1}-D_h s_{\mathrm{drive}}\dot q^k]_+.
\end{aligned}
\label{eq:actuator-single-acting}
\end{equation}
The time constant $\tau_h$ represents effort buildup and decay, and $D_h$ sets joint-side flow damping. The passive return spring and damper are added independently of this unilateral drive restriction. The implicit solve retains the flow-damping term across the zero-output boundary as described in \S~\ref{sec:actuator-solver}. Eq.~\eqref{eq:actuator-single-acting} defines the evaluated output, not an exact implicit solution of that boundary.

The model is a joint-effort-level approximation, not a resolved hydraulic circuit with explicit valves, chamber volumes, or fluid compressibility.

\subsection{State integration and dynamics coupling}
\label{app:actuator-coupling}

\paragraph{Implicit coupling approximations}
Transmission geometry and local coefficients are held fixed during each substep, so the coupling of \S~\ref{sec:actuator-solver} does not solve all electrical, thermal, and mechanical states simultaneously. The coefficients retained through saturation are not the exact derivatives of the clipped output law, and the unilateral hydraulic transition is regularized without locating its event time. Controller anti-windup addresses accumulated tracking error, while the implicit terms address the numerical response within the substep. The split update does not remove temporal-discretization error, so rapid changes in transmission geometry, saturation, or friction still require adequate substep resolution.

\paragraph{Bristle stiffness in the articulated solve}
During the articulated solve, the implicit motor-side bristle stiffness uses the nonnegative approximation
\begin{equation}
K_z=\sigma_0\operatorname{clip}\!\left(e^{-\lambda h}\left[1-\frac{\operatorname{sgn}(\omega_m)\sigma_0z^k}{g(\omega_m)}\right],0,2\right).
\label{eq:actuator-bristle-tangent}
\end{equation}
This approximation holds the Stribeck envelope fixed and is not a full derivative of the nonlinear friction law. After the solve, Eq.~\eqref{eq:actuator-bristle-update} is evaluated from the old bristle state using the solved velocity and the frozen transmission ratio. The pre-solve provisional state is not integrated a second time.

\clearpage
\section{Asset generation details}
\label{app:asset-generation}
This appendix provides additional details for
\S~\ref{sec:eris}.
\subsection{Constraint types}
\label{app:eris-constraints}
As described in \S~\ref{sec:eris-routing},
the structured set of task constraints
$\widehat{\mathcal{R}}_{\mathrm{task}}$
is instantiated from a fixed vocabulary of constraint types.
\autoref{tab:eris-constraints} summarizes the seven
categories and their representative constraint types.
Each type specifies the fields needed to instantiate it,
and the resulting constraints are used by the
rule-based router $\rho$ for branch selection.
\begin{table}[!htb]
\centering
\caption{
Constraint types used by \eris, grouped into seven categories.
Representative types are shown for each category.
}
\label{tab:eris-constraints}
\small
\begin{tabularx}{\linewidth}{@{}l l X@{}}
\toprule
Category & Representative constraint types & Description \\
\midrule
Property
& \texttt{dimension}, \texttt{part\_count},
  \texttt{min\_thickness}, \texttt{appearance}
& Dimensions, counts, wall thickness, and appearance \\

Access
& \texttt{open\_top}, \texttt{access}, \texttt{graspable}
& Openings and manipulation accessibility \\

Fit
& \texttt{fits}, \texttt{clearance}, \texttt{aligns}, \ldots
& Geometric compatibility, fit, and clearance \\

Articulation
& \texttt{dof}, \texttt{actuated}, \texttt{reaches\_state}, \ldots
& Relative motion and actuation \\

Support
& \texttt{supports}, \texttt{retains}, \texttt{stable}, \ldots
& Load support, retention, and stability \\

Transport
& \texttt{transports}, \texttt{guides}, \texttt{presents}, \ldots
& Object transport and presentation \\

Interaction
& \texttt{inserts}, \texttt{removes}, \texttt{engages}
& Part-to-part interactions \\
\bottomrule
\end{tabularx}
\end{table}
\subsection{SDF construction and collider layout}
\label{app:eris-sdf}
This section provides additional details on the collider
verification stage summarized in
\autoref{tab:eris-collision-checks} (\S~\ref{sec:eris-verification}).
We describe adaptive SDF authoring and simulator-side validation in \hdaynewton,
the collision-cost model, and bake-verified collider-layout optimization.

\paragraph{Adaptive SDF authoring and validation}
Before baking each collider, \eris applies the
resolution-admissibility check to determine which SDF
voxel sizes are geometrically admissible.
It then authors the SDF parameters used by \hdaynewton,
most importantly the voxel size that determines collision resolution.
For each collider $c_i$, \eris derives an upper bound on voxel size from geometric information in the constructed asset.
Let $\ell_i^{\min}$ denote the minimum side length of the collider's axis-aligned bounding box, $w_i$ a measured wall thickness when available, and $f_i$ the scale of an explicitly declared collision feature.
These quantities constrain the resolution according to the collider's overall scale, locally thin structures, and declared functional features, respectively.
The factors $\kappa_{\mathrm{size}}$, $\kappa_{\mathrm{wall}}$, and $\kappa_{\mathrm{feat}}$ specify the corresponding sampling requirements:
\begin{equation}
    u_i
    =
    \min
    \left\{
        \frac{\ell_i^{\min}}{\kappa_{\mathrm{size}}},
        \frac{w_i}{\kappa_{\mathrm{wall}}},
        \frac{f_i}{\kappa_{\mathrm{feat}}}
    \right\}.
    \label{eq:eris-sdf-bound}
\end{equation}
\eris then initializes the search at the coarsest supported voxel size $\Delta_{\mathrm{vox},i}$ that satisfies $\Delta_{\mathrm{vox},i} \leq u_i$, subject to any additional simulator- or layout-specific caps, where smaller voxel sizes correspond to finer SDF resolutions.
Terms corresponding to unavailable measurements or declarations are omitted, so $u_i$ is determined by the most restrictive applicable geometric constraint.
The resolution-admissibility check therefore excludes candidate voxel sizes that are too coarse relative to known geometric scales, without attempting to predict the exact surface produced by SDF discretization.

\eris therefore performs a second, complementary check after conversion.
\hdaynewton bakes the collider, and the SDF-fidelity check runs on the resulting simulation representation.
Whereas resolution admissibility evaluates the sampling conditions before baking, SDF fidelity evaluates the geometry that the simulator will actually use.
Let $\mathcal{S}_i$ denote samples drawn from the source surface, and $\Sigma_i$ denote the resulting SDF isosurface.
We define the surface coverage $C_i$ and surface area $A_i$ as:
\[
C_i =
\frac{1}{|\mathcal{S}_i|}
\sum_{x\in\mathcal{S}_i}
\mathbf{1}\!\left[
d(x,\Sigma_i) < 2\Delta_{\mathrm{vox},i}
\right],
\qquad
A_i = \operatorname{Area}(\Sigma_i).
\]
Here, $d(x,\Sigma_i)$ is the distance from $x$ to $\Sigma_i$. Let $E_i$ denote the number of open boundaries of $\Sigma_i$, and let $g_i$ and $n_i$ denote its genus and connected-component count, respectively.
A candidate representation passes the SDF-fidelity check when
\begin{equation}
\begin{aligned}
\frac{C_i}{C_i^{\mathrm{ref}}} &\geq \tau_{\mathrm{cov}},
\qquad
\frac{A_i}{A_i^{\mathrm{ref}}} \geq \tau_{\mathrm{area}}
&& \text{(surface retention)}, \\[2pt]
E_i &\leq \tau_{\mathrm{open}}
&& \text{(boundary validity)}, \\[2pt]
g_i-g_i^{+} &\leq \Delta_g,
\qquad
n_i-n_i^{+} \leq \Delta_n
&& \text{(topology guard)}.
\end{aligned}
\label{eq:eris-sdf-fidelity}
\end{equation}
Coverage and area provide complementary measures of surface retention.
$C_i$ detects regions of the source surface that are no longer represented nearby in the baked geometry, whereas $A_i$ detects substantial surface loss that can occur even when the remaining baked surface stays close to the source.
Surface collapse can preserve high coverage while reducing area, whereas localized geometric loss can reduce coverage even when the total baked surface area remains large.
Open boundaries identify invalid extracted surfaces, while increases in genus and connected-component count detect spurious perforation and fragmentation, respectively.
Here, ``ref'' denotes the finest available reference bake, while $(\cdot)^{+}$ denotes the corresponding quantity at the next finer voxel size.
When a candidate fails the SDF-fidelity check, \eris moves to a more conservative SDF parameterization and rebakes the collider.

\paragraph{Collision-cost model}
We now specify the cost term $M_{\mathrm{SDF}}$ in
Eq.~\eqref{eq:eris-collision-objective}.
A valid collision representation may still be unnecessarily expensive for
large-scale simulation.
Let $N=|\Pi|$ denote the number of colliders in a layout, let $B_i$ denote the
block count of collider $i$, and let
$B_{\max}=\max_i B_i$.
For a reference number $D$ of dynamic bodies, \eris estimates the collision-memory contribution of the asset with a simulator-derived model:
\begin{equation}
    M_{\mathrm{SDF}}
    =
    a_1 \sum_i B_i
    +
    a_2 D N B_{\max}
    +
    a_3 D N
    +
    a_4 N .
    \label{eq:eris-sdf-memory}
\end{equation}
Here, $a_1$ to $a_4$ are coefficients derived from \hdaynewton's collision-memory model.
The terms capture different sources of collision cost:
finer SDF resolutions increase the aggregate block count $\sum_i B_i$,
retaining many independent colliders increases $N$, and merging parts can increase $B_{\max}$ by placing a spatially larger region into a single collider.
Consequently, SDF resolution and collider partition cannot generally be optimized independently.

\paragraph{Collider-layout optimization}
For multipart code-native assets, $\Pi$ in Eq.~\eqref{eq:eris-collision-objective}
introduces an additional optimization variable.
Keeping every part separate can increase collider-dependent memory overhead, while merging parts too aggressively can cause small or thin source parts to be distorted or partially absorbed by the discretized union.
\eris therefore searches for lower-cost groupings of eligible static parts and subjects the selected merged layout to a separate layout-preservation check after baking.

Moving parts, conveying surfaces, parts carrying explicitly declared collision
features, non-closed geometry, and parts belonging to different rigid bodies
are excluded from merging.
The remaining eligible parts form a graph in which
edges connect touching parts.
For each candidate grouping, \eris reapplies the resolution-admissibility rule
to the geometry of the union and recomputes its block count, since merging
changes both the spatial extent and geometric scales of the collider being
baked.

\eris greedily considers unions of adjacent groups under several deterministic
orderings.
A merge is considered only when the resulting union forms a valid connected
solid, satisfies the per-collider block constraint
\begin{equation}
    B_i \leq B_{\mathrm{cap}},
    \label{eq:eris-block-cap}
\end{equation}
and reduces the predicted collision cost in
Eq.~\eqref{eq:eris-sdf-memory}.
The block cap prevents a locally cost-reducing merge from producing an
individually oversized collider.
The original per-part layout is retained as a candidate throughout the search,
so reducing collider count alone is never sufficient to select a merge.

Predicted cost alone is insufficient because merging changes the geometry that
is ultimately discretized.
\eris therefore bakes the selected layout and applies the layout-preservation check to each source part participating in a merged collider.
For each source part $j$, area-weighted surface samples from both the source union and the baked collider are attributed to the nearest source part whose surface lies within $2\Delta_{\mathrm{vox},i}$ and has a sufficiently aligned normal.
Let $s_j^{\mathrm{src}}$ and $s_j^{\mathrm{bake}}$ denote the resulting surface-sample shares attributed to part $j$ before and after baking.
From the attributed samples, we estimate local wall thickness using the first percentile of distances to opposing surfaces, denoted by $w_j^{\mathrm{src}}$ and $w_j^{\mathrm{bake}}$, respectively.
We quantify local geometric integrity using
\begin{equation}
    \frac{w_j^{\mathrm{bake}}}
         {w_j^{\mathrm{src}}}
    \geq \tau_{\mathrm{wall}},
    \qquad
    \frac{s_j^{\mathrm{bake}}}
         {s_j^{\mathrm{src}}}
    \geq \tau_{\mathrm{share}} .
    \label{eq:eris-layout-preservation}
\end{equation}
Here, $\tau_{\mathrm{wall}}$ specifies the minimum wall-retention ratio, while $\tau_{\mathrm{share}}$ specifies the minimum retained surface-share ratio.
The former detects local squeezing or necking, whereas the latter detects parts that are partially erased or absorbed by the merged SDF even when their remaining thickness is preserved.
These failures can be hidden by collider-level statistics when the affected part occupies only a small fraction of a merged collider, motivating the part-wise preservation check.

Parts for which the preservation measurement is inconclusive are treated conservatively as failures.
Parts that fail the check or yield inconclusive measurements
are forced to remain separate, after which the remaining
eligible parts are laid out and baked again.
If no verified merged layout is obtained within the retry budget, \eris falls back to the original per-part layout.
The retry budget limits repeated baking in the simulator and does not relax the preservation requirements.
\autoref{alg:merge-parts} summarizes this bake-verified optimization.

When adaptive SDF construction and collider-layout optimization still cannot produce an acceptable collision representation, the measurements return through the representation-specific revision path of \S~\ref{sec:eris-verification}.
\begin{algorithm}[t]
\caption{Bake-verified collider-layout optimization}
\label{alg:merge-parts}
\small
\begin{algorithmic}[1]
\Require Eligible static parts $\mathcal A$, touching graph $\mathcal G$,
         block cap $B_{\mathrm{cap}}$, retry budget $R$, greedy orderings $\mathcal O$
\Ensure Bake-verified layout $\Pi$, or the per-part layout $\Pi_{\mathrm{part}}$
\State $\mathcal C_{\mathrm{merge}} \gets \varnothing$
    \Comment{parts forced to remain separate}
\For{$r=1,\ldots,R$}
    \State $\mathcal P \gets \{\Pi_{\mathrm{part}}\}$
    \For{greedy ordering $o\in\mathcal O$}
        \State $\Pi_o \gets
        \Call{GreedyMerge}{\mathcal A,\mathcal G,\mathcal C_{\mathrm{merge}},o,
                           M_{\mathrm{SDF}},B_{\mathrm{cap}}}$
        \State $\mathcal P \gets \mathcal P\cup\{\Pi_o\}$
    \EndFor
    \State $\Pi \gets
        \arg\min_{\Pi'\in\mathcal P}M_{\mathrm{SDF}}(\Pi')$
    \State $\widehat{\Pi}\gets\Call{Bake}{\Pi}$
        \Comment{\hdaynewton SDF bake}
    \State $\mathcal F\gets\Call{Verifyparts}{\Pi,\widehat{\Pi}}$
        \Comment{part-wise layout preservation}
    \If{$\mathcal F=\varnothing$}
        \State \Return $\Pi$
    \EndIf
    \State $\mathcal C_{\mathrm{merge}}\gets\mathcal C_{\mathrm{merge}}\cup\mathcal F$
        \Comment{failed or unjudged parts stay separate}
\EndFor
\State \Return $\Pi_{\mathrm{part}}$
\end{algorithmic}
\end{algorithm}

\subsection{Models, budgets, and thresholds}
\label{app:eris-config}

The generators, model settings, and common generation budget
used in the \eris experiments are summarized in
\autoref{tab:eris-generation-config}.
All branches allow up to $K=4$ generation rounds,
including the initial generation and subsequent revisions.
Observed generation cost and wall-clock time are reported
separately in \autoref{tab:eris-generation-profile},
since they depend strongly on the structure and interaction
complexity of the requested asset.

The fixed hyperparameters used for collider construction
and verification are reported in
\autoref{tab:eris-verification-config}.
Task-specific geometric and interaction tolerances
are instantiated from the corresponding task constraints
and are therefore not included as global hyperparameters.

\begin{table}[H]
\centering\small
\caption{Generation configuration and round budgets used by \eris.}
\label{tab:eris-generation-config}
\begin{tabular}{llll}
\toprule
Branch & Generator & Model setting & Budget \\
\midrule
Mesh
& FLUX.2-klein-9B $\rightarrow$ BiRefNet $\rightarrow$ SAM 3D $\rightarrow$ PaMO
& pipeline defaults
& $K=4$ \\
Code-Rigid
& executable Blender program
& claude-opus-5-5 (high)
& $K=4$ \\
Code-Articulated
& Blender program + joint declaration
& claude-opus-5-5 (high)
& $K=4$ \\
\bottomrule
\end{tabular}
\end{table}

\begin{table}[H]
\centering\small
\caption{Observed generation cost and wall-clock time for the code-native \eris pipeline
measured with claude-opus-5-5.}
\label{tab:eris-generation-profile}
\begin{tabular}{lcc}
\toprule
Task class & Wall-clock / unit & Cost / unit \\
\midrule
Static      & 1.9\,min  & \$1.11 \\
Insertion   & 1.8\,min  & \$0.80 \\
Articulated & 12.5\,min & \$3.13 \\
Transport   & 21.1\,min & \$4.29 \\
Machinery   & 11.2\,min & \$2.78 \\
\midrule
Overall     & 10.1\,min & \$2.48 \\
\bottomrule
\end{tabular}
\end{table}

\begin{table}[H]
\centering\small
\caption{Fixed hyperparameters used for collider construction and verification.}
\label{tab:eris-verification-config}
\begin{tabular}{lll}
\toprule
Stage & Parameters & Values \\
\midrule
Resolution admissibility
& $\kappa_{\mathrm{size}},\ \kappa_{\mathrm{wall}},\ \kappa_{\mathrm{feat}}$
& $1.25,\ 1,\ 2$ \\
& Supported voxel sizes
& $\{48,32,24,16,12,8,6,4,3,2,1\}$\,mm \\
\midrule
SDF fidelity
& $\tau_{\mathrm{cov}},\ \tau_{\mathrm{area}},\ \tau_{\mathrm{open}}$
& $0.99,\ 0.80,\ 0$ \\
& $\Delta_g,\ \Delta_n$
& $2,\ 2$ \\
\midrule
Collider layout
& $B_{\mathrm{cap}},\ \tau_{\mathrm{wall}},\ \tau_{\mathrm{share}}$
& $1900,\ 0.90,\ 0.25$ \\
& Maximum collider-layout optimization rounds
& $8$ \\
\bottomrule
\end{tabular}
\end{table}
\clearpage
\section{Scenario generation details}
\label{app:worldgen-config}
This appendix gives the models, budgets, prompts, protocols, and verification thresholds behind \S~\ref{sec:world-generation}, together with the background pipeline, re-rendering, and the post-hoc rubric evaluation.
Values are taken from the system's default parameters and the stored records of the generation runs.
Quoted prompts in this and the following appendices are reproduced verbatim, with line breaks adjusted for typesetting.

\subsection{Inverse scene generation}
\subsubsection{Models and budgets}
\label{app:scene-generation-config}
Model calls return typed JSON and use read-only file access.
Codex-served models use a command-line agent, while Claude models use an adapter described under Controls below (\autoref{tab:app-scene-models}).
The planner first summarizes the request in an outline and then compiles it separately into a scene specification and a task plan. The scene-specification prompt forbids task steps and process state so that the two remain separate.
A run stops when the same checks fail a third time or when an identical candidate is resubmitted.

\begin{table}[H]
\centering\small
\caption{Models and budgets for inverse scene generation in \autoref{tab:model-ablation} and \autoref{tab:plan-critic}, with settings for asset generation.}
\label{tab:app-scene-models}
\begin{tabularx}{\linewidth}{@{}>{\raggedright\arraybackslash}p{0.26\linewidth}>{\raggedright\arraybackslash}p{0.2\linewidth}Y@{}}
\toprule
\tablehead{Stage} & \tablehead{Model (effort)} & \tablehead{Budget and settings}\\
\midrule
Inverse scene generation & gpt-6-astra (high), gpt-6.1-sol (medium), gpt-6-luna (medium), claude-opus-5-5 (high), claude-sonnet-5-5 (medium), gemini-4-argon (high) & 900\,s per call, 2{,}400\,s and 40 calls per run, or 7{,}200\,s and 80 calls with the larger budget, at most 6 scene-specification, 8 plan, and 3 outline rounds, 2 scene critics in fresh contexts, asset generation off unless stated\\
Asset generation in the loop & claude-opus-5-5 (high), claude-sonnet-5-5 fallback & Parametric path of the asset generator, at most 1{,}800\,s per request, used only in the last two rows of \autoref{tab:plan-critic}\\
\bottomrule
\end{tabularx}
\end{table}

The scenes in \autoref{fig:scene-generation} come from runs in \autoref{tab:model-ablation} and \autoref{tab:plan-critic}. The packaging scene in \autoref{fig:gs-worlds} is written by claude-opus-5-5 as a coding agent and reviewed by a person (\S~\ref{par:bg-comparison}).

\subsubsection{Experimental protocols}
\label{app:ablation-protocol}
\autoref{tab:scene-modes} summarizes the experimental settings alongside full ISG and the human review used for the background study.

\begin{table}[H]
\centering\small
\caption{Input and assessment settings for scene generation. Full ISG starts from a task and workplace. The other settings start from a supplied scene description and omit task-plan generation.}
\label{tab:scene-modes}
\begin{tabularx}{\linewidth}{@{}>{\raggedright\arraybackslash}p{0.2\linewidth}YYY@{}}
\toprule
 & \tablehead{Full ISG} & \tablehead{Scene description} & \tablehead{Human review}\\
\midrule
Input & Task and workplace & Scene description & Scene description\\
Task-plan stage & Generated and checked & Not run & Not run\\
Scene specification & Agent-authored & Agent-authored & Agent-authored\\
Deterministic checks & Plan and scene checks & Scene checks & Static verification\\
Assessment & Plan critic and scene critics & Reference-based fidelity checks & Human review\\
\bottomrule
\end{tabularx}
\end{table}

\paragraph{Specification-generation trials}
These trials measure scene-specification writing alone. The input already describes the scene, so no task plan is generated, and each result is scored against a reference specification.
Each model writes scene specifications for the three fidelity tiers of \S~\ref{sec:scene-generation}, with three cold-start trials per tier and at most five attempts per trial.
Each attempt starts a fresh coding agent that writes a scene specification from the supplied description. The result is scored against the tier's reference specification on item count, object set, target placement, alignment, an empty container where the tier requires one, catalog asset identity, position within 3\,cm, support contact within 1\,cm and penetration within 5\,mm, and overlap.
After a failed attempt, the next agent receives the previous specification and a critic report that names each failed criterion, the object involved, and the kind or direction of the error, without reference coordinates.
The reference specifications follow the rubric of each tier and meet every criterion when scored against themselves.
Agents run in a sandbox that is read-only except for a working directory per attempt, the reference specifications are hidden, and every attempt is checked for reads of benchmark definitions.

\begin{promptbox}{Example critic report after a failed attempt (verbatim excerpt)}
\small
position: box\_left: placed moderately off the expected place; it should move toward the viewer's right (+Y).\\
contact: apple\_in\_left\_box: sunk into its support (penetrating it moderately).
\end{promptbox}

\paragraph{Full ISG trials}
Each model runs every stage of inverse scene generation on the same ten tasks with a 2{,}400\,s and 40-call budget per run and asset generation disabled, so that asset supply is held fixed across generation models.
Each task is written together with its workplace as one request, and the ten are every third entry of an industrial corpus of 15 atomic and 15 long-horizon tasks.
The atomic tasks are CNC chuck exchange, adhesive bead application, mixed-SKU case packing, go/no-go gauging with reject, and battery-cell tray loading.
The long-horizon tasks are a mixed-part CNC tending shift, secondary packaging line clearance, injection-molding machine tending, a preventive-maintenance route, and a multi-station pump build.
We report the number of runs with an accepted plan, with at least one candidate that passes every deterministic check, and with an accepted scene.
\autoref{tab:scene-cost} gives the validity and cost of these runs.

\paragraph{Bottleneck removal}
\autoref{tab:plan-critic} repeats the full ISG protocol with gpt-6.1-sol, the Codex-served model with the most accepted plans in \autoref{tab:model-ablation}, adding one change per condition. The conditions add a larger budget of 7{,}200\,s and 80 calls, then asset generation, and then a stated affordance format.
With asset generation, a missing asset triggers the parametric path of the asset generator described above, and each run writes its new assets to its own scratch copy of the catalog, so runs cannot reuse each other's assets.
The asset-generation condition adds to the supplier prompt the evidence fields that the acceptance check reads.
The last condition further requires that the words of each required affordance appear in the description or tags of a registered asset, because acceptance matches them literally.
An asset counts as accepted only after it passes geometry, conversion, mass, collision, and resting checks in simulation, in addition to this match.
Runs whose asset supplier reaches a usage limit of its model provider, two in the asset-generation condition and one in the last condition, are repeated once in a fresh catalog.

\paragraph{Critic swap}
For \autoref{fig:critic-swap}, we take the last fully checked candidate of every run in \autoref{tab:model-ablation}, 25 in total, and re-judge each with gpt-6.1-sol (medium), claude-opus-5-5 (high), and claude-sonnet-5-5 (medium).
Each critic receives, in two fresh contexts, the prompt of the original run rebuilt from its stored plan and candidate, and accepts only if both return a pass with a score of at least 0.8 and no open concern.
A stand-in asset is one whose catalog category, read from the asset evidence of the candidate, differs from the category of the planned entity. A critic counts as blocking a stand-in asset when it raises that asset as a concern in at least one of its two passes.
The critics see the same evidence and disagree on how much category fidelity matters, as the following excerpts show.
\begin{promptbox}{Three critics on the same raw billet (verbatim excerpts)}
\textbf{claude-sonnet-5-5, pass.} Optional, not a defect: the billet asset is catalogued as a toy yellow cylinder [\ldots] and is scaled 2x. Its geometry fits a raw cylindrical billet, so this is acceptable at category fidelity.\\[0.3em]
\textbf{claude-opus-5-5, revise.} Because the catalog evidence contradicts the industrial category the plan requires [\ldots], the scene shows a toy block standing in for machining stock.\\[0.3em]
\textbf{gpt-6.1-sol, revise.} raw\_billet uses a cataloged yellow toy building block with a matte finish. Scaling it by 2 does not make it a raw steel billet.
\end{promptbox}

\paragraph{Shared critic}
\autoref{tab:fixed-critic} repeats the full ISG protocol for gpt-6.1-sol and claude-sonnet-5-5 with every plan and scene critic call served by claude-opus-5-5 (high) in a fresh context, with unchanged prompts and schemas, while all other calls stay on the planner model.
Across the 20 runs, all 104 critic calls are served by claude-opus-5-5 and all 299 other calls by the planner model.

\paragraph{Controls}
The catalog, thresholds, tasks, and budgets are fixed across models.
In the generation runs, critics use the planner's model except in the shared-critic experiment.
Claude models run the same procedure through an adapter that replaces the command-line agent and reads files with read-only search tools instead of a read-only shell.
The adapter checks each structured output against the schema after generation, whereas the Codex-served models receive the schema as a strict output format.
gemini-4-argon runs the same procedure through its own command-line agent with read-only file tools and its thinking level set to high, receives the schema in the prompt with the same check after generation, and executes one run at a time.
One claude-opus-5-5 run ends when its output violates a cross-field rule of the schema, and it counts as a failed run.

\begin{table}[H]
\centering\small
\caption{Validity and cost of inverse scene generation over the ten runs of \autoref{tab:model-ablation}. Candidates are proposed scene specifications, with the number that pass static verification and the number that pass every deterministic check. Calls and minutes are per accepted scene. Dashes mark models without an accepted scene.}
\label{tab:scene-cost}
\begin{tabular*}{\linewidth}{@{\extracolsep{\fill}}lccccc@{}}
\toprule
\tablehead{Model (effort)} & \tablehead{Candidates} & \tablehead{Static} & \tablehead{All checks} & \tablehead{Calls} & \tablehead{Min}\\
\midrule
gpt-6-astra (high) & 1 & 0 & 0 & -- & --\\
gpt-6.1-sol (medium) & 14 & 10 & 8 & 186 & 274\\
gpt-6-luna (medium) & 11 & 6 & 6 & -- & --\\
claude-opus-5-5 (high) & 24 & 21 & 16 & 105 & 147\\
claude-sonnet-5-5 (medium) & 40 & 34 & 31 & \textbf{28} & \textbf{17}\\
gemini-4-argon (high) & 31 & 17 & 8 & 32 & 37\\
\bottomrule
\end{tabular*}
\end{table}

\subsection{Background generation}
\label{app:bg-config}
\begin{figure}[t]
\centering
\begin{subfigure}[t]{0.495\linewidth}
  \includegraphics[width=\linewidth]{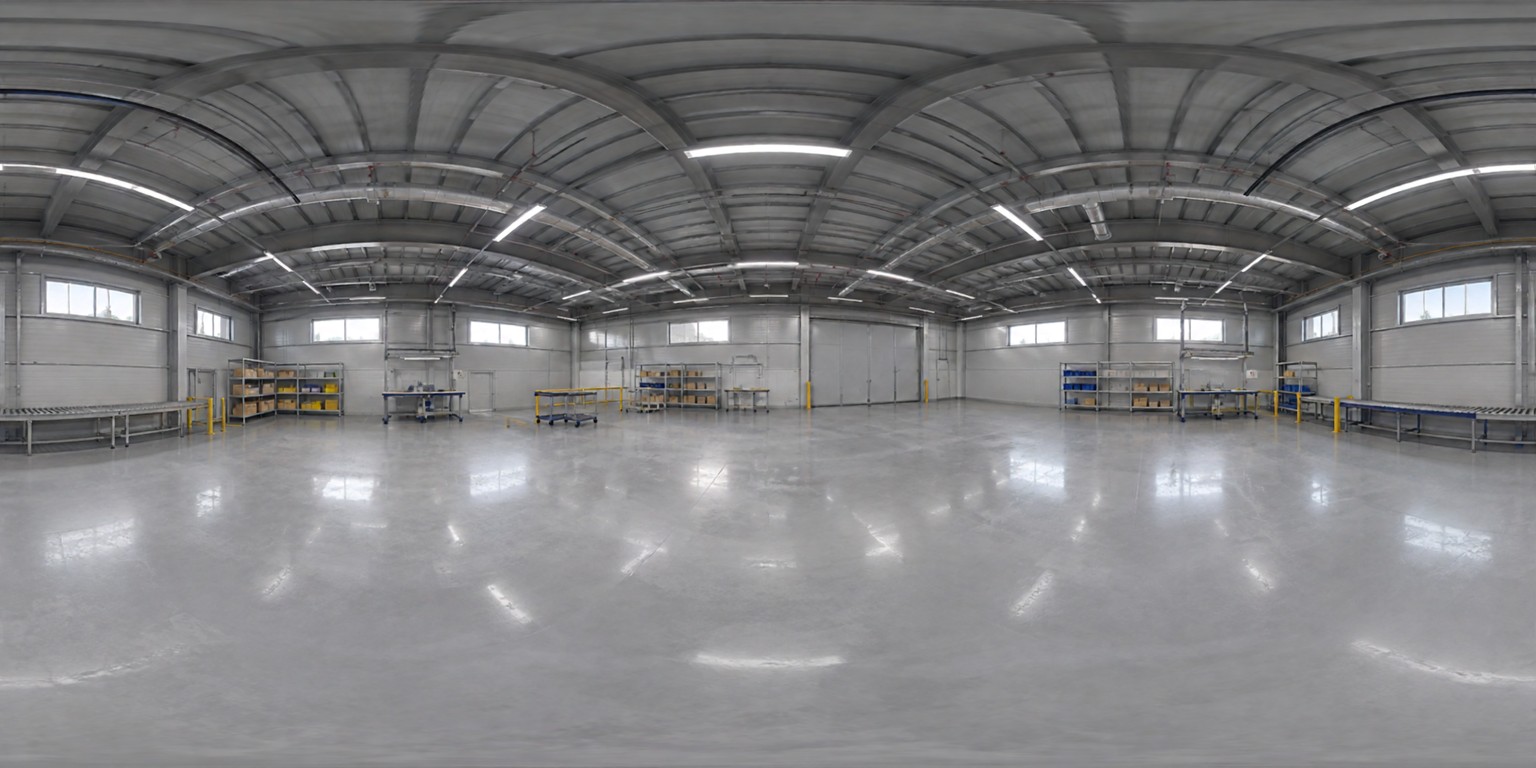}
  \caption{Generated $360^\circ$ panorama (equirectangular)}
\end{subfigure}\hfill
\begin{subfigure}[t]{0.495\linewidth}
  \includegraphics[width=\linewidth]{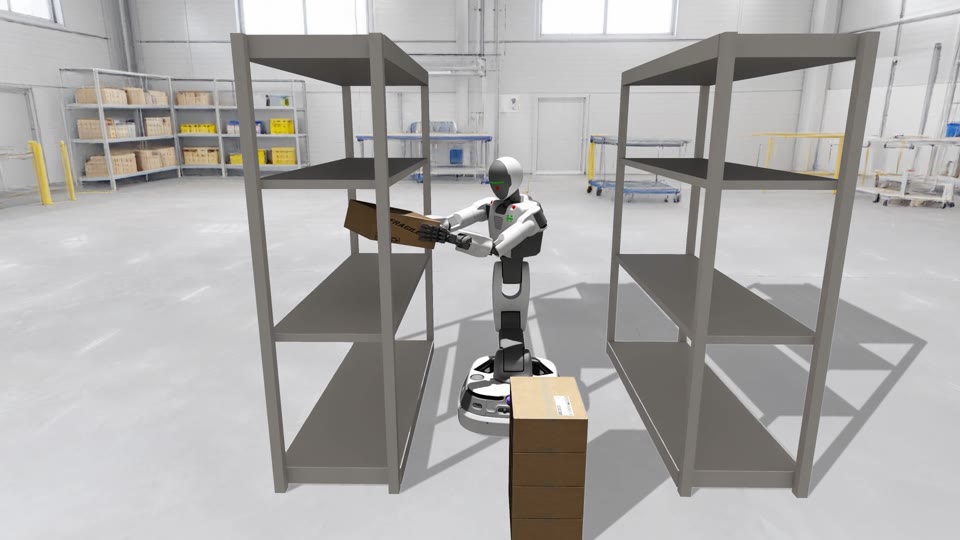}
  \caption{\friday at a shelf rack inside the splat of (a)}
\end{subfigure}
\caption{A generated background in use.
(a) Panorama produced by the proprietary 3D world generator, an enclosed hall with an empty central floor as the prompt requires.
(b) \friday inside the reconstructed splat of (a), replaying a recorded box-stacking motion.
The splat is rendered by the tiled camera and the simulated racks, cartons, and robot by ray tracing, and the two are composited.}
\label{fig:background-pipeline}
\end{figure}

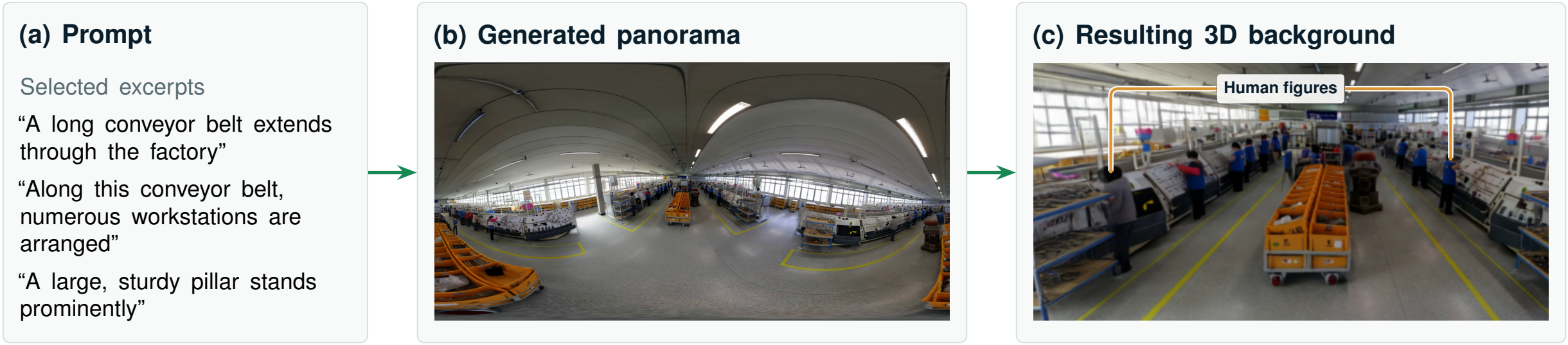
\begin{figure}[t]
\centering
\begin{tikzpicture}[
  panel/.style={draw=holidayline, rounded corners=2pt, fill=holidaypaper,
    inner sep=5pt, minimum height=35mm, anchor=north west, align=left},
  flow/.style={-{Stealth[length=2mm]}, holidaygreendark, line width=1pt}
]
\node[panel, text width=34mm] (prompt) at (0,0) {%
  {\footnotesize\bfseries\color{holidayink}(a) Prompt}\par\smallskip
  {\scriptsize\color{holidayslate}Selected excerpts}\par\smallskip
  {\scriptsize\raggedright
  ``A long conveyor belt extends through the factory''\par\smallskip
  ``Along this conveyor belt, numerous workstations are arranged''\par\smallskip
  ``A large, sturdy pillar stands prominently''\par}
};
\node[panel, text width=53mm, right=5mm of prompt.north east, anchor=north west] (panorama) {%
  {\footnotesize\bfseries\color{holidayink}(b) Generated panorama}\par\smallskip
  \includegraphics[width=\linewidth]{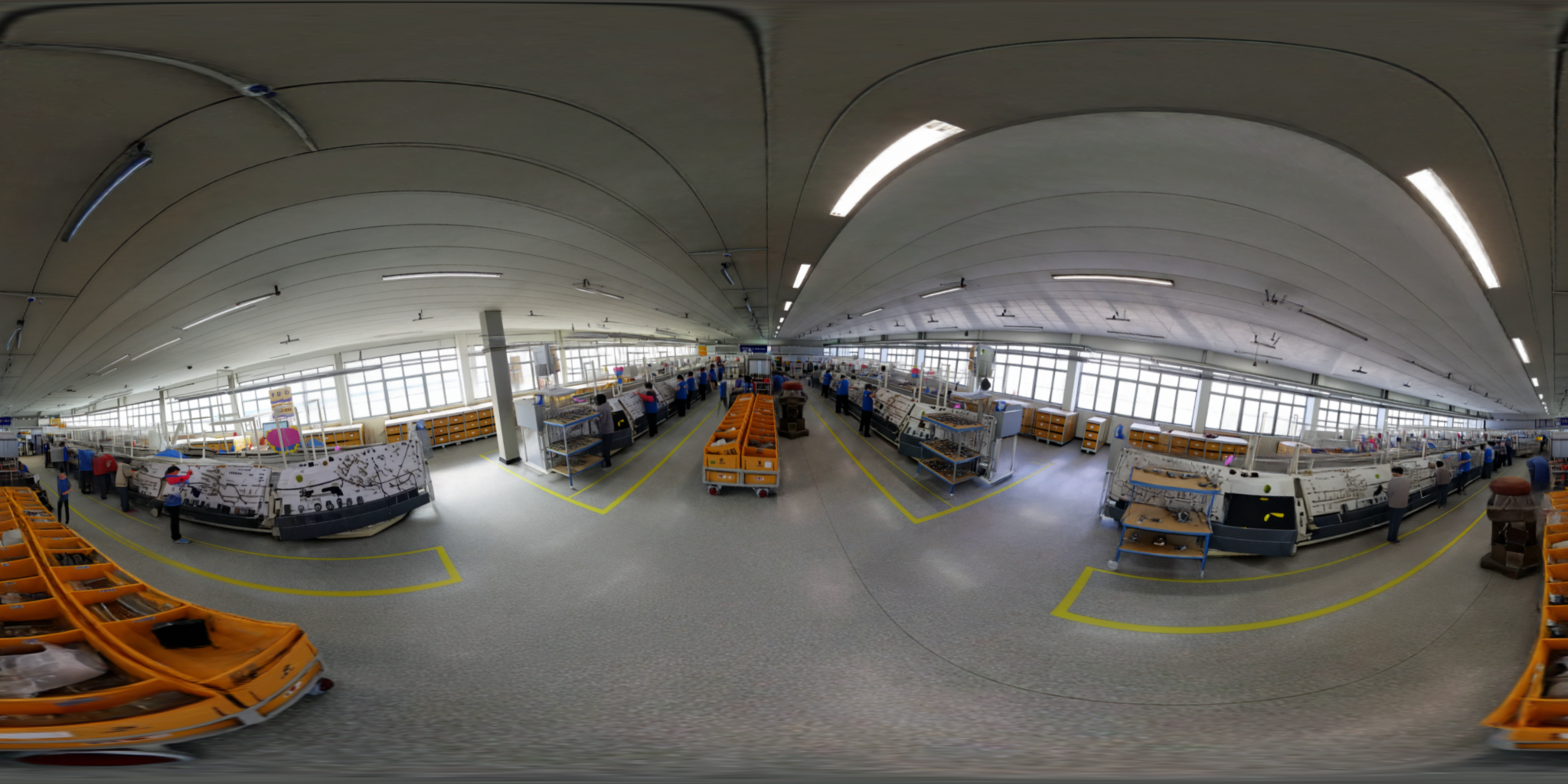}
};
\node[panel, text width=53mm, right=5mm of panorama.north east, anchor=north west] (world) {%
  {\footnotesize\bfseries\color{holidayink}(c) Resulting 3D background}\par\smallskip
  \begin{tikzpicture}[every node/.style={minimum width=0pt, minimum height=0pt,
    text width=, anchor=center, align=center}]
    \node[inner sep=0, outer sep=0, anchor=south west] (detail) at (0,0) {%
      \includegraphics[width=\linewidth,trim=180bp 46bp 300bp 100bp,clip]{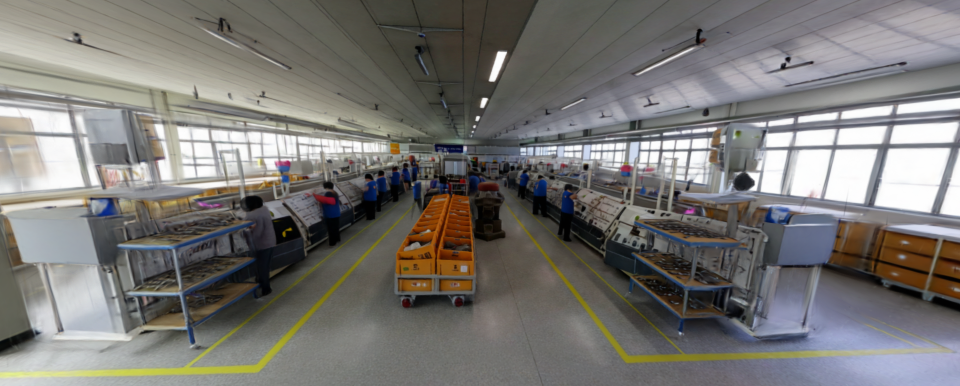}};
    \begin{scope}[x={(detail.south east)},y={(detail.north west)}]
      \node[fill=white, fill opacity=0.9, text opacity=1, rounded corners=1pt,
        inner sep=2pt, font=\tiny\bfseries, text=holidayink] (people) at (0.48,0.90) {Human figures};
      \draw[white, line width=1.6pt] (people.west) -- (0.15,0.90) -- (0.15,0.58);
      \draw[white, line width=1.6pt] (people.east) -- (0.81,0.90) -- (0.81,0.63);
      \draw[holidayamber, line width=0.8pt] (people.west) -- (0.15,0.90) -- (0.15,0.58);
      \draw[holidayamber, line width=0.8pt] (people.east) -- (0.81,0.90) -- (0.81,0.63);
      \fill[holidayamber] (0.15,0.58) circle[radius=0.6pt];
      \fill[holidayamber] (0.81,0.63) circle[radius=0.6pt];
    \end{scope}
  \end{tikzpicture}
};
\draw[flow] (prompt.east) -- (panorama.west);
\draw[flow] (panorama.east) -- (world.west);
\end{tikzpicture}
\caption{A background-generation failure relative to the benchmark requirements.
Selected prompt excerpts specify a factory with a conveyor, numerous workstations, and a prominent pillar.
The generated panorama and the resulting background contain human figures and central equipment, conflicting with the no-people and open-center requirements for work-cell placement.
Panel (c) is cropped and enlarged, and amber markers indicate two human figures.}
\label{fig:background-failure}
\end{figure}

The procedure in \autoref{alg:background} combines background generation, post-processing, work-cell placement, and verification.
\autoref{tab:app-bg-models} lists the models of the proprietary and open-source approaches compared in \S~\ref{sec:photoreal-background}.
The presets organize the factors of \autoref{tab:bg-design-space} into plausible facility types and allow variation within each preset.
The presets cover customer industries that IFR tracks for industrial robots, namely electronics, automotive, metal and machinery, and food and beverage~\citep{ifr2025worldrobotics}.
The factors vary lighting, surface appearance, and surrounding objects because visual changes can affect manipulation-policy performance~\citep{xie2024generalization}.
\autoref{fig:background-pipeline} shows a generated background in use, and \autoref{fig:background-failure} shows one that fails because it places people and equipment where the work cell must go.

\begin{table}[H]
\centering\small
\caption{Background-generation models. gpt-5.6-sol was the Codex default model at the time of the background study.}
\label{tab:app-bg-models}
\begin{tabularx}{\linewidth}{@{}>{\raggedright\arraybackslash}p{0.22\linewidth}>{\raggedright\arraybackslash}p{0.26\linewidth}Y@{}}
\toprule
\tablehead{Stage} & \tablehead{Model} & \tablehead{Parameters}\\
\midrule
\rowcolor{holidaymint}
\multicolumn{3}{@{}l}{\emph{Shared}}\\
Reference image & gemini-3.1-flash-image & Optional, 16:9, 1K, at most three attempts, not used in the paired comparison\\
Image and view checks & gpt-5.6-sol (medium) & Panorama with four $90^\circ$ crops, five renders of the splat\\
Rubric judge & gpt-5.6-sol (medium) & Nine matched frames for each background in the paired comparison, four criteria scored 1 to 5, three shuffled repeats, pairwise comparison in both orders\\
\rowcolor{holidaymint}
\multicolumn{3}{@{}l}{\emph{Proprietary}}\\
Background & marble-1.1-plus & Image input, seed 17, full resolution, 4.32\,M Gaussians\\
\rowcolor{holidaymint}
\multicolumn{3}{@{}l}{\emph{Open-source}}\\
Panorama & Matrix-3D image-to-panorama~\citep{yang2025matrix3d}, FLUX.1-Fill-dev, WorldGen LoRA~\citep{xie2025worldgen} & $2048\times1024$, 50 steps, guidance 30, $100^\circ$ field of view, seed 7341\\
Multi-view & HY-World 2.0 WorldStereo & Three views, forward, left, and right paths, 8 kept and 1 rejected, 262 cameras\\
3DGS & HY-World 2.0 trainer & 4{,}000 steps, seed 29211, $10^6$ initial points, degree-0 color, depth and normal losses, 3.78\,M Gaussians, scale 0.631\\
\rowcolor{holidaymint}
\multicolumn{3}{@{}l}{\emph{Rendering}}\\
Robot camera & Newton tiled camera & Fast Gaussian mode, transmittance 0.49, at most 20 hits per ray\\
Figures and checks & RTX path & 4 samples per pixel, denoising, 64 warm-up frames\\
\bottomrule
\end{tabularx}
\end{table}

\begin{table}[H]
\centering\small
\caption{Design space of background generation: 17 factors, with the number of options in parentheses. Presets: modern general, electronics, food, cleanroom, automotive, machining, logistics hybrid, and small automated.}
\label{tab:bg-design-space}
\begin{tabularx}{\linewidth}{@{}>{\raggedright\arraybackslash}p{0.2\linewidth}Y@{}}
\toprule
\tablehead{Group} & \tablehead{Factors}\\
\midrule
Structure & Ceiling (5), floor (6), wall (6), window (5), door (5)\\
Lighting & Lighting (5), ceiling equipment (4)\\
Perimeter & Conveyor (5), shelving (5), cart (4), boxes (5), production equipment (6), safety (5)\\
Arrangement & Layout (5), density (3), condition (5), palette (7)\\
\bottomrule
\end{tabularx}
\end{table}

\begin{algorithm}[t]
\caption{Background generation and placement}
\label{alg:background}
\small
\begin{algorithmic}[1]
\Require Preset or prompt $c$, a 3D scene reconstruction pipeline, footprint samples $\mathcal F$, recorded work-cell rollout $r$ with camera configuration, attempts $R_{\mathrm{bg}}$
\Ensure Placed splat $(\mathcal G,\theta^\star)$, or the last failed check
\Statex A failed check ends the attempt at its stage. Stages are ordered Panorama, Lift, PostProcess.
\State $s \gets \textsc{Panorama}$ \Comment{stage to rerun}
\For{$k=1,\ldots,R_{\mathrm{bg}}$}
  \State \textbf{if} $s = \textsc{Panorama}$ \textbf{then} $u \gets \Call{Prompt}{c}$; \ $I \gets \Call{Panorama}{u}$ \Comment{checks: requirements, image}
  \State \textbf{if} $s \le \textsc{Lift}$ \textbf{then} $\mathcal G \gets \Call{Lift}{I}$ \Comment{3D scene reconstruction}
  \State $\mathcal G \gets \Call{PostProcess}{\mathcal G}$ \Comment{check: background views}
  \State $\theta^\star \gets \Call{PlaceWorkCell}{\mathcal G,\mathcal F}$ \Comment{footprint clearance (\S~\ref{app:bg-config})}
  \State \Call{VerifyIntegratedViews}{$\mathcal G,\theta^\star,r$} \Comment{checks: coverage, stability}
  \State \textbf{if} every check passes \textbf{then return} $(\mathcal G,\theta^\star)$
  \State $f \gets$ the first failed check; \ $s \gets \Call{Route}{f}$ \Comment{alignment or outliers rerun PostProcess. Inconsistent views rerun Lift with new input. A failed image check, no free floor, or missing coverage rerun Panorama with a new source. A stability failure fixes the renderer instead}
  \State \textbf{if} $f$ has failed twice with the same input \textbf{then break} \Comment{the candidate is closed}
\EndFor
\State \Return the last failed check
\end{algorithmic}
\end{algorithm}

\begin{figure}[t]
\centering
\begin{tikzpicture}[
  panel/.style={draw=holidayline, rounded corners=2pt, fill=holidaypaper,
    inner sep=5pt, minimum height=35mm, text width=48mm,
    anchor=north west, align=left}
]
\node[panel] (exterior) at (0,0) {%
  {\footnotesize\bfseries\color{holidayink}(a) Exterior view}\par\smallskip
  \includegraphics[width=\linewidth]{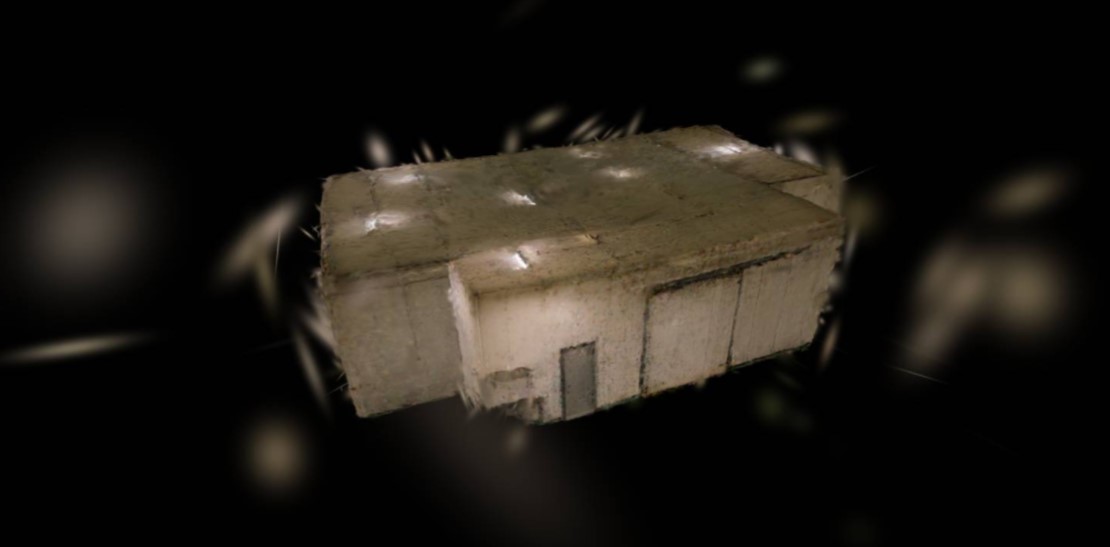}
};
\node[panel, right=4mm of exterior.north east, anchor=north west] (enclosure) {%
  {\footnotesize\bfseries\color{holidayink}(b) Estimated enclosure}\par\smallskip
  \includegraphics[width=\linewidth]{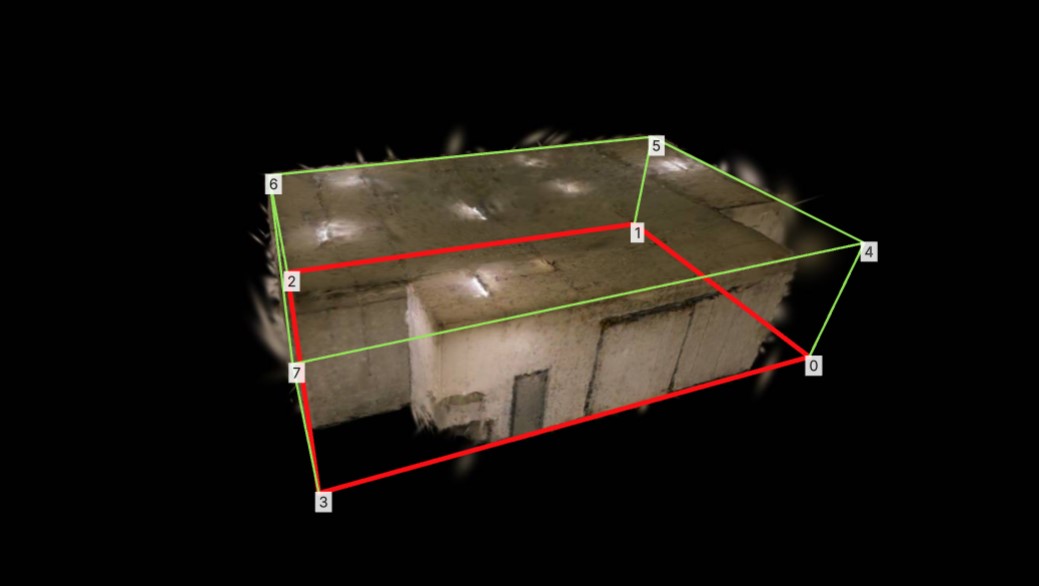}
};
\node[panel, right=4mm of enclosure.north east, anchor=north west] (region) {%
  {\footnotesize\bfseries\color{holidayink}(c) Placement region}\par\smallskip
  \includegraphics[width=\linewidth]{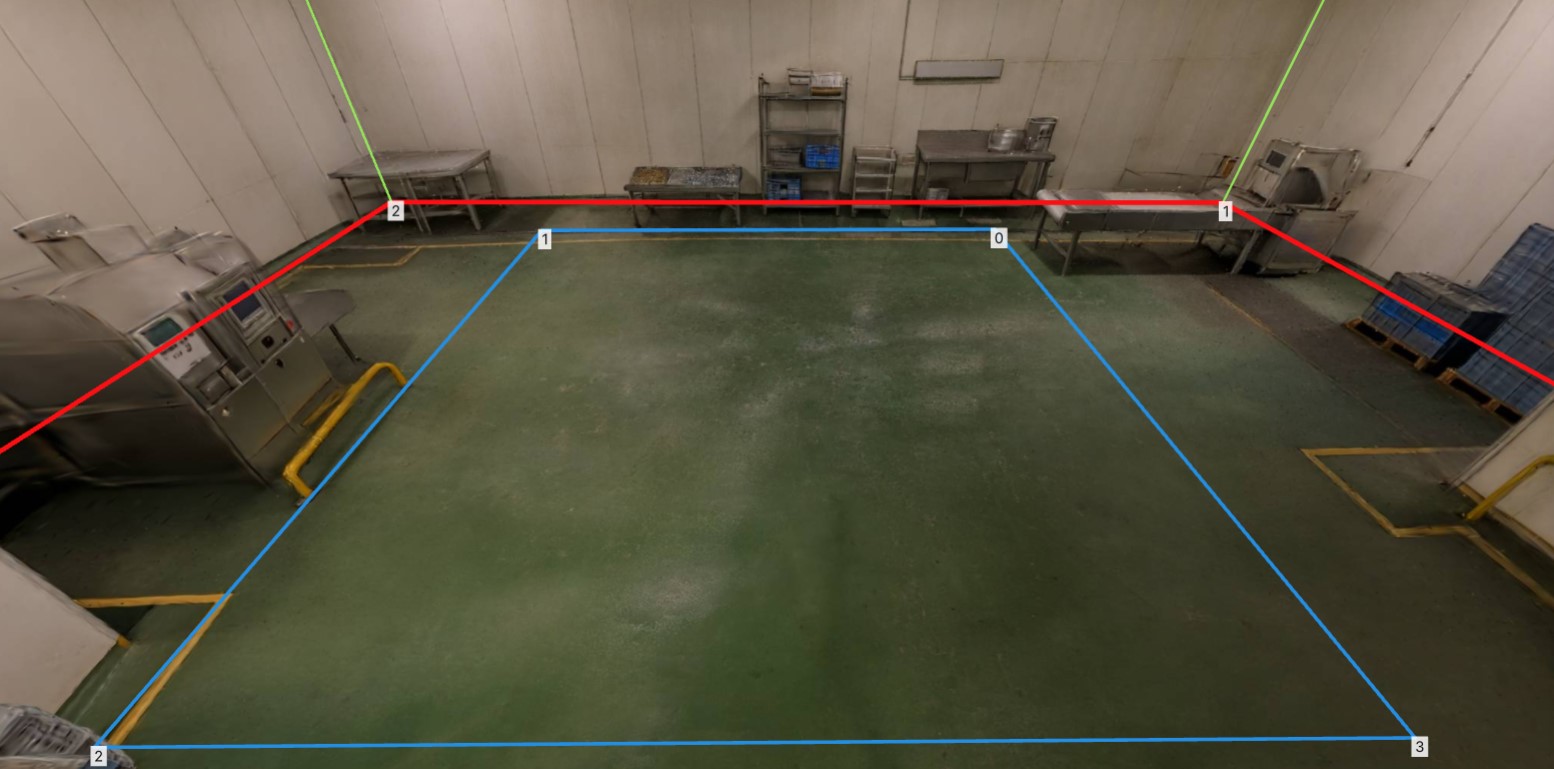}
};
\end{tikzpicture}
\caption{Preparation of generated backgrounds for work-cell placement.
(a) Exterior view of a generated Gaussian splat.
(b) Estimated room enclosure, shown by the red and green overlay.
(c) Interior view with the identified placement region, outlined in blue.
The estimated enclosure provides the geometric reference for coordinate alignment and out-of-bounds filtering, and the floor rectangle identifies the region available for work-cell placement.}
\label{fig:background-preparation}
\end{figure}

\paragraph{Post-processing a generated splat}
We estimate the room boundaries as a six-plane enclosure (\autoref{tab:gs-postprocess} and \autoref{fig:background-preparation}).
Floor and ceiling are fitted by angle-constrained RANSAC~\citep{fischler1981ransac}.
Each wall is found by scanning inward and measuring, at each scan position $c$, the fraction of the cross-section that Gaussians block, $B(c)=\frac{1}{K}\sum_{k=1}^{K}(1-e^{-\tau_k(c)})$ over $K$ cross-section samples with optical depth $\tau_k(c)$, so that large, opaque Gaussians count more than small, faint ones.
A Gaussian is removed when its one-sigma extent lies beyond face $j$ by more than a margin $m_j$,
\begin{equation}
  \mathbf n_j^{\top}\boldsymbol\mu_i-\left\lVert \mathrm{diag}(\mathbf s_i)\,R_i^{\top}\mathbf n_j\right\rVert_2-c_j>m_j .
  \label{eq:floater-removal}
\end{equation}
Here, $\boldsymbol\mu_i$, $\mathbf s_i$, and $R_i$ are the center, per-axis scales, and rotation of Gaussian $i$, and face $j$ is the plane $\mathbf n_j^{\top}\mathbf x=c_j$ with outward unit normal $\mathbf n_j$.
For each face, we also record whether it is observed or, when detection fails, placed at the 1st or 99th percentile of the Gaussian positions along its normal.

\begin{table}[H]
\centering\small
\caption{Post-processing of a generated splat, with default parameters.}
\label{tab:gs-postprocess}
\begin{tabularx}{\linewidth}{@{}>{\raggedright\arraybackslash}p{0.18\linewidth}YY@{}}
\toprule
\tablehead{Step} & \tablehead{Method} & \tablehead{Defaults}\\
\midrule
Faces & Floor and ceiling by RANSAC on the lowest and highest deciles, walls by blocked fraction $B(c)$ & Normal within $20^\circ$, $\max_c B(c)\ge 0.13$\\
Alignment & Floor to $z=0$, yaw from wall azimuths folded by $90^\circ$ & --\\
Floaters & Eq.~\eqref{eq:floater-removal} & $m_j$: 0.2 (floor, ceiling), 0.1 (walls) of center distance\\
Placement region & Largest free rectangle, with the origin moved to its center & Band 0.2--1.2\,m, wall margin 1\,m\\
\bottomrule
\end{tabularx}
\end{table}

\paragraph{Work-cell placement}
We build a 0.25\,m occupancy grid from opaque splats 0.15 to 2.5\,m above the floor, treating unscanned space as occupied.
Candidate work-cell positions are sampled on a 0.5\,m grid with rotations in $90^\circ$ increments.
For each candidate, we measure the distance to the nearest occupied grid cell at sampled points covering the work-cell footprint, including the robot's spawn poses and a 1.5\,m margin.
We reject candidates with any sample in occupied space and select the remaining candidate with the largest minimum distance.
If no candidate remains, placement fails.
For the variants of one scenario, we keep several well-separated work-cell poses with at least 0.5\,m clearance, so they occupy different parts of the same building.

\paragraph{Packaging scene for the comparison}

\label{par:bg-comparison}
A person supplies the following brief for the packaging scene used in the background study (\autoref{fig:gs-worlds}).

\begin{promptbox}{Packaging-cell brief}
A home-appliance packaging cell with an infeed conveyor of boxed washing machines, two wooden pallets, a flow rack of small-parts totes, a packing table, and a cart. Leave a clear 1.5\,m aisle for a mobile manipulator.
\end{promptbox}

The agent searches the catalog, writes the scene specification, applies static verification, and repairs it at most twice. A person reviews the scene in place of model critics.
This scene is used only to study backgrounds, because its bounding-box checks flag 2\,mm gaps as overlaps.
The full results of the comparison in \S~\ref{sec:photoreal-background} are reported in \autoref{tab:bg-paths}. The rubric judgments are post hoc and do not decide installation.

\begin{table}[t]
\centering\small
\caption{Proprietary and open-source backgrounds generated from the same reference image and compared on nine matched frames. Rubric scores come from a blind VLM judge over three shuffled repeats, and pairwise preference counts wins over both presentation orders. Better values are in bold. Stability is the temporal-stability score (limit 1.0, lower is better). The proprietary splat has 4.32\,M Gaussians and is kept in the background catalog, and the open-source splat has 3.78\,M and is not.}
\label{tab:bg-paths}
\begin{tabular*}{\linewidth}{@{\extracolsep{\fill}}lcccccccc@{}}
\toprule
 & \multicolumn{5}{c}{\tablehead{Rubric score (1 to 5)}} & & & \\
\cmidrule(lr){2-6}
\tablehead{Background} & \tablehead{Mean} & \tablehead{Detail} & \tablehead{Integrity} & \tablehead{Realism} & \tablehead{Consistency} & \tablehead{\makecell{Preferred\\(of 18)}} & \tablehead{\makecell{Usable\\(of 27)}} & \tablehead{Stability}\\
\midrule
Proprietary API & \textbf{2.85} & \textbf{3.00} & \textbf{2.67} & \textbf{2.41} & \textbf{3.33} & \textbf{18} & 0 & \textbf{0.37}\\
Open-source chain & 1.78 & 1.67 & 1.78 & 1.33 & 2.33 & 0 & 0 & 0.98\\
\bottomrule
\end{tabular*}
\end{table}

\subsection{Re-rendering recorded trajectories}
\label{app:rendering}
Render targets grow with simulation instances, cameras, and background variants, so the chunk size is fixed before allocation, instead of shrinking the batch after an out-of-memory failure.
Variants that share their objects and robot and differ only in background or lighting share $F_{s,\ell}$ and differ only in $O_s$, so a recorded trajectory can be re-rendered under each of them without rerunning the policy (\S~\ref{sec:overview-preliminaries}). Demonstrations are replayed in splat renderers in the same way~\citep{kim2026legs}.
We choose the number of variants per chunk using pixel and video-writer budgets.
For $N_{\mathrm{env}}$ instances and $C$ cameras with image width $w_c$ and height $h_c$, each variant requires $p=N_{\mathrm{env}}\sum_{c=1}^{C} w_c h_c$ pixels and $N_{\mathrm{env}}C$ writers, giving a chunk size of
\begin{equation}
  B=\max\!\left(1,\ \min\!\left(\left\lfloor P_{\max}/p\right\rfloor,\ \left\lfloor K/(N_{\mathrm{env}}C)\right\rfloor\right)\right),
  \label{eq:render-chunk}
\end{equation}
where $P_{\max}=16$ megapixels by default and $K=256$.
If a single variant requires more than $K$ writers, rendering stops before replay. If it exceeds the pixel budget, the renderer warns and proceeds with a chunk of one variant, which can still exhaust GPU memory.
Each chunk replays the trajectory once and releases its sensors before the next begins.
For 32 variants, chunks of eight reduce peak memory from 15.9 to 6.4\,GiB at 7\% lower throughput (\autoref{tab:render-memory}).

\begin{table}[t]
\centering\small
\caption{Chunking reduces the measured peak GPU memory of batched re-rendering. A 24-frame recording with one simulation instance and three cameras is re-rendered with a different lighting environment per variant on a 24\,GB GPU.}
\label{tab:render-memory}
\begin{tabular}{@{}rrrr@{}}
\toprule
\tablehead{Variants} & \tablehead{Per chunk} & \tablehead{Peak memory (GiB)} & \tablehead{Images/s}\\
\midrule
4 & 4 & 4.5 & 129.0\\
16 & 16 & 9.8 & 135.0\\
32 & 32 & 15.9 & 140.6\\
32 & 8 & \textbf{6.4} & 130.3\\
\bottomrule
\end{tabular}
\end{table}

\subsection{Process logic representation}
\label{app:process-logic}
The scenario-authoring agent writes the four typed parts of process logic $\ell$ in a process-logic file using role placeholders to bind scenario entities.
An \emph{effect} is a continuous process applied at every step. The only type currently supported is a conveyor belt, whose speed events can change.
An \emph{event} pairs a trigger with one or more actions applied in order. The trigger is a predicate of \S~\ref{sec:evaluation} that fires while it holds, on its rising edge, or after it has held for a dwell time, and each action teleports a body, sets a joint target, or sets the speed of an effect.
A \emph{queue} delivers bodies first in, first out. It names its member bodies, the slots where they wait, and the region and pose where each is delivered, and it can wait for an upstream belt, delay each dispatch, and let a removed body re-enter.
An \emph{idle target} sets the initial joint target of articulated equipment, such as a press ram parked at the top of its stroke.
Changes that must happen together, such as removing a blank and placing the stamped part, form one event with two actions rather than two events with the same trigger.

\subsection{Verification thresholds across components}
\autoref{tab:app-thresholds} lists the reported thresholds and resource limits in \S~\ref{sec:world-generation}, including the checks at each stage of background generation. Deterministic checks run first, every judged criterion must pass, and these checks determine installation eligibility.

\begin{table}[H]
\centering\small
\caption{Thresholds of the checks in \S~\ref{sec:world-generation}.}
\label{tab:app-thresholds}
\begin{tabularx}{\linewidth}{@{}>{\raggedright\arraybackslash}p{0.22\linewidth}Y@{}}
\toprule
\tablehead{Check} & \tablehead{Threshold}\\
\midrule
Static geometry & Fixture bounding-box overlap and ground penetration at most 5\,mm, specified clearance of an item above its support at most 1\,cm\\
Plan & Up to $10^4$ reachable predicate states, critic score at least 0.8 with no open concern\\
Variants & At least two distinct resolved layouts when variation is requested, every variant collision-checked\\
Specification fidelity & Position within 3\,cm, contact gap within 1\,cm, penetration within 5\,mm, at most five attempts\\
Background prompt & Ten requirements present\\
Panorama geometry & Aspect ratio $2{:}1 \pm 0.01$, width at least 1{,}024 pixels, seam ratio at most 4.0, pole standard deviation at most 24 (8-bit)\\
Panorama VLM & Valid panorama, open center, plausible, no people\\
Splat & Single splat\\
View VLM & Intact geometry, open center, plausible, no people, consistent with the panorama\\
Background coverage & Missing background at most 2\% and at most 1 percentage point above the scene without background, with sharpness, contrast, and clipping checks\\
Temporal stability & Mean frame-to-frame RGB difference at most 1.0 per fixed region, region standard deviation at least 2, at least 99\% valid frames\\
Placement & Splat opacity at least 0.3, obstacle band 0.15 to 2.5\,m, grid 0.25\,m, margin 1.5\,m, step 0.5\,m, yaw in $90^\circ$ steps, extra poses at least 0.5\,m clear and separated by yaw or 3\,m\\
Collision memory & Resolution from 32 to 1\,mm, refined while a part costs at most 1\,MB per simulation instance, parts above 500\,MB per instance rejected\\
Re-rendering & Pixel budget of 16 megapixels per chunk by default, at most 256 open video writers\\
\bottomrule
\end{tabularx}
\end{table}

\subsection{Post-hoc rubric evaluation}
\label{app:rubric-prompts}
Generated scenes and generated evaluations are scored by three judges, gpt-6-astra (high), claude-opus-5-5 (high), and gemini-4-argon (high), each under three personas.
Every input is text only, with model names, paths, run identifiers, and timestamps removed and the inputs presented in shuffled order.
All judges receive identical prompts and return typed JSON with an integer score from 1 to 5 and a rationale for each criterion and overall.
Each prompt states one persona, the artifact type, and the criteria of that type with their anchors (\autoref{tab:rubric-criteria}), and asks for a rationale of one to three sentences before each score.
These judges are used only for post-hoc analysis. The criteria use the word contract for an evaluation.
\begin{promptbox}{Personas (verbatim)}
\small
\textbf{Process engineer}. You are a senior industrial process engineer with many years of experience designing and commissioning robotic work cells. You judge the artifact as you would before signing off a cell design or a cell acceptance test.\\[0.3em]
\textbf{Evaluation researcher}. You are a robot-learning evaluation researcher who designs benchmarks and success criteria for manipulation policies. You judge the artifact as you would before using it to compare policies in a paper.\\[0.3em]
\textbf{Adversarial QA reviewer}. You are an adversarial QA reviewer. Assume the artifact may be wrong and actively look for ways it is wrong: contradictions, missing pieces, loopholes, mismatches with the request. Score using the anchors below and ground each deduction in specific evidence from the text; do not deduct for speculative problems that you cannot point to in the artifact.
\end{promptbox}
\begin{table}[H]
\centering\small
\caption{Rubric criteria. Each criterion in the prompt also carries anchors for scores 1, 3, and 5.}
\label{tab:rubric-criteria}
\begin{tabularx}{\linewidth}{@{}>{\raggedright\arraybackslash}p{0.27\linewidth}Y@{}}
\toprule
\tablehead{Criterion} & \tablehead{Question to the judge}\\
\midrule
\multicolumn{2}{@{}l}{\emph{Scenes}}\\
\texttt{process\_realism} & Would this cell, its equipment, and its material flow exist on a real factory floor?\\
\texttt{task\_conditioning} & Does the scene enable exactly the requested work, no more and no less?\\
\texttt{entity\_asset\_fit} & Is each entity the right kind of thing, at the right scale, with an asset whose category matches its role?\\
\texttt{layout\_coherence} & Are positions, supports, clearances, access, and reach coherent?\\
\texttt{meaningful\_variation} & Do the variation axes match what the request asks to vary, with meaningful ranges?\\
\texttt{completeness} & Is every element and constraint named in the request present and honoured?\\
\addlinespace
\multicolumn{2}{@{}l}{\emph{Evaluations}}\\
\texttt{task\_faithfulness} & Does the contract measure what the task text asks, including every allowed route to success?\\
\texttt{requirement\_coverage} & Is each requirement owned by a predicate, event, job, metric, or acceptance rule?\\
\texttt{measurement\_validity} & Are regions, tolerances, hold durations, and thresholds consistent with the task and physically sensible?\\
\texttt{failure\_coverage} & Would the contract report failure for the plausible ways the task can go wrong?\\
\texttt{robustness} & Does the contract stay correct under noise and resist degenerate or exploiting executions?\\
\texttt{clarity} & Is the contract understandable and internally consistent, with its choices declared?\\
\bottomrule
\end{tabularx}
\end{table}

\begin{figure}[t]
\centering
\includegraphics[width=\linewidth]{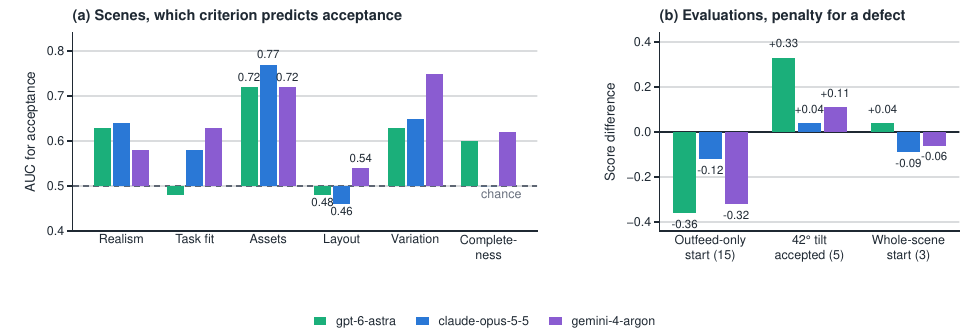}
\caption{Rubric judges, scores averaged over three personas. (a) AUC of each scene criterion for predicting acceptance of the 25 candidates by the planner's critic. (b) Overall-score difference for each defect that execution misses, with the number of defective evaluations in parentheses.}
\label{fig:rubric-judges}
\end{figure}

\paragraph{Scene results}
Plan and scene critics accept or reject candidates during generation. Rubric judges score the candidates afterward for analysis.
The pooled scores of gpt-6-astra and claude-opus-5-5 align closely with acceptance by a shared critic, but are less predictive of outcomes under the original pipelines.
Asset fit largely drives the scores, while layout coherence provides little discrimination (\autoref{fig:rubric-judges}a).
Scores cluster in a narrow range, agreement between judges is modest, and generated assets improve local fit without substantially affecting overall scores.
The third judge, gemini-4-argon, scores higher than both with a wider spread across personas and agrees more with claude-opus-5-5 than with gpt-6-astra. It also separates acceptance by asset fit and variation but barely by layout coherence.

\clearpage
\section{Evaluation generation details}
\label{app:evaluation-generation}
This appendix gives the details behind \S~\ref{sec:eval-authoring}, namely the reasons for which the compiler rejects a proposal, the three checks of verification with an audit of them, the protocol of the generation study, and examples of its feedback.

\subsection{Rejections at compilation}
\label{app:eval-rejections}
\autoref{tab:eval-rejections} groups the reasons for which the compiler of \S~\ref{sec:eval-compilation} rejects a confirmed proposal.

\begin{table}[!htb]
\centering\small
\caption{Reasons for which the compiler rejects a confirmed proposal. A rejection returns the reason to the person and produces no evaluation.}
\label{tab:eval-rejections}
\begin{tabular*}{\linewidth}{@{\extracolsep{\fill}}lp{0.74\linewidth}@{}}
\toprule
\tablehead{Group} & \tablehead{The proposal is rejected when}\\
\midrule
Names & a name is missing from the scenario, is not declared in the proposal, does not resolve to an object in the scene, or is reserved for the evaluator although the task gives it to the policy.\\
Predicates and metrics & a predicate has invalid parameters, or a metric is not one the evaluator can compute.\\
Job lifecycle & the start, completion, and failure events do not form a valid job, or the rubric of an episodic task ends on anything other than one job.\\
Authority & a visual requirement is not bound to events, a success criterion refers to a visual or unmeasured requirement, or a requirement that the task needs is labeled unmeasured.\\
\bottomrule
\end{tabular*}
\end{table}

\subsection{Verification checks}
\label{app:eval-verification}
This subsection states the three checks of \autoref{tab:eval-checks} and reports a controlled evaluation-mutation audit on the nominal first scene variant of conveyor loading.
An evaluation that fails the static check is not simulated.
The reset and idle checks share one environment per scene variant, and an evaluation is accepted when it passes all three.

\paragraph{Static check}
The check first composes the job, metrics, and rubric again, so an error in the schema, in a reference, or in the ownership of a requirement fails it.
It then tests the composed evaluation against the rules of \autoref{tab:eval-static-rules}.
A single-step completion is reported as an advisory, independently of required sustained events or protocol-violation thresholds.
The persistence needed for completion belongs to task intent and is evaluated against labeled trajectories, not inferred from a one-frame event.
Events on both edges require a rubric that tolerates no protocol violation, and a sustained event does not justify this lifecycle.
A declared justification must name a condition of the evaluation and connect the job start to a required event, so a justification cannot outlive the condition it is written for.

\begin{table}[!htb]
\centering\small
\caption{Rules of the static check. Reported conditions appear in the result of the check but do not reject the evaluation.}
\label{tab:eval-static-rules}
\begin{tabularx}{\linewidth}{@{}>{\raggedright\arraybackslash}p{0.22\linewidth}YY@{}}
\toprule
\tablehead{Rule} & \tablehead{Rejects when} & \tablehead{Admits or only reports when}\\
\midrule
Single-step completion & never & the completion event holds for one step\\
\addlinespace
Events on both edges & a predicate has events on its rising and its falling edge, or its declared justification names no condition of the evaluation & the rubric tolerates no protocol violation\\
\addlinespace
Nonpositive drop threshold & a drop predicate uses a height threshold $z \leq 0$ and no region failure is declared & the evaluation declares it a guard only, which is reported\\
\addlinespace
Time limit beyond the run & the job timeout or the stall limit exceeds the length of the run & the limit equals the length of the run, which is reported\\
\addlinespace
No failure event & never & no failure event is declared, which is reported\\
\bottomrule
\end{tabularx}
\end{table}

\paragraph{Reset check}
The check runs 4\,s of physics with the robot uncontrolled, at four substeps, in every scene variant.
It runs once with every body at its nominal position and once at each corner of the randomization range of body positions, and the axes that it leaves at nominal are listed in its result.
It covers the units that the job names and, when a queue delivers the units as in conveyor loading, the units that the queue delivers.
It fails when a unit expected to start a job is outside its start condition at reset, when the queue delivers no unit at reset, or when a unit leaves its start condition during the 4\,s.
For each unit initially inside the start condition, the implementation records the first step at which it becomes inactive, if any.

\paragraph{Idle check}
In the same environment, each evaluation then runs from reset to its termination with no action, at its time limit or its stall limit.
The check fails when a job completes, when the run passes the success criterion, or when a start event fires on a job that is already open.
When the units miss their start condition at reset, the idle run is not judged, because the reset check already reports it.

\paragraph{Controlled evaluation mutations}
The audit in \autoref{tab:eval-check-mutations} changes evaluation documents while keeping the scene, robot, and physics fixed. The control is an evaluation written for the audit rather than the benchmark evaluation of conveyor loading. It uses a 3,600-step fixed-duration run at 60\,Hz and requires at least one completed job with zero protocol violations. No policy or model is run. The first 240 steps of the passive rollout provide the 4\,s reset check, and the same rollout continues to 60\,s for the idle check, with four physics substeps.
The dual-edge test uses a separate static control whose completion requires two frames and whose rubric allows at most one protocol violation, which removes the zero-violation waiver. The mutant adds the falling start-predicate event. The static control passes and the mutant fails the rule on events at both edges. This control is not submitted for physical acceptance.

\begin{table}[t]
\centering\small
\caption{Verification outcomes on the nominal first conveyor-loading variant. Expected is the rejection stage specified before physics, and None predicts no rejection. NR means not run, and NJ means not judged because Reset fails, despite an executed rollout. Accepted refers only to this nominal scope. The static-only control has no final acceptance decision.}
\label{tab:eval-check-mutations}
\begin{tabularx}{\linewidth}{@{}Ylcccc@{}}
\toprule
Case & Expected & Static & Reset & Idle & Accepted\\
\midrule
Conveyor-loading control & None & Pass & Pass & Pass & Yes\\
Undeclared completion event & Static & Fail & NR & NR & No\\
Static-rule control (two-frame completion) & None & Pass & NR & NR & --\\
Add falling start-predicate event & Static & Fail & NR & NR & No\\
Add drop failure at floor ($z=0$) & Static & Fail & NR & NR & No\\
Stall limit: 3,600 to 3,601 steps & Static & Fail & NR & NR & No\\
Start region excludes presented tote & Reset & Pass & Fail & NJ & No\\
Start region: 0.2\,mm vertical slab & Reset & Pass & Pass & Pass & Yes\\
Start region widened to whole hall & Idle & Pass & Pass & Fail & No\\
Completed-job threshold: 1 to 0 & Idle & Fail & NR & NR & No\\
Upright tolerance: $10^\circ$ to $45^\circ$ & None & Pass & Pass & Pass & Yes\\
Upright tolerance: $10^\circ$ to $1^\circ$ & None & Pass & Pass & Pass & Yes\\
\bottomrule
\end{tabularx}
\end{table}

The control passes all three checks. Five mutations are rejected by the static check (including schema validation), one by the reset check, and one by the idle check. The zero-job threshold is rejected by schema validation before the idle check because acceptance requires a positive completed-job threshold. The whole-hall start region passes the static and reset checks but completes four jobs without robot action and satisfies the rubric during the 60\,s rollout, where the control completes none.
The narrow start slab passes, contrary to the predeclared reset-rejection expectation, because the presented tote is initially inside it and never leaves during the reset window. No settling departure is established, so this outcome is neither evidence of a missed violation nor a test of departure-detection sensitivity. In contrast, the distant start region rejects an initial eligibility failure, and its Idle result is excluded because the evaluation fails the reset check.
Both changes to the upright tolerance pass all three checks. Verification does not receive the tolerance that the person intends and therefore does not reject these semantic changes. This audit does not measure their verdict errors on tilted trajectories or estimate a statistical detection rate.

\paragraph{Scope}
The audit covers one nominal variant without randomization offsets and does not establish acceptance across all variants or the seven tasks.

\subsection{Generation study protocol}
\label{app:eval-generation-protocol}
The four models of \autoref{tab:eval-semantic-outcomes} run at the following effort levels, claude-opus-5-5 (high), gpt-6.1-sol (medium), gpt-6-astra (high), and gpt-6-luna (medium), and the simulated user is gpt-6.1-sol (medium).
Each model runs three fresh conversations for direct and for question-guided generation, which are repeated model calls rather than controlled random seeds.
The 24 generation trials, twelve with direct evaluation generation and twelve with question-guided generation, yield the 48 condition outcomes of the table because the first candidate of each trial also forms its condition without feedback.
The study makes 91 model calls, 18 in direct evaluation generation, 43 in question-guided generation, and 30 by the simulated user, who answers 64 questions in fifteen answer rounds and reviews fifteen proposals, approving thirteen.
In direct evaluation generation, the model receives the task instruction, neutral interface documentation, and the public scenario, and only the simulated user sees the private description of the intended task.
Neither the generating models nor the simulated user sees the held-out recordings, their labels, or any verdict.
Twenty-nine distinct executable candidates are scored on all fifteen recordings, and candidates shared between conditions are reused in the reported totals.
The recorded costs include the simulated user and the compiler diagnostics, which in this implementation carry the records of earlier calls, so one failed trial spends 247,042 input tokens in its last proposal call.

\begin{table}[!htb]
\centering\small
\setlength{\tabcolsep}{4pt}
\caption{Evaluations generated for single tote transfer, by model and condition, three trials each. Accepted counts trials whose evaluation passes the guards and the three checks on the first scene variant. TP, FN, FP, and TN count pairs of an executable evaluation and one of the fifteen labeled recordings, including evaluations that verification rejects, and a trial without an executable evaluation stays unscored.}
\label{tab:eval-semantic-outcomes}
\begin{tabularx}{\linewidth}{@{}lYccccc@{}}
\toprule
Model & Condition & Accepted & TP & FN & FP & TN\\
\midrule
claude-opus-5-5 & Direct, no feedback & 3/3 & 12 & 0 & 3 & 30\\
 & Direct, with feedback & 3/3 & 12 & 0 & 3 & 30\\
 & Questions, no feedback & 3/3 & 12 & 0 & 0 & 33\\
 & Questions, with feedback & 3/3 & 12 & 0 & 0 & 33\\
\addlinespace
gpt-6.1-sol & Direct, no feedback & 0/3 & 12 & 0 & 3 & 30\\
 & Direct, with feedback & 3/3 & 12 & 0 & 3 & 30\\
 & Questions, no feedback & 3/3 & 12 & 0 & 0 & 33\\
 & Questions, with feedback & 3/3 & 12 & 0 & 0 & 33\\
\addlinespace
gpt-6-astra & Direct, no feedback & 0/3 & 12 & 0 & 3 & 30\\
 & Direct, with feedback & 3/3 & 12 & 0 & 3 & 30\\
 & Questions, no feedback & 3/3 & 12 & 0 & 0 & 33\\
 & Questions, with feedback & 3/3 & 12 & 0 & 0 & 33\\
\addlinespace
gpt-6-luna & Direct, no feedback & 3/3 & 0 & 12 & 0 & 33\\
 & Direct, with feedback & 3/3 & 0 & 12 & 0 & 33\\
 & Questions, no feedback & 1/3 & 4 & 0 & 0 & 11\\
 & Questions, with feedback & 2/3 & 8 & 0 & 0 & 22\\
\bottomrule
\end{tabularx}
\end{table}

\subsection{Feedback examples}
\label{app:eval-feedback-examples}
These examples come from the generation study reported in \autoref{tab:eval-semantic-outcomes}, not from new trials. The excerpts below are shortened paraphrases of saved outputs and feedback. They illustrate the repair process rather than provide additional independent evidence.

\paragraph{Questions, with feedback: repairing event order}
gpt-6-luna trial 1 asks which physical condition defines completion, what tilt counts as upright, and whether evidence must persist. The simulated user answers that uprightness and physical support on the conveyor must be observed together before or at inlet entry, with a tilt of at most 10$^\circ$, and that one step of confirmation suffices. Conveyor placement alone does not complete the job. These details are supplied through the answers and are absent from the original instruction.
The first candidate is rejected for malformed predicate-parameter JSON. \autoref{tab:eval-feedback-example} shows the exchange for candidates 2 and 3.

\begin{table}[!htb]
\centering\small
\caption{Saved proposal repair in gpt-6-luna trial 1 (Questions, with feedback). Text is condensed, and field names identify the corresponding saved proposal. The first candidate, omitted from this excerpt, fails JSON validation.}
\label{tab:eval-feedback-example}
\begin{tabularx}{\linewidth}{@{}lY@{}}
\toprule
Stage & Relevant content\\
\midrule
Candidate 2 & \texttt{transfer\_complete} combines support, on-top geometry, uprightness, and inlet membership in one conjunction. \texttt{required\_before\_complete} is empty, and the separately defined \texttt{transfer\_qualification} is unused.\\
\addlinespace
Simulated-user feedback & Candidate 2 accepts qualification only at entry and not earlier, and it also completes when qualification first holds after entry. Record qualification as an event, require it before or on completion, and complete on the rising inlet-entry event.\\
\addlinespace
Candidate 3 & \texttt{transfer\_qualified} becomes a one-step rising event for the support/upright conjunction and is added to \texttt{required\_before\_complete}. \texttt{transfer\_complete} becomes inlet membership alone, with completion on its one-step rising event.\\
\bottomrule
\end{tabularx}
\end{table}

The simulated user approves candidate 3, which compiles and passes the static, reset, and idle checks on the nominal first variant. It produces TP=4, FN=0, FP=0, and TN=11 on the frozen recordings. The temporal repair is requested by simulated-user review, not by the three verification checks or by held-out verdict feedback.

\paragraph{Feedback does not guarantee acceptance}
In gpt-6-luna trial 2, the simulated user approves candidates 2 and 3, but both retain a required pickup condition labeled unmeasured. The compiler returns \texttt{required\_unmeasured} for \texttt{pickup\_unmeasured}. The three-candidate budget ends without an executable evaluation. This is the failed Questions trial in \autoref{tab:eval-semantic-outcomes}, whose fifteen recording pairs remain unscored.

\clearpage
\section{\benchmark details}
\label{app:benchmark-details}
This appendix gives the settings, the evaluations, and the reporting behind \S~\ref{sec:tasks}.

\subsection{Task selection}
\label{app:task-selection}
The work behind the seven tasks employs millions of people in each of the five largest robot markets (\autoref{tab:task-regions}), with robot data from the IFR~\citep{ifr2026worldrobotics,ifr2026us,ifr2026eu,ifr2026density} and employment from national labor surveys~\citep{eurostat2026lfs,stat2026lfsjp,mods2026lfskr,bls2026oews} and the 2020 census of China~\citep{nbs2022census}.
Occupation groups differ between countries. For example, Japan's handling group includes cleaning and South Korea's counts all elementary work, so the table supports which tasks enter the benchmark but does not weight them.

\begin{table}[H]
\centering\footnotesize
\caption{Industrial robots and matching workers by market. Installations are for 2025, density is robots per 10,000 manufacturing employees in 2024, and workers are counted in the national occupation groups closest to handling and to production, with their share of employment.}
\label{tab:task-regions}
\begin{tabular*}{\linewidth}{@{\extracolsep{\fill}}lrrrr@{}}
\toprule
\tablehead{Region} & \tablehead{Robots installed} & \tablehead{Robot density} & \tablehead{Handling workers} & \tablehead{Production workers}\\
\midrule
China & 354{,}000 & 166 & 6.1\% & 15.8\%\\
European Union & 60{,}500 & 231 & 5.33\,M (2.6\%) & 5.90\,M (2.8\%)\\
Japan & 36{,}200 & 446 & 5.05\,M (7.4\%) & 8.44\,M (12.4\%)\\
South Korea & 30{,}000 & 1{,}220 & 3.92\,M (13.6\%) & 2.94\,M (10.2\%)\\
United States & 38{,}400 & 307 & 2.95\,M (1.9\%) & 8.57\,M (5.5\%)\\
\bottomrule
\end{tabular*}
\end{table}

\subsection{Robot and simulation settings}

\label{app:robot-simulation}
\friday is the wheeled humanoid robot used throughout this paper, simulated from a single USD description (\autoref{tab:robot-specifications}).

\begin{table}[H]
\centering\small
\caption{\friday robot model. Counts exclude fixed joints and the floating base.}
\label{tab:robot-specifications}
\begin{tabularx}{\linewidth}{@{}>{\raggedright\arraybackslash}p{0.2\linewidth}Y@{}}
\toprule
\tablehead{Component} & \tablehead{Description}\\
\midrule
Mobile base & Differential drive, two driven wheels on prismatic suspension, four passive casters\\
Torso & Five-joint waist\\
Arms & Two arms with seven revolute joints each\\
Hands & Two five-finger hands with 20 joints each\\
Head & Three joints\\
Active joints & 64 (2 wheels, 5 waist, 14 arm, 40 hand, and 3 head joints)\\
Passive joints & 11 (2 suspension and 9 caster joints)\\
Overall & 75 joints in total, about 1.8\,m tall in the default configuration\\
Cameras & Head and wrist cameras\\
Base sensors & IMU, LiDAR, and odometry\\
Joint telemetry & Joint state\\
Tactile sensing & Magnetic tactile skin on the palms and inner finger surfaces\\
\bottomrule
\end{tabularx}
\end{table}

The robot's magnetic tactile skin produces Hall field vectors from contact-induced magnet displacements (\S~\ref{sec:tactile} and \S~\ref{app:tactile-kernel}).
These readings are available to a policy but are not used by the benchmark's success criteria, which are evaluated from privileged simulator state.

\autoref{tab:simulation-configuration} lists the default timing and actuation settings for \friday. All actuator controllers produce motor-current commands. Joint-specific parameters are defined in the robot configuration, and individual experiments specify any overrides.

\begin{table}[H]
\centering\small
\caption{Default simulation and actuation settings for \friday.}
\label{tab:simulation-configuration}
\begin{tabularx}{\linewidth}{@{}>{\raggedright\arraybackslash}p{0.30\linewidth}Y@{}}
\toprule
\tablehead{Setting} & \tablehead{Configuration}\\
\midrule
Frame/control period & $1/60$\,s\\
Control interface & Joint position and velocity targets\\
Physics substeps & 4 per frame ($1/240$\,s each)\\
Evaluation rate & 60\,Hz\\
\bottomrule
\end{tabularx}
\end{table}

\subsection{Evaluations of the seven tasks}
\label{app:task-specs}
This subsection states the predicates (\autoref{tab:predicates}), the margins, the temporal operators, and the job semantics behind \S~\ref{sec:evaluation}, and then the evaluation of each task in \autoref{tab:task-events}.

\begin{table}[!htb]
\centering\small
\caption{Predicate vocabulary. All predicates read privileged state and return one value per simulation instance.}
\label{tab:predicates}
\begin{tabularx}{\linewidth}{@{}>{\raggedright\arraybackslash}p{0.16\linewidth}Y@{}}
\toprule
\tablehead{Group} & \tablehead{Predicates}\\
\midrule
Object state & In a region, upright within a tilt bound, below a height, at rest\\
Relation & Touching, supported by, on top of, next to, stable on (on top of, supported by, and at rest)\\
Joint & Within a range, completed a stroke\\
Composition & Conjunction\\
\bottomrule
\end{tabularx}
\end{table}

\paragraph{State and margins}
Let $x_t$ be the privileged state at step $t$ of a run sampled at $\Delta t=1/60$\,s. The default $N_{\mathrm{sub}}=4$ physics substeps per frame have duration $h=\Delta t/N_{\mathrm{sub}}=1/240$\,s. For a run of $N_{\mathrm{step}}$ frames, $t_{\mathrm{run}}=N_{\mathrm{step}}\Delta t$, which can be shorter than $T_{\mathrm{run}}$.
A unit $u$ has pose $(\mathbf{p}_u,\mathbf{R}_u)\in SE(3)$ and twist $(\mathbf{v}_u,\boldsymbol{\omega}_u)$, and joint $j$ has position $q_j$.
A region $\mathcal{R}=(b,\mathbf{c},\mathbf{h})$ is a box with center $\mathbf{c}$ and half extent $\mathbf{h}\in\mathbb{R}^3_{>0}$ in the frame of a body $b$ with pose $(\mathbf{p}_b,\mathbf{R}_b)$, the world for fixed regions and a jig or a tote for moving ones.
The margins are
\begin{align}
  \mu_{\mathsf{in}}(u,\mathcal{R}) &= \min_{i\in\{1,2,3\}}\Big(h_i-\big|\mathbf{e}_i^{\top}\big(\mathbf{R}_b^{\top}(\mathbf{p}_u-\mathbf{p}_b)-\mathbf{c}\big)\big|\Big), &
  \mu_{\mathsf{up}_{\alpha}}(u) &= \alpha-\arccos\big(\mathbf{e}_z^{\top}\mathbf{R}_u\mathbf{e}_z\big), \nonumber\\
  \mu_{\mathsf{st}}(u) &= \min\big(\epsilon_v-\|\mathbf{v}_u\|,\ \epsilon_\omega-\|\boldsymbol{\omega}_u\|\big), &
  \mu_{[a,b]}(q_j) &= \min\big(q_j-a,\ b-q_j\big),
  \label{eq:predicate-margins}
\end{align}
with the standard basis $\mathbf{e}_1,\mathbf{e}_2,\mathbf{e}_3=\mathbf{e}_z$, tilt bound $\alpha$, $\epsilon_v=0.02$\,m/s, $\epsilon_\omega=0.1$\,rad/s, and $a=-\infty$ or $b=\infty$ for one-sided joint bounds.
Each atomic predicate thresholds its margin at zero, $\mathsf{in}(u,\mathcal{R})=\mathbf{1}[\mu_{\mathsf{in}}(u,\mathcal{R})\ge0]$ and likewise for $\mathsf{up}_{\alpha}$ and joint bounds, while rest is strict, $\mathsf{st}(u)=\mathbf{1}[\mu_{\mathsf{st}}(u)>0]$.
A union of regions takes the maximum of the margins and a conjunction the minimum, so composite truth values can likewise be determined by thresholding the resulting margin, retaining the strictness of the constituent tests.
For a set $U$ of bodies, $\mathsf{cnt}(U,\mathcal{R})=\sum_{u\in U}\mathsf{in}(u,\mathcal{R})$ enters through integer bounds $n\le\mathsf{cnt}(U,\mathcal{R})\le n'$.
Margins are used for their sign, and their magnitudes are not compared across predicates with different units. A geometric margin can also express distance to a boundary, for example $-0.04$\,m for a tote whose origin rests 4\,cm outside its region.

\paragraph{Temporal operators}
For a Boolean sequence $b_t$, set $b_t=0$ for $t<0$ and define the hold operator as $H_{K_{\mathrm{hold}}}b_t=\min_{0\le s<K_{\mathrm{hold}}}b_{t-s}$.
The rising event $\evup{\Delta}b$ with $\Delta=K_{\mathrm{hold}}\,\Delta t$ fires at step $t$ when $H_{K_{\mathrm{hold}}}b_t\wedge\neg H_{K_{\mathrm{hold}}}b_{t-1}$, which equals $H_{K_{\mathrm{hold}}}b_t\wedge\neg b_{t-K_{\mathrm{hold}}}$ and is what Eq.~\eqref{eq:debounce} computes, the falling event is $\evdn{\Delta}b=\evup{\Delta}(1-b)$, and the subscript is omitted for $K_{\mathrm{hold}}=1$. In the task formulas below, the temporal operator applies to the entire bracketed predicate. Durations written as 0.17\,s denote 10 evaluation steps (10/60\,s).

\paragraph{Jobs and success criterion}
A job of unit $u$ is given by a start event $e_{\mathrm{start}}$, required events $e_{\mathrm{req},1},\dots,e_{\mathrm{req},m}$, a completion event $e_{\mathrm{complete}}$, failure events $e_{\mathrm{fail},1},\dots,e_{\mathrm{fail},l}$, and a limit $\tau_{\mathrm{job}}$.
It opens at the step $t_0$ at which $e_{\mathrm{start}}(u)$ fires while no job of $u$ is open, and a start event of $u$ during an open job counts as a protocol violation.
At every later step the job fails if some $e_{\mathrm{fail},j}(u)$ fires, and otherwise completes if $e_{\mathrm{complete}}(u)$ fires and every $e_{\mathrm{req},i}(u)$ has fired since $t_0$.
A job still open when $(t-t_0)\Delta t\ge\tau_{\mathrm{job}}$ times out, and one still open when the run ends is censored.
Every job also fails when its unit remains in the floor region, $\evup{\Delta}\mathsf{in}(u,\mathcal{R}_{\mathrm{floor}})$ with $\Delta$ of 0.5 or 1\,s, or leaves the world, $z_u=\mathbf{e}_z^{\top}\mathbf{p}_u<z_{\min}$.
An event $e_{\mathrm{monitor}}$ that no job references, such as a hand in the die while the ram is away from home, can fire at any step, also after the job of its unit completes.
A metric counts its firings $N_{\mathrm{monitor}}$, as for any other event of the evaluation.
A run with $N_{\mathrm{done}}$ completed jobs ends at $N_{\mathrm{done}}=N_{\mathrm{goal}}$, at the limit $T_{\mathrm{run}}$, or after 50\,s to 3\,min without an event, and
\begin{equation}
  \mathrm{pass}\iff N_{\mathrm{done}}\ge N_{\mathrm{goal}}\ \wedge\ N_{\mathrm{fail}}=N_{\mathrm{timeout}}=N_{\mathrm{stall}}=N_{\mathrm{violation}}=0\ \wedge\ N_{\mathrm{monitor}}=0 .
  \label{eq:run-acceptance}
\end{equation}
Here, $N_{\mathrm{fail}}$, $N_{\mathrm{timeout}}$, $N_{\mathrm{stall}}$, and $N_{\mathrm{violation}}$ count failed jobs, timed-out jobs, stalls, and protocol violations.
A rubric that only reports, as in conveyor loading, has no $N_{\mathrm{goal}}$ and no success criterion, and its run ends at $T_{\mathrm{run}}$.
$N_{\mathrm{goal}}$, $T_{\mathrm{run}}$, $\tau_{\mathrm{job}}$, and the inactivity limit are parameters of the evaluation, not constants of the benchmark, and an evaluation may set its own. \autoref{tab:rubric-parameters} lists the values of the seven reference evaluations used below.

\begin{table}[!htb]
\centering\small
\caption{Rubric parameters of the seven reference evaluations. A report-only rubric has no job target and no job limit.}
\label{tab:rubric-parameters}
\begin{tabular}{@{}lccc@{}}
\toprule
\tablehead{Task} & \tablehead{$N_{\mathrm{goal}}$} & \tablehead{$T_{\mathrm{run}}$ (min)} & \tablehead{$\tau_{\mathrm{job}}$ (min)}\\
\midrule
Conveyor loading & -- & 5 & --\\
Flow-rack loading & 3 & 10 & 7\\
Cart push & 8 & 20 & 18\\
Jig loading & 8 & 20 & 18\\
Press tending & 3 & 12 & 9\\
Valve operation & 3 & 10 & 7\\
Barcode routing & 5 & 12 & 6\\
\bottomrule
\end{tabular}
\end{table}

\paragraph{Properties}
Three properties follow from these definitions.
(P1) For a fixed evaluation and initial job state, the verdict can be recomputed from timestamped events and the run termination time. The event times preserve the information needed for job timeouts and inactivity limits, and the termination time determines censoring.
(P2) A predicate excursion shorter than $K_{\mathrm{hold}}$ steps fires no event, a run of at least $K_{\mathrm{hold}}$ steps fires exactly one, and its time stamp lags the transition by exactly $K_{\mathrm{hold}}-1$ steps.
(P3) Every job ends in exactly one of the states completed, failed, timed out, or censored, a failure overrides a completion in the same step, and a job completes only after all of its required events, so a unit that reaches its destination by another route does not count.

\paragraph{Conveyor loading}
The instruction reads ``Pick up each tote presented on the pallet and place it upright on the running conveyor.''
A queue of five totes returns every tote that enters the X-ray tunnel to the pallet, so the same totes serve every job.
A tote opens a job when the queue delivers it to the pallet. It must be upright on and supported by the belt, and the job completes when the tote enters the tunnel, a single-step event because the queue recycles the tote at once.
The job has no limit $\tau_{\mathrm{job}}$ of its own, and the rubric runs for 5\,min and only reports. Single tote transfer shares this job, with a rubric that ends after one job or 1\,min and passes with one completed job and no protocol violations (\autoref{tab:eval-example}).
\begin{align*}
  e_{\mathrm{start}}(u) &= \evup{}\mathsf{in}(u,\mathcal{R}_{\mathrm{pallet}}),\\
  e_{\mathrm{req}}(u) &= \evup{}\big(\mathsf{on}(u,\mathrm{belt})\wedge\mathsf{sup}(u,\mathrm{belt})\wedge\mathsf{up}_{10^\circ}(u)\big),\\
  e_{\mathrm{complete}}(u) &= \evup{}\mathsf{in}(u,\mathcal{R}_{\mathrm{tunnel}}),
\end{align*}
where $\mathsf{on}$ and $\mathsf{sup}$ are the relations \emph{on top of} and \emph{supported by} in \autoref{tab:predicates}.

\paragraph{Flow-rack loading}
The instruction reads ``Move each container from the pallet stack into a lane of the flow rack, upright.''
A container opens a job in its slot of the pallet stack and crosses the aisle, and the job completes when the container is upright in any of the 12 lanes $\mathcal{R}_1,\dots,\mathcal{R}_{12}$.
\begin{align*}
  e_{\mathrm{start}}(u) &= \evup{0.5\mathrm{s}}\mathsf{in}(u,\mathcal{R}_{\mathrm{stack}}(u)),\\
  e_{\mathrm{req}}(u) &= \evup{0.17\mathrm{s}}\mathsf{in}(u,\mathcal{R}_{\mathrm{aisle}}),\\
  e_{\mathrm{complete}}(u) &= \evup{0.5\mathrm{s}}\bigl[\mathsf{in}\big(u,\textstyle\bigcup_{\ell=1}^{12}\mathcal{R}_{\ell}\big)\wedge\mathsf{up}_{15^\circ}(u)\bigr],\\
  e_{\mathrm{fail}}(u) &= \evdn{0.5\mathrm{s}}\mathsf{up}_{45^\circ}(u).
\end{align*}

\paragraph{Cart push}
The instruction reads ``Load the cartons from the staging tables onto a hand cart, push it to a dock bay, and clear the loading deck.''
A carton opens a job on its staging table, leaves the table, and crosses the aisle, and the job completes when the carton is upright and at rest in a dock bay and no carton of the set $U$ remains on the loading deck.
The completion condition checks that the loading deck is clear but does not verify that an empty cart has returned.
A carton that leaves its bay fires an event $e_{\mathrm{monitor}}$.
\begin{align*}
  e_{\mathrm{start}}(u) &= \evup{0.5\mathrm{s}}\mathsf{in}(u,\mathcal{R}_{\mathrm{stage}}(u)),\\
  e_{\mathrm{req},1}(u) &= \evdn{0.17\mathrm{s}}\mathsf{in}(u,\mathcal{R}_{\mathrm{stage}}(u)),\\
  e_{\mathrm{req},2}(u) &= \evup{0.5\mathrm{s}}\mathsf{in}(u,\mathcal{R}_{\mathrm{aisle}}),\\
  e_{\mathrm{complete}}(u) &= \evup{1\mathrm{s}}\bigl[\mathsf{in}(u,\mathcal{R}_{\mathrm{dock}})\wedge\mathsf{up}_{15^\circ}(u)\wedge\mathsf{st}(u)\wedge\mathsf{cnt}(U,\mathcal{R}_{\mathrm{load}})=0\bigr],\\
  e_{\mathrm{monitor}}(u) &= \evdn{0.5\mathrm{s}}\mathsf{in}(u,\mathcal{R}_{\mathrm{dock}}).
\end{align*}

\paragraph{Jig loading}
The instruction reads ``Fill the pockets of the bench jig with parts from the tray, store the full jig on the rack, and seat an empty jig.''
A part opens a job in the input tray, leaves it, and is seated in a pocket of the bench jig.
The job completes when the part's jig is stored on a rack shelf while a jig of the set $G$ is seated on the bench and no part of the set $U$ remains there.
Pockets $\mathcal{R}_{\mathrm{pocket}}$ move with their jigs. The evaluation does not enforce a pocket-filling order.
A stored part that leaves its jig, a jig on the floor, and a jig tilted beyond $30^\circ$ fire events $e_{\mathrm{monitor}}$.
\begin{align*}
  e_{\mathrm{start}}(u) &= \evup{0.5\mathrm{s}}\mathsf{in}(u,\mathcal{R}_{\mathrm{tray}}),\\
  e_{\mathrm{req},1}(u) &= \evdn{0.17\mathrm{s}}\mathsf{in}(u,\mathcal{R}_{\mathrm{tray}}),\\
  e_{\mathrm{req},2}(u) &= \evup{1\mathrm{s}}\bigl[\mathsf{in}(u,\mathcal{R}_{\mathrm{pocket}})\wedge\mathsf{in}(u,\mathcal{R}_{\mathrm{bench}})\bigr],\\
  e_{\mathrm{complete}}(u) &= \evup{1\mathrm{s}}\bigl[\mathsf{in}(u,\mathcal{R}_{\mathrm{pocket}})\wedge\mathsf{in}(u,\mathcal{R}_{\mathrm{shelf}})\wedge\mathsf{st}(u)\wedge\mathsf{cnt}(G,\mathcal{R}_{\mathrm{seat}})\ge1\wedge\mathsf{cnt}(U,\mathcal{R}_{\mathrm{bench}})=0\bigr].
\end{align*}

\paragraph{Press tending}
The instruction reads ``Load a blank into the die, start the press with both buttons, and move the stamped part to the outfeed table.''
A blank opens a job on the infeed table, is nested in the die, is pressed with both buttons while nested, and is stamped, when the press event replaces it by its stamped part $u'$.
The job completes when the origin of $u'$ remains in the outfeed region for 1\,s. This region is a 7\,cm height band above the table. The dwell is a proxy for placement at rest and does not include a velocity check.
A hand or wrist link of the set $H$ inside the die while the ram is away from home fires an event $e_{\mathrm{monitor}}$. Here, $q_{\mathrm{L}}$ and $q_{\mathrm{R}}$ are the positions of the two button joints and $q_{\mathrm{ram}}$ that of the ram.
\begin{align*}
  e_{\mathrm{start}}(u) &= \evup{0.5\mathrm{s}}\mathsf{in}(u,\mathcal{R}_{\mathrm{infeed}}),\\
  e_{\mathrm{req},1}(u) &= \evup{0.5\mathrm{s}}\mathsf{in}(u,\mathcal{R}_{\mathrm{nest}}),\\
  e_{\mathrm{req},2}(u) &= \evup{0.1\mathrm{s}}\bigl[\mathsf{in}(u,\mathcal{R}_{\mathrm{nest}})\wedge q_{\mathrm{L}}\le-19\,\mathrm{mm}\wedge q_{\mathrm{R}}\le-19\,\mathrm{mm}\bigr],\\
  e_{\mathrm{req},3}(u) &= \evup{0.5\mathrm{s}}\mathsf{stamped}(u),\\
  e_{\mathrm{complete}}(u) &= \evup{1\mathrm{s}}\mathsf{in}(u',\mathcal{R}_{\mathrm{outfeed}}),\\
  e_{\mathrm{monitor}} &= \evup{0.05\mathrm{s}}\bigl[\mathsf{cnt}(H,\mathcal{R}_{\mathrm{die}})\ge1\wedge q_{\mathrm{ram}}\le-20\,\mathrm{mm}\bigr].
\end{align*}

\paragraph{Valve operation}
The instruction reads ``Place a beaker under the spout, open the valve until it is filled, close the valve, and move the beaker to the output side.''
A beaker opens a job on the input side, and the job completes when the beaker is upright and at rest on the output side with the valve closed.
Because the simulator does not model the liquid, filling is proxied by holding the beaker upright under the spout with the valve open for 3\,s.
Opening the valve with no beaker of the set $U$ under the spout fires an event $e_{\mathrm{monitor}}$, where $q_{\mathrm{valve}}$ is the valve joint angle.
\begin{align*}
  e_{\mathrm{start}}(u) &= \evup{0.5\mathrm{s}}\mathsf{in}(u,\mathcal{R}_{\mathrm{input}}),\\
  e_{\mathrm{req}}(u) &= \evup{3\mathrm{s}}\bigl[\mathsf{in}(u,\mathcal{R}_{\mathrm{spout}})\wedge\mathsf{up}_{10^\circ}(u)\wedge q_{\mathrm{valve}}\ge60^\circ\bigr],\\
  e_{\mathrm{complete}}(u) &= \evup{1\mathrm{s}}\bigl[\mathsf{in}(u,\mathcal{R}_{\mathrm{output}})\wedge\mathsf{up}_{15^\circ}(u)\wedge\mathsf{st}(u)\wedge q_{\mathrm{valve}}\le5^\circ\bigr],\\
  e_{\mathrm{fail}}(u) &= \evdn{0.25\mathrm{s}}\mathsf{up}_{30^\circ}(u),\\
  e_{\mathrm{monitor}} &= \evup{0.1\mathrm{s}}\bigl[\mathsf{cnt}(U,\mathcal{R}_{\mathrm{spout}})=0\wedge q_{\mathrm{valve}}\ge10^\circ\bigr].
\end{align*}

\paragraph{Barcode routing}
The instruction reads ``Put each scanned parcel into the tote assigned to its barcode, and a parcel with an unreadable code into the exception tote.''
A parcel opens a job once it passes the scan gantry. It then leaves the end of the belt, and the job completes when the parcel is at rest in the tote $\mathcal{R}_{\mathrm{dest}}(u)$ that the routing table assigns to its code, the exception tote for an unreadable code.
Resting in any other tote fails the job, and a parcel that leaves its assigned tote fires an event $e_{\mathrm{monitor}}$.
\begin{align*}
  e_{\mathrm{start}}(u) &= \evup{0.5\mathrm{s}}\mathsf{in}(u,\mathcal{R}_{\mathrm{scan}}),\\
  e_{\mathrm{req}}(u) &= \evdn{0.17\mathrm{s}}\mathsf{in}(u,\mathcal{R}_{\mathrm{end}}),\\
  e_{\mathrm{complete}}(u) &= \evup{1\mathrm{s}}\bigl[\mathsf{in}(u,\mathcal{R}_{\mathrm{dest}}(u))\wedge\mathsf{st}(u)\bigr],\\
  e_{\mathrm{fail}}(u) &= \evup{1\mathrm{s}}\bigl[\mathsf{in}(u,\mathcal{R}_{\mathrm{other}}(u))\wedge\mathsf{st}(u)\bigr],\\
  e_{\mathrm{monitor}}(u) &= \evdn{0.5\mathrm{s}}\mathsf{in}(u,\mathcal{R}_{\mathrm{dest}}(u)).
\end{align*}

\paragraph{Checks with scripted trajectories}
\label{par:eval-scripted}
The evaluations of the six tasks other than conveyor loading are also tested with scripted trajectories, which stand in for a policy and are therefore not part of acceptance (\S~\ref{sec:eval-compilation}).
All 185 region checks against scene bodies match, the 20 start-state probes behave as expected, including a false result for each plausible non-start state, and 32 scripted trajectories labeled with their intended verdict reproduce all 32 labels (\autoref{tab:spec-checks}).
The replay exposes one defect that the static and reset checks miss. Carry paths that dip back into a start volume 15 to 25\,cm tall re-fire the start event and count protocol violations in nominal runs, 5 for cart push and 3 for barcode routing.
Holding every start volume within 5\,cm of the resting height and the start event for 30 frames removes all false alarms.
No policy completes the jobs yet, so the replay uses scripted trajectories instead of recorded rollouts.

\begin{table}[!htb]
\centering\small
\caption{Further verification of the evaluations of six tasks. Region checks place a scene body inside and outside each region, start-state probes move one item to a plausible non-start pose, and replay traces are scripted trajectories whose verdict is declared before the run, one success and the rest violations.}
\label{tab:spec-checks}
\begin{tabular*}{\linewidth}{@{\extracolsep{\fill}}lccc@{}}
\toprule
\tablehead{Task} & \tablehead{Region checks} & \tablehead{Start probes} & \tablehead{Replay traces}\\
\midrule
Flow-rack loading & 33/33 & 3/3 & 5/5\\
Cart push & 30/30 & 4/4 & 5/5\\
Jig loading & 25/25 & 3/3 & 5/5\\
Press tending & 23/23 & 3/3 & 6/6\\
Valve operation & 15/15 & 3/3 & 7/7\\
Barcode routing & 59/59 & 4/4 & 4/4\\
\midrule
All six & 185/185 & 20/20 & 32/32\\
\bottomrule
\end{tabular*}
\end{table}

\subsection{Visual evidence}
\label{app:visual-evidence}
All seven benchmark evaluations determine job outcomes and run verdicts from privileged simulator state.
The visual progress estimates reported below are diagnostic and carry no weight in these outcomes.
We distinguish this use from an optional evaluator extension that can include explicitly declared visual conditions in the success criterion. That extension is not used by the seven benchmark evaluations, and its visual accuracy is not established here.

\paragraph{Optional visual conditions}
The evaluator can bind a visual condition to frame windows selected by named events and combine its results with physical conditions when this additional rule is declared before execution.
This differs from the evaluation generation pipeline of \S~\ref{sec:eval-authoring}, whose compiler records visual requirements as event-bound evidence but rejects their use in a success criterion.
Physical predicates retain their simulator-derived values, and visual evidence cannot replace them.
For a visual condition, model judgments or reviewed annotations resolve the available evidence, while missing or unresolved results remain unknown.
The number of satisfying frames is therefore bounded by $[L_\kappa,U_\kappa]$, where $L_\kappa$ counts known satisfying frames and $U_\kappa$ also includes unresolved frames.
A declared threshold passes or fails only when these bounds determine the result, and otherwise it remains undecided.
For an evaluation using this extension, let $\mathcal K_{\mathrm{eval}}=\mathcal K_{\mathrm{phys}}\cup\mathcal K_{\mathrm{vis}}$ be its declared condition set. The outcomes of condition $\kappa$ and of the run are
\begin{equation}
  y_\kappa = A_\kappa\big(\operatorname{Aggregate}(\mathcal{D}(x_{0:N_{\mathrm{step}}}))\big)\ \ (\kappa\in\mathcal{K}_{\mathrm{phys}}),\qquad
  y_\kappa = A_\kappa\big([L_\kappa,U_\kappa]\big)\ \ (\kappa\in\mathcal{K}_{\mathrm{vis}}),\qquad
  Y = \min_{\kappa\in\mathcal{K}_{\mathrm{eval}}} y_\kappa,
  \label{eq:eval-channels}
\end{equation}
where $\mathcal{D}$ detects events, $\operatorname{Aggregate}$ aggregates them into jobs and metrics, $A_\kappa$ applies the threshold of condition $\kappa$, and the minimum follows the order fail $<$ undecided $<$ pass.
Thus, a physical failure remains a failure, while a run that passes its physical conditions may fail or remain undecided on an additional visual condition.
This rule specifies how evidence affects the verdict, not how accurately a visual judge recognizes the condition.

\paragraph{Visual judge}
A visual judge receives one criterion and one ordered frame window, for example whether the tote remains visibly upright as it enters the X-ray tunnel, and returns satisfied or violated only when the frames show it, not visible when an occluder hides the relevant object, and unclear when the visible evidence does not resolve the criterion, with a rationale that names the visible basis rather than a reconstructed account of hidden robot actions.
Uprightness is measured from simulator state in the actual conveyor-loading evaluation, and this example does not report a validation of visual acceptance.

\begin{figure}[!htb]
\centering
\includegraphics[width=0.68\linewidth]{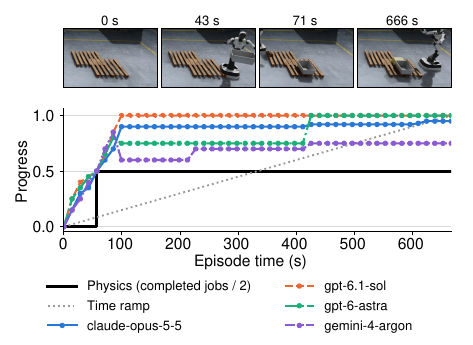}
\caption{Progress estimated by proprietary models against progress from physics on line unloading, conveyor loading in reverse.}
\label{fig:progress-calibration}
\end{figure}

\paragraph{Whole-run progress judges}
\label{par:progress-judges}
The evaluator computes progress from the number of completed jobs and, within a job, the fraction of required events that have occurred, using the event definitions in the evaluation specification without frame-level progress labels.
A vision-language model that estimates progress from frames acts as a value function~\citep{ma2024gvl,rao2026argon}, and we record such whole-run judgments beside physical progress as a diagnostic.
Following the in-context progress-judging setup of \citet{ma2024gvl}, four models assess 36 labeled runs of line unloading, a diagnostic task that reverses conveyor loading: 8 successes and 28 failures (\autoref{tab:progress-judge-line}).
Each run supplies 48 frames, and the judges return frame-level progress and an outcome using the prompt structure given at the end of this section.
We compare outcomes with the physical evaluation and progress with completed-job counts and an elapsed-time baseline.
The three models gpt-6.1-sol, gpt-6-astra, and claude-opus-5-5 call the same 8 failures successful and agree with one another on at least 34 of the 36 outcomes.
gemini-4-argon calls 7 failures successful, 6 of them among those 8. It rejects a tote left on its side and the upside-down tote of \autoref{fig:progress-calibration}, which the others accept, and accepts a tote wedged at $42^\circ$, which they reject.
All four agree on 32 of the 36 outcomes, so their errors are largely but not entirely shared.
Across the four models, 25--29\% of failed runs are judged successful. Their progress estimates correlate with completed-job counts more strongly than elapsed time does for three of the four judges, but whole-run visual judgments cannot substitute for the physical conditions in these trials, and they say nothing about the accuracy of the optional event-bound visual conditions above.
Calibrated estimates may still direct a reviewer to runs worth inspecting.

\begin{table}[!htb]
\centering\small
\caption{Progress judges against physics on line unloading, over 36 labeled runs. $\tau_b$ is the Kendall rank correlation between judged progress and the completed-job count over 48 frames, step-up the fraction of job completions at which judged progress rises by at least 0.1, agreement the fraction of runs whose judged outcome matches physics among those the judge does not leave unclear, and false success the fraction of failed runs judged achieved. With shuffled frames the outcome is undefined. Time ramp reports the elapsed fraction of the run as progress.}
\label{tab:progress-judge-line}
\begin{tabular*}{\linewidth}{@{\extracolsep{\fill}}llcccc@{}}
\toprule
\tablehead{Judge (effort)} & \tablehead{Frames} & $\tau_b$ & Step-up & Agreement & False success\\
\midrule
Time ramp & Ordered & 0.53 & -- & -- & --\\
\addlinespace
gpt-6.1-sol (medium) & Ordered & 0.68 & 27\% & 78\% & 29\%\\
 & Shuffled & 0.73 & 52\% & -- & --\\
\addlinespace
gpt-6-astra (high) & Ordered & 0.69 & 26\% & 78\% & 29\%\\
 & Shuffled & 0.73 & 58\% & -- & --\\
\addlinespace
claude-opus-5-5 (high) & Ordered & 0.55 & 50\% & 76\% & 29\%\\
 & Shuffled & 0.61 & 52\% & -- & --\\
\addlinespace
gemini-4-argon (high) & Ordered & 0.63 & 44\% & 81\% & 25\%\\
 & Shuffled & 0.65 & 50\% & -- & --\\
\bottomrule
\end{tabular*}
\end{table}

\paragraph{Progress judge prompt}
\label{par:progress-prompt}
The judge receives the goal of the run, the frame indices with their times in the run, and five progress anchors from 0 (nothing relevant has changed) to 1 (the goal is visibly satisfied and stable), is told that the physics predicates decide success, and returns per-frame progress with a rationale and a whole-run outcome as JSON.
With shuffled frames, the order and times are withheld.

Validating the optional visual condition path requires independently labeled conditions, including physical failures, visual failures, and missing or occluded evidence.
A comparison with whole-run judging should hold the visual model and evidence budget fixed and report false acceptance, false rejection, and unresolved rates.
For $N>0$ assessed intervals, $N_a$ non-abstaining answers, and $E_a$ errors among those answers, coverage is $N_a/N$ and conditional error is $E_a/N_a$ when $N_a>0$.
Both are needed because abstaining on difficult cases can reduce error without improving coverage.
No such validation of event-bound visual conditions is reported here.

\end{document}